\documentclass{article} 
\usepackage{iclr2025_conference,times}

\usepackage{amsmath,amsfonts,bm}

\def\eqref#1{equation~\ref{#1}}

\def\1{\bm{1}}

\def\rvepsilon{{\mathbf{\epsilon}}}

\def\rvh{{\mathbf{h}}}

\def\rvs{{\mathbf{s}}}

\def\rvv{{\mathbf{v}}}

\def\rvx{{\mathbf{x}}}
\def\rvy{{\mathbf{y}}}
\def\rvz{{\mathbf{z}}}

\DeclareMathAlphabet{\mathsfit}{\encodingdefault}{\sfdefault}{m}{sl}
\SetMathAlphabet{\mathsfit}{bold}{\encodingdefault}{\sfdefault}{bx}{n}

\usepackage{hyperref}
\usepackage{url}

\usepackage{xspace}
\usepackage{dsfont}
\usepackage{afterpage}

\usepackage{enumitem}
\usepackage{booktabs}
\usepackage{caption} 
\usepackage{multirow} 
\usepackage{enumitem}
\usepackage{colortbl} 
\usepackage{bbm}
\usepackage{algorithm}
\usepackage{algpseudocode}                  
\usepackage{setspace}                       
\usepackage{lipsum} 
\usepackage{subcaption} 

\usepackage{mathtools} 
\usepackage{bm} 
\usepackage{soul} 
\usepackage{nicefrac}
\usepackage{csquotes}

\usepackage{duckuments}
\usepackage{wrapfig}

\newcommand{\eg}{\emph{e.g.}} 
\newcommand{\ie}{\emph{i.e.}}

\newcommand{\bz}{\mathbf{z}}

\newcommand{\lname}{REPresentation Alignment\xspace}
\newcommand{\sname}{REPA\xspace}
\newcommand{\bestfid}{1.80\xspace}
\newcommand{\bestfidinterval}{1.42\xspace}

\definecolor{cornellred}{rgb}{0.7, 0.11, 0.11}
\definecolor{cadmiumgreen}{rgb}{0.0, 0.42, 0.24}
\definecolor{aliceblue}{rgb}{0.91, 0.94, 0.97}
\definecolor{darkblue}{rgb}{0.83, 0.89, 0.97}
\definecolor{Red7}{rgb}{0.941, 0.243, 0.243}
\definecolor{Green7}{RGB}{55, 178, 77}
\definecolor{Blue9}{rgb}{0.098,0.3,0.9}

\usepackage{pifont}

\def\pz{{\phantom{0}}}
\sethlcolor{aliceblue}

\hypersetup{
  linkcolor = cornellred,
  citecolor  = cadmiumgreen,
  colorlinks = true,
  urlcolor = Blue9
}

\newif\ifunderreview
\underreviewfalse 

\iclrfinalcopy

\title{
Representation Alignment for Generation: \\
Training Diffusion Transformers \\
Is Easier Than You Think
}
\author{Sihyun Yu$^1${\quad}Sangkyung Kwak$^{1,3}{\quad}$Huiwon Jang$^1$\\
\textbf{Jongheon Jeong$^2${\quad}Jonathan Huang$^3${\quad}Jinwoo Shin$^{1}$\thanks{Equal advising. \hfill \textbf{Project page:} {\scriptsize\url{https://sihyun.me/REPA}}}{\quad}Saining Xie$^{4}$$^\ast$}  \\
$^1$KAIST\,\,\,$^2$Korea University\,\,\,$^3$Scaled Foundations\,\,\,$^4$New York University \\
}

\newcommand*{\ShowNotes}{} 

\ifdefined\ShowNotes
  \newcommand{\colornote}[3]{{\color{#1}\bf{#2: #3}\normalfont}}
\else
  \newcommand{\colornote}[3]{}
\fi

\begin{document}
\maketitle

\begin{abstract}
Recent studies have shown that the denoising process in (generative) diffusion models can induce meaningful (discriminative) representations inside the model, though the quality of these representations still lags behind those learned through recent self-supervised learning methods. We argue that one main bottleneck in training large-scale diffusion models \emph{for generation} lies in effectively learning these representations. Moreover, training can be made easier by incorporating high-quality external visual representations, rather than relying solely on the diffusion models to learn them independently. We study this by introducing a straightforward regularization called \emph{REPresentation Alignment (\sname)}, which aligns the projections of noisy input hidden states in denoising networks with clean image representations obtained from external, pretrained visual encoders. The results are striking: our simple strategy yields significant improvements in both training efficiency and generation quality when applied to popular diffusion and flow-based transformers, such as DiTs and SiTs.
For instance, our method can speed up SiT training by over 17.5$\times$, matching the performance (without classifier-free guidance) of a SiT-XL model trained for 7M steps in less than 400K steps. In terms of final generation quality, our approach achieves state-of-the-art results of FID=\bestfidinterval using classifier-free guidance with the guidance interval.
\end{abstract}

\begin{figure*}[ht!]
 \ifunderreview
\else
 \vspace{-0.21in}
\fi
    \centering
    \includegraphics[width=.37\linewidth]{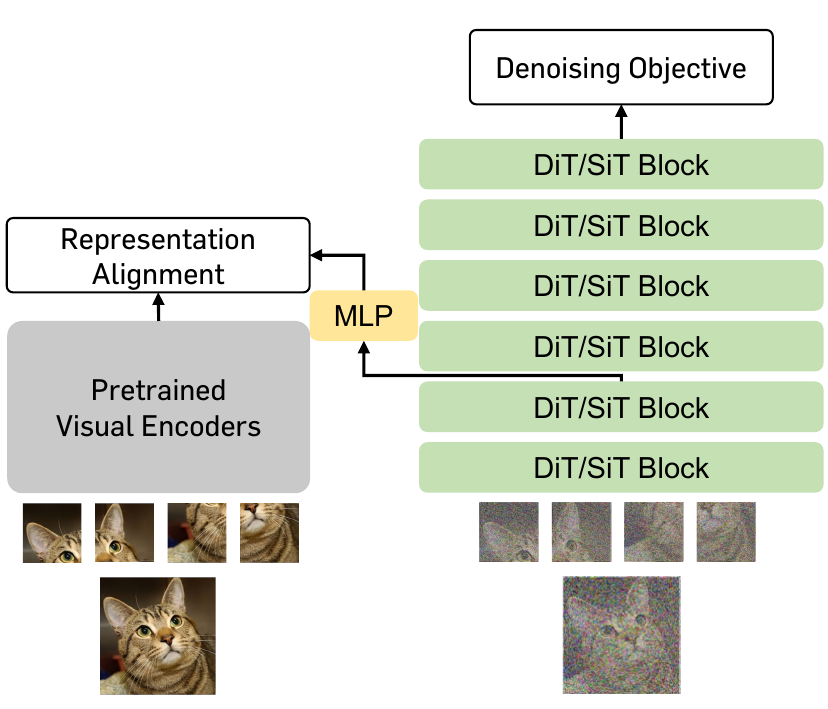}
    ~~~
    \includegraphics[width=.42\linewidth]{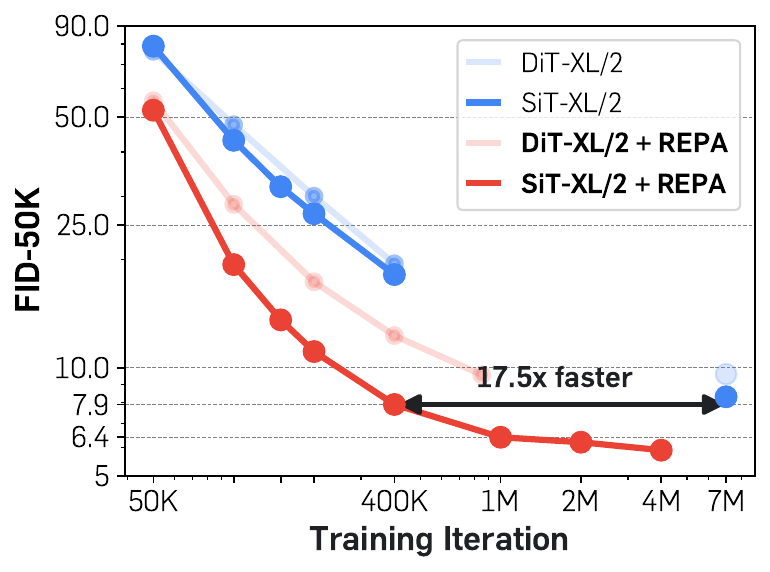}
    \caption{
    \textbf{Representation alignment makes diffusion transformer training significantly easier.}
    Our framework, \sname, explicitly aligns the diffusion model representation with powerful pretrained visual representation through a simple regularization. Notably, model training becomes significantly more efficient and effective, and achieves $>$17.5$\times$ faster convergence than the vanilla model.
    }
    \label{fig:teaser}
\end{figure*}
\vspace{-0.04in}
\section{Introduction}
\vspace{-0.04in}
\label{sec:intro}
Generative models based on \emph{denoising}, such as diffusion models~\citep{ho2021denoising,song2021scorebased} and flow matching models~\citep{albergo2023building,lipman2022flow,liu2022flow}, have been a scalable approach in generating high-dimensional visual data. They achieve remarkably successful results in challenging tasks such as zero-shot text-to-image~\citep{podell2023sdxl,saharia2022photorealistic, esser2024scaling} or text-to-video \citep{moviegen, videoworldsimulators2024} generation. 

Recent works have explored the use of diffusion models as representation learners~\citep{li2023your, xiang2023denoising,chen2024deconstructing,mukhopadhyay2023diffusion} and have shown that they learn discriminative features in their hidden states, and better diffusion models learn better representations \citep{xiang2023denoising}. In fact, this observation is closely related to earlier approaches that employ \emph{denoising score matching} \citep{vincent2011connection} as a self-supervised learning method \citep{bengio2013representation}, which implicitly learns a representation $\rvh$ as a hidden state of a denoising autoencoder $\rvs_{\theta}(\tilde{\rvx})$ through a \emph{reconstruction} of $\rvx$ from the corrupted data $\tilde{\rvx}$ \citep{yang2023diffusion}. However, the reconstruction task may not be a suitable task for learning good representations, as it is not capable of eliminating unnecessary details in $\rvx$ for representation learning \citep{lecun2022path, assran2023self}.

\begin{figure*}[t!]
\centering\small
\begin{subfigure}{.3\textwidth}
\centering
\includegraphics[width=\textwidth]{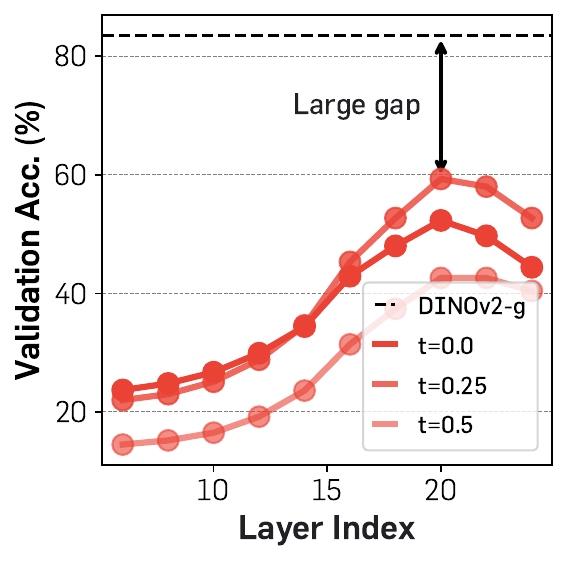}
\caption{Semantic gap: Linear probing} 
\label{subfig:sit_lin_eval}
\end{subfigure}
~~
\begin{subfigure}{.3\textwidth}
\centering
\includegraphics[width=1.\textwidth]{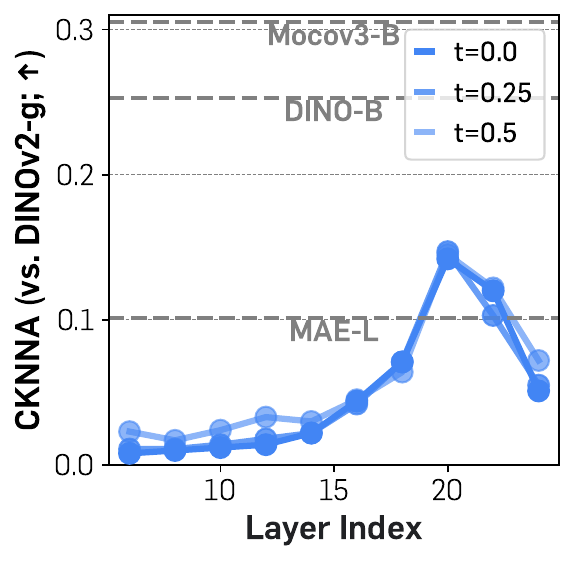}
\caption{Alignment to DINOv2-g} 
\label{subfig:sit_cknna_dino}
\end{subfigure}
~~~
\begin{subfigure}{.3\textwidth}
\centering
\includegraphics[width=1.\textwidth]{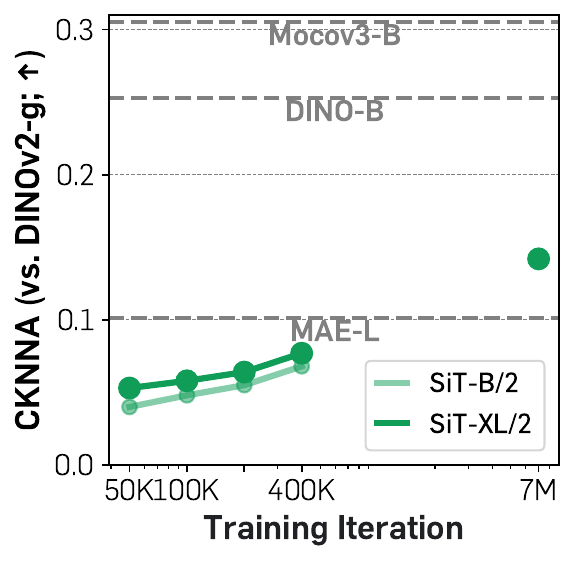}
\caption{Alignment progression} 
\label{subfig:sit_cknna_progression}
\end{subfigure}
\caption{
\textbf{Alignment behavior for a pretrained SiT model.} We empirically investigate the feature alignment between DINOv2-g and the original SiT-XL/2 checkpoint trained for 7M iterations. (a) While SiT learns semantically meaningful representations, a significant gap remains compared to DINOv2. (b) Using CKNNA \citep{huh2024platonic}, we observe that SiT already shows some alignment with DINOv2, though its absolute value is lower compared to other vision encoders. (c) Alignment improves with a larger model and longer training, but the progress remains slow and insufficient.}
\label{fig:sit_observation}
\vspace{-0.15in}
\end{figure*}

\textbf{Our approach.}
In this paper, we identify that the main challenge in training diffusion models stems from the need to learn a high-quality internal representation $\rvh$. We demonstrate that the training process for generative diffusion models becomes significantly easier and more effective when supported by an external representation, $\rvy_\ast$. Specifically, we propose a simple regularization technique that leverages recent advances in self-supervised visual representations as $\rvy_\ast$, leading to substantial improvements in both training efficiency and the generation quality of diffusion transformers.

We start by performing an empirical analysis with recent diffusion transformers~\citep{Peebles2022DiT, ma2024sit} and the state-of-the-art self-supervised vision model, DINOv2~\citep{oquab2024dinov}. 
Similar to prior studies \citep{xiang2023denoising}, we first observe that pretrained diffusion models do indeed learn meaningful discriminative representations (as shown by the linear probing results in Figure~\ref{subfig:sit_lin_eval}). However, these representations are significantly inferior to those produced by DINOv2.
Next, we find that the alignment between the representations learned by the diffusion model and those of DINOv2 (Figure~\ref{subfig:sit_cknna_dino}) is still considered weak,\footnote{We describe this as ``weak'' because relatively, the alignments are much poorer than those seen with other self-supervised encoders (\eg, MoCov3 \citep{chen2021empirical}), even after extensive training.} which we study by measuring their \emph{representation alignment} \citep{huh2024platonic}. Finally, we observe this alignment between diffusion models and DINOv2 improves consistently with longer training and larger models (Figure~\ref{subfig:sit_cknna_progression}).

These insights inspire us to enhance generative models by incorporating external self-supervised representations. However, this approach is not straightforward when using off-the-shelf self-supervised visual encoders (\eg, by fine-tuning an encoder for generation tasks). The first challenge is an input mismatch: diffusion models work with noisy inputs $\tilde{\rvx}$, whereas most self-supervised learning encoders are trained on clean images ${\rvx}$. This issue is even more pronounced in modern \emph{latent diffusion} models, which take a compressed latent image $\rvz = E(\rvx)$ from a pretrained VAE encoder \citep{rombach2022high} as input. Additionally, these off-the-shelf vision encoders are not designed for tasks like reconstruction or generation. To overcome these technical hurdles, we guide the feature learning of diffusion models using a \emph{regularization} technique that distills pretrained self-supervised representations into diffusion representations, offering a flexible way to integrate high-quality representations.

\begin{figure*}[t!]
\vspace{-0.15in}
\centering\small
\begin{subfigure}{.3\textwidth}
\centering
\includegraphics[width=\textwidth]{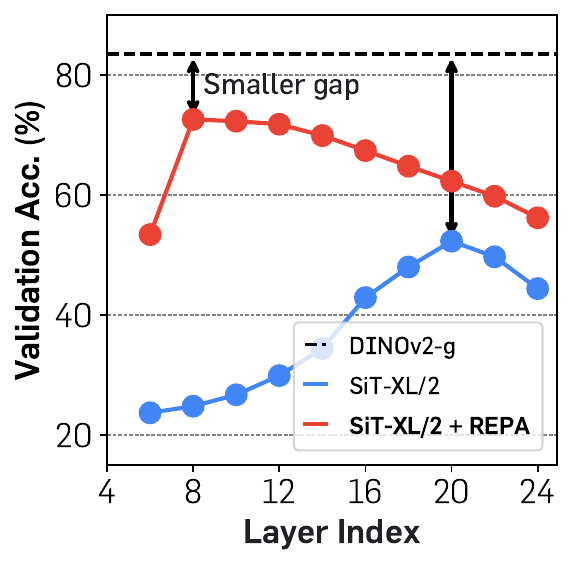}
\caption{Semantic gap: Linear probing} 
\label{subfig:lineval_diff}
\end{subfigure}
~
\begin{subfigure}{.3\textwidth}
\centering
\includegraphics[width=\textwidth]{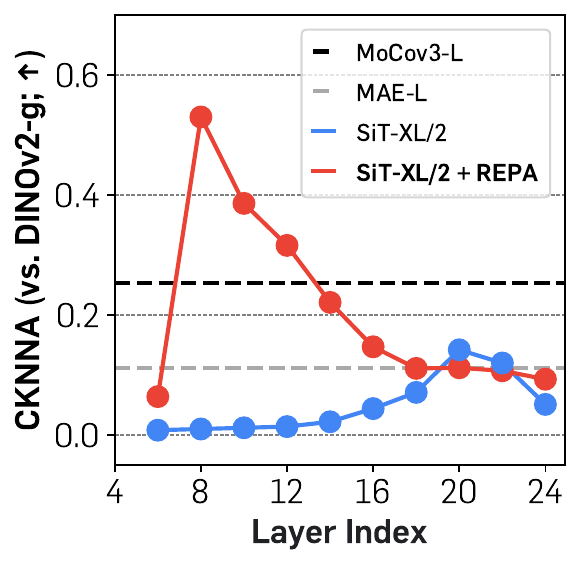}
\caption{Alignment to DINOv2-g} 
\label{subfig:cknna_diff}
\end{subfigure}
~
\begin{subfigure}{.3\textwidth}
\centering
\includegraphics[width=\textwidth]{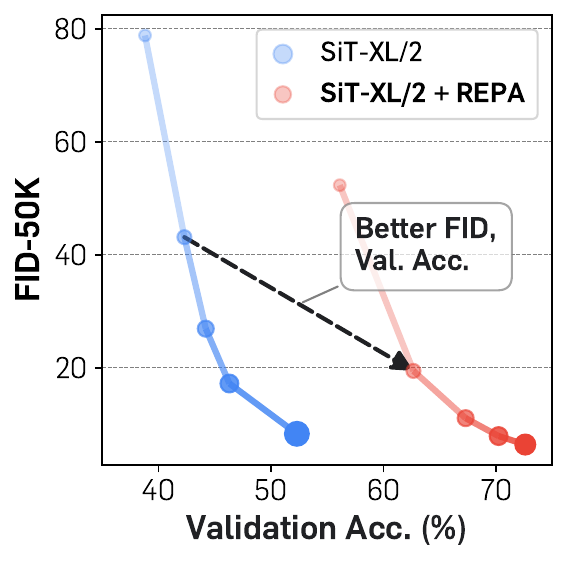}
\caption{Acc. and FID progression} 
\label{subfig:lineval_fid}
\end{subfigure}
\vspace{-0.05in}
\caption{
\textbf{Bridging the representation gap:} 
(a) Our method, \sname significantly reduces the ``semantic gap'' between diffusion transformers and DINOv2, as demonstrated by the linear probing results on ImageNet classification. (b) With \sname, the alignment between diffusion transformers and DINOv2 improves substantially, even after just a few (\eg, 8) layers. (c) Notably, with improved alignment, we can push the SiT model's generation-representation envelope: within the same number of training iterations, it delivers both better generation quality and stronger linear probing results. We use a single network trained with \sname at layer 8 and perform the evaluation at different layers.
}
\vspace{-0.12in}
\label{fig:summary}
\end{figure*}

Specifically, we introduce \emph{\lname} (\sname), a simple regularization technique built on recent diffusion transformer architectures \citep{Peebles2022DiT}. In essence, \sname distills the pretrained self-supervised visual representation $\rvy_\ast$ of a clean image $\rvx$ into the diffusion transformer representation $\rvh$ of a noisy input $\tilde{\rvx}$.  This regularization reduces the semantic gap in the representation $\rvh$ (Figure~\ref{subfig:lineval_diff}) and better aligns it with the target self-supervised representations $\rvy_\ast$ (Figure~\ref{subfig:cknna_diff}). Notably, this enhanced alignment significantly boosts the \emph{generation} performance of diffusion transformers (Figure~\ref{subfig:lineval_fid}). Interestingly, with \sname, we observe that sufficient representation alignment can be achieved by aligning only the first few transformer blocks. This, in turn, allows the later layers of the diffusion transformers to focus on capturing high-frequency details based on the aligned representations, further improving generation performance.

Based on our analysis, we conduct a system-level comparison to demonstrate the effectiveness of our scheme by applying it to two recent diffusion transformers: DiTs~\citep{Peebles2022DiT} and SiTs~\citep{ma2024sit}. For SiT training, we show the model achieves FID$=$7.9 on class-conditional ImageNet~\citep{deng2009imagenet} generation only using 400K training iteration (without classifier-free guidance; \citealt{ho2022classifier}) which is $>$17.5$\times$ faster than the vanilla SiTs. Moreover, with classifier-free guidance, our scheme shows an improved FID at the final from 2.06 to \bestfid and achieves state-of-the-art results of FID$=$\bestfidinterval with guidance interval \citep{kynkaanniemi2024applying}.

We highlight the main contributions of this paper below:
\vspace{-0.05in}
\begin{itemize}[leftmargin=*,itemsep=0mm]
    \item We hypothesize that learning high-quality representations in diffusion transformers is essential for improving their generation performance.
    \item We introduce \sname, a simple regularization for aligning diffusion transformer representations with strong self-supervised visual representations.
    \item Our framework improves the generation performance of diffusion transformers, \eg, for SiTs, we achieve a 17.5$\times$ faster training for SiTs and improved FID scores on ImageNet generation.
\end{itemize}
\section{Preliminaries}
\label{sec:prelim}
We present a brief overview of \emph{flow and diffusion-based} models through the unified perspective of \emph{stochastic interpolants}~\citep{albergo2023stochastic,ma2024sit}; see Appendix~\ref{appen:si} for more details.

We consider a continuous time-dependent process with a data $\rvx_\ast \sim p(\rvx)$ and a Gaussian noise $\rvepsilon~\sim \mathcal{N}(\mathbf{0}, \mathbf{I})$ on $t \in [0, T]$:
\begin{equation}
    \rvx_t = \alpha_t \rvx_\ast+ \sigma_t \rvepsilon,\quad \alpha_0 = \sigma_T = 1,\,\, \alpha_T = \sigma_0 = 0,
    \label{eq:flow}
\end{equation}
where $\alpha_t$ and $\sigma_t$ are a decreasing and increasing function of $t$, respectively. Given such a process, there exists a \emph{probability flow ordinary differential equation} (PF ODE) with a velocity field
\begin{equation}
\dot{\rvx}_t = \rvv (\rvx_t, t),
\label{eq:pf_ode}
\end{equation}
where the distribution of this ODE at $t$ is equal to the marginal $p_t(\rvx)$. Thus, data can be sampled by solving this PF ODE in Eq.~(\ref{eq:pf_ode}) through existing ODE samplers (\eg, Euler sampler) starting from a random Gaussian noise $\epsilon \sim \mathcal{N} (\mathbf{0}, \mathbf{I})$ \citep{lipman2022flow,ma2024sit}.

This velocity $\rvv (\rvx, t)$ is represented as the following sum of two conditional expectations
\begin{equation}
    \rvv (\rvx, t) = \mathbb{E}[\dot{\rvx}_t| \rvx_t = \rvx] = \dot{\alpha}_t \mathbb{E}[\rvx_\ast| \rvx_t = \rvx] + \dot{\sigma}_t\mathbb{E}[\rvepsilon| \rvx_t = \rvx], 
\end{equation}
which can be approximated with model $\rvv_{\theta}(\rvx_t, t)$ by minimizing the following training objective:
\begin{equation}
    \mathcal{L}_{\text{velocity}}(\theta)\coloneqq
    \mathbb{E}_{\rvx_\ast, \bm{\epsilon}, t}
    \big[
    ||\rvv_{\theta}(\rvx_t, t) - \dot{\alpha}_t\rvx_\ast - \dot{\sigma}_t\rvepsilon||^2 
    \big].
    \label{eq:v_prediction}
\end{equation}

Moreover, there exists a reverse \emph{stochastic differential equation} (SDE) that the marginal $p_t(\rvx)$ coincides with the one of PF ODE in Eq.~(\ref{eq:pf_ode}) with a diffusion coefficient $w_t$ \citep{ma2024sit}:
\begin{equation}
    d\rvx_t = \rvv(\rvx_t, t)dt -\frac{1}{2}w_t \rvs(\rvx_t, t)dt + \sqrt{w_t} d\bar{\mathbf{w}}_t,
    \label{eq:sde}
\end{equation}
where the score $\rvs(\rvx_t, t)$ is the following conditional expectation 
\begin{equation}
    \rvs(\rvx_t, t) = -{\sigma_t^{-1}}\mathbb{E}[\rvepsilon| \rvx_t = \rvx].
\end{equation}
and it can be directly computed using the velocity $\rvv(\rvx, t)$ for $t > 0$ as 
\begin{equation}
    \rvs(\rvx, t) = {{\sigma_t^{-1}}} \cdot \frac{\alpha_t\rvv(\rvx, t) - \dot{\alpha}_t\rvx}{\dot{\alpha}_t\sigma_t - {\alpha}_t\dot{\sigma}_t},
\end{equation}
implying that data can be alternatively generated through Eq.~(\ref{eq:sde}) with SDE solvers.

Following \citet{ma2024sit}, we mainly consider a simple linear interpolant with restricting $T=1$: $\alpha_t = 1-t$ and $\sigma_t = t$. However, our approach is applicable to any similar variants (\eg, DDPM; \citealt{ho2021denoising}), which has a similar formulation but uses a discretized process and different $\alpha_t, \sigma_t$ that $\mathcal{N} (\mathbf{0}, \mathbf{I})$ becomes an equilibrium distribution (\ie, $\rvx_t$ converges to $\mathcal{N} (\mathbf{0}, \mathbf{I})$ only if $ t\to \infty$).
\section{\sname: Regularization for Representation Alignment}
\label{sec:method}

\subsection{Overview}
\label{subsec:overview}

Let $p(\rvx)$ be an unknown target distribution for data $\rvx \in \mathcal{X}$. Our goal is to approximate $p(\rvx)$ through a model distribution using a dataset drawn from $p(\rvx)$. To lower computational costs, we adopt the recent prevalent \emph{latent diffusion} \citep{rombach2022high}. This involves learning a latent distribution $p(\rvz)$, which is defined as the distribution of a compressed latent variable $\bz = E(\rvx)$, where $E$ is an encoder from a pretrained autoencoder (\eg, KL-VAE; \citealt{rombach2022high}), with $\rvx \sim p_{\text{data}}(\rvx)$. 

We aim to learn this distribution by training a diffusion model $\rvv_{\theta}(\rvz_t, t)$ using objectives such as velocity prediction, as described in Section~\ref{sec:prelim}. Here, we revisit denoising score matching within the context of self-supervised representation learning \citep{bengio2013representation}. From this perspective, one can think of the diffusion model $\rvv_{\theta}(\rvz_t, t)$ as a composition of two functions $g_{\theta} \circ f_{\theta}$ with an encoder $f_\theta: \mathcal{Z} \to \mathcal{H}$ 
with $f_{\theta}(\rvz_t) = \rvh_t$ and a decoder $g_\theta: \mathcal{H} \to \mathcal{Z}$ with $g_{\theta}(\rvh_t) = \rvv_t$, where the encoder $f_\theta$ implicitly learns a representation $\rvh_t$ that reconstructs the target $\rvv_t$.

However, learning a good representation through producing a prediction of the input space (\eg, generating pixels) can be challenging, as the model is often not capable of eliminating unnecessary details, which is crucial for developing a strong representation \citep{lecun2022path,assran2023self}. We argue that a key bottleneck in the training of large-scale diffusion models \emph{for generation} lies in representation learning, an area where current diffusion models fall short. We also hypothesize that the training process can be made easier by guiding the model with high-quality external visual representations, rather than relying solely on the diffusion model to learn them independently.

To address this challenge, we introduce a simple regularization method called \emph{\lname} (\sname) using the recent diffusion transformer architectures \citep{Peebles2022DiT,ma2024sit} (see Appendix~\ref{appen:architecture_detail} for an illustration). In a nutshell, our regularization distills pretrained self-supervised visual representations to diffusion transformers in a simple and effective way. This allows the diffusion model to leverage these semantically rich external representations for generation, leading to a substantial boost in performance.

\begin{figure*}[t!]
    \vspace{-0.15in}
    \centering
    \includegraphics[width=.99\linewidth]{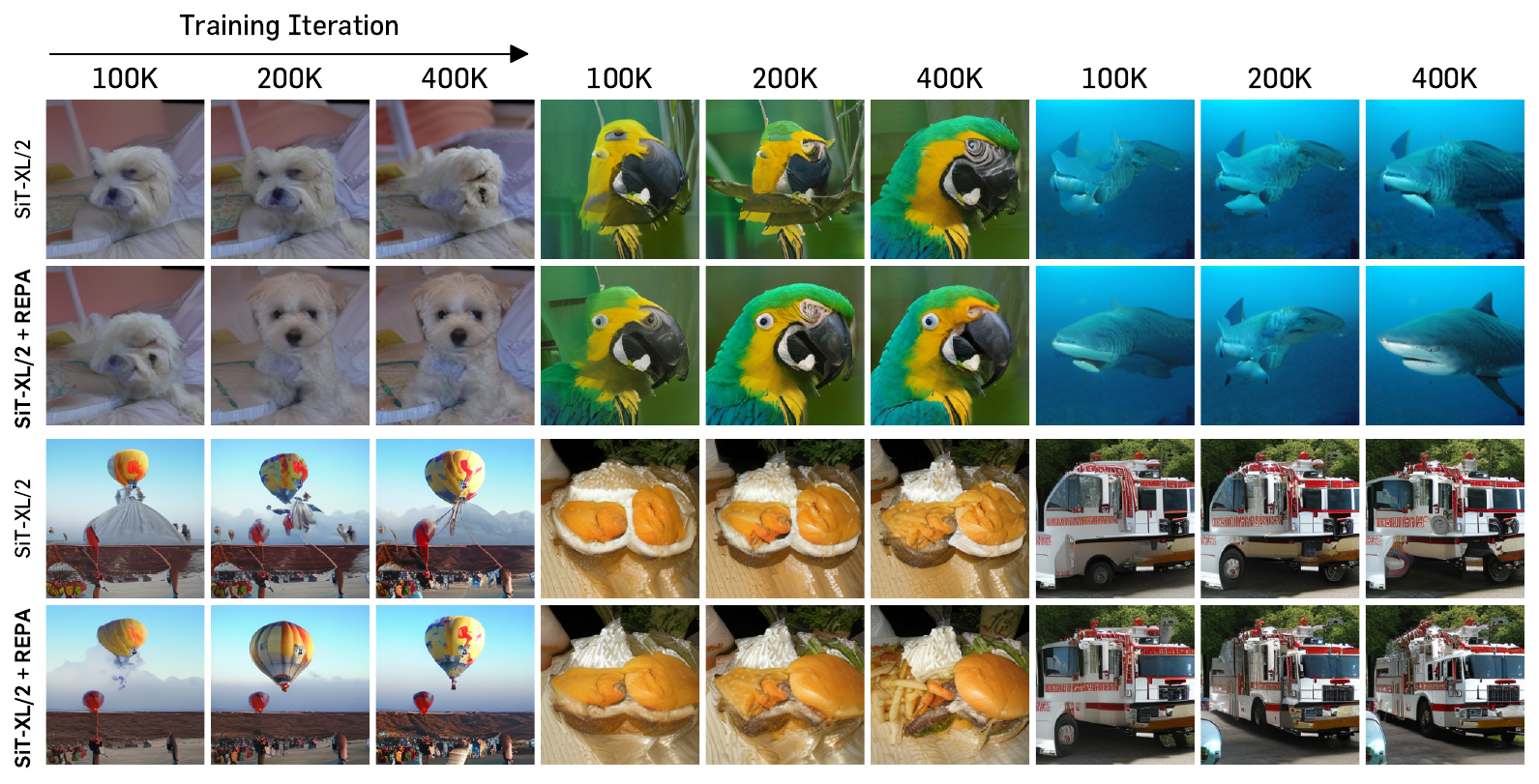}
    \vspace{-0.05in}
    \caption{\textbf{\sname improves visual scaling.} We compare the images generated by two SiT-XL/2 models during the first 400K iterations, with \sname applied to one of the models. Both models share the same noise, sampler, and number of sampling steps, and neither uses classifier-free guidance.}
    \label{fig:qual_progression}
    \vspace{-0.15in}
\end{figure*}

\subsection{Observations}
\label{subsec:hypothesis}

To take a deeper dive into this, we first investigate the layer-wise behavior of the pretrained SiT model~\citep{ma2024sit} on ImageNet~\citep{deng2009imagenet}, which uses linear interpolants and velocity prediction for training. In particular, we focus on measuring the
\emph{representation gap} between the diffusion transformer and the state-of-the-art self-supervised DINOv2 model \citep{oquab2024dinov}. We examine this from three angles: semantic gap, feature alignment progression, and their final feature alignment. For the \emph{semantic gap}, we compare linear probing results using DINOv2 features with those from SiT models trained for 7M iterations, following the same protocol as in \citet{xiang2023denoising}, which involves linear probing on globally pooled hidden states of the diffusion transformer. Next, to measure \emph{feature alignments}, we use CKNNA \citep{huh2024platonic}, a kernel alignment metric related to CKA \citep{kornblith2019similarity}, but based on mutual nearest neighbors. This allows for a quantitative assessment of alignment between different representations. We summarize the result in Figure~\ref{fig:sit_observation} and more details (\eg, definition of CKNNA) in Appendix~\ref{appen:subsec:metric}. 

\textbf{Diffusion transformers exhibit a significant semantic gap from state-of-the-art visual encoders.} As shown in Figure~\ref{subfig:sit_lin_eval}, we observe that the hidden state representation of the pretrained diffusion transformer, in line with prior works~\citep{xiang2023denoising, chen2024deconstructing}, achieves a reasonably high linear probing peak at layer 20. However, its performance remains well below that of DINOv2, indicating a substantial semantic gap between the two representations. Additionally, we find that after reaching this peak, linear probing performance quickly declines, suggesting that the diffusion transformer must shift away from focusing solely on learning semantically-rich representations in order to generate images with high-frequency details.

\textbf{Diffusion representations are already (weakly) aligned with other visual representations.}
In Figure \ref{subfig:sit_cknna_dino}, we report representational alignments between SiT and DINOv2 using CKNNA. In particular, the SiT model representation already shows better alignment than MAE \citep{he2022masked}, which is also a self-supervised learning approach based on the reconstruction of masked patches. However, the absolute alignment score remains lower than that observed between other self-supervised learning methods (\eg, MoCov3 \citep{chen2021empirical} \emph{vs.} DINOv2). These results suggest that while diffusion transformer representations exhibit some alignment with self-supervised visual representations, the alignment remains weak.

\textbf{Alignment improves with larger models and extended training.} We also measure CKNNA values across different model sizes and training iterations. As depicted in Figure~\ref{subfig:sit_cknna_progression}, we observe improved alignment with larger models and extended training. However, the absolute alignment remains low and does not reach the levels observed between other self-supervised visual encoders (\eg, MoCov3 and DINOv2), even after extensive training of 7M iterations.

These findings are not unique to the SiT model but are also observed in other denoising-based generative transformers. 
For instance, in Figure~\ref{fig:sit_observation}, we present a similar analysis using a DiT model~\citep{Peebles2022DiT} pretrained on ImageNet with the DDPM objective~\citep{ho2021denoising,nichol2021improved}. See Appendix~\ref{appen:subsec:dit_analysis} for more details.

\subsection{Representation alignment with self-supervised representations}
\label{subsec:method}
\sname aligns patch-wise projections of the model's hidden states with pretrained self-supervised visual representations. Specifically, we use the \emph{clean} image representation as the target and explore its impact. The goal of this regularization is for the diffusion transformer's hidden states to predict noise-invariant, clean visual representations from noisy inputs that contain useful semantic information. This provides meaningful guidance for the subsequent layers to reconstruct the target.

Formally, let $f$ be a pretrained encoder and consider a clean image $\mathbf{x}_\ast$. Let $\rvy_\ast = f(\rvx_\ast) \in \mathbb{R}^{N \times D}$ be an encoder output, where $N, D>0$ are the number of patches and the embedding dimension of $f$, respectively. \sname aligns $h_{\phi}(\rvh_t) \in \mathbb{R}^{N \times D}$ with $\rvy_\ast$, where $h_{\phi}(\rvh_t)$ is a projection of an diffusion transformer encoder output $\rvh_t = f_{\theta} (\rvz_t)$ that through a trainable projection head $h_{\phi}$. In practice, we simply parameterize $h_{\phi}$ using a multilayer perceptron (MLP). 

In particular, \sname achieves alignment through a maximization of patch-wise similarities between the pretrained representation $\rvy_\ast$ and the hidden state $\rvh_{t}$:
\begin{equation}
    \mathcal{L}_{\text{REPA}}(\theta, \phi) \coloneqq -\mathbb{E}_{\rvx_\ast, \bm{\epsilon}, t} \Big[ \frac{1}{N}\sum_{n=1}^{N} \mathrm{sim}(\rvy_\ast^{[n]}, h_{\phi}(\rvh_{t}^{[n]})) \Big],
\end{equation}
where $n$ is a patch index and $\mathrm{sim}(\cdot,\cdot)$ is a pre-defined similarity function. 

In practice, we add this term to the original diffusion-based objectives described in Section~\ref{sec:prelim} and Appendix~\ref{appen:si}. For instance, for the training of a velocity model in Eq.~(\ref{eq:v_prediction}), the objective becomes:
\begin{equation}
    \mathcal{L} \coloneqq \mathcal{L}_{\text{velocity}} + \lambda\mathcal{L}_{\text{REPA}}
\end{equation}
where $\lambda>0$ is a hyperparameter that controls the tradeoff between denoising and representation alignment. We primarily investigate the impact of this regularization on two popular objectives: Improved DDPM \citep{nichol2021improved} used in DiT~\citep{Peebles2022DiT} and linear stochastic interpolants used in SiT~\citep{ma2024sit}, though other objectives can also be considered.
\section{Experiments}
\label{sec:exp}

We validate the performance of \sname and the effect of the proposed components through extensive experiments. In particular, we investigate the following questions:
\begin{itemize}[leftmargin=*,itemsep=0mm]
\item Can \sname improve diffusion transformer training significantly? (Table~\ref{tab:detailed_design}, \ref{tab:wo_cfg}, \ref{tab:main}, Figure~\ref{fig:qual_progression}, \ref{fig:main_qual})
\item Is \sname scalable in terms of model size and representation quality? (Table~\ref{tab:detailed_design}, Figure \ref{fig:scalability})
\item Can diffusion model representations be aligned with various visual representations? (Figure \ref{fig:abla_different_enc})
\end{itemize}

\subsection{Setup}
\label{subsec:setup}
\begin{wrapfigure}[6]{R}{0.4\textwidth}
\centering\small
\vspace{-0.18in}
\captionof{table}{
Model configuration details.
}
\vspace{-0.1in}
\resizebox{0.38\textwidth}{!}{
\begin{tabular}{l c c c}
    \toprule
     Config & \#Layers & Hidden dim & \#Heads  \\
     \midrule
     B/2 & 12 & 768 & 12 \\
     L/2 & 24 & 1024 & 16 \\
     XL/2 & 28 & 1152 & 16 \\
     \bottomrule
     & 
\end{tabular}
}
\label{tab:config}
\vspace{-0.3in}
\end{wrapfigure}
\textbf{Implementation details.}
We strictly follow the setup in DiT~\citep{Peebles2022DiT} and SiT~\citep{ma2024sit} unless otherwise specified. Specifically, we use ImageNet~\citep{deng2009imagenet}, where each image is preprocessed to the resolution of 256$\times$256 (denoted as ImageNet 256$\times$256), and follow ADM~\citep{dhariwal2021diffusion} for other data preprocessing protocols. Each image is then encoded into a compressed vector $\rvz \in \mathbb{R}^{32 \times 32 \times 4}$ using the Stable Diffusion VAE~\citep{rombach2022high}. For model configurations, we use the B/2, L/2, and XL/2 architectures introduced in the DiT and SiT papers, which process inputs with a patch size of 2 (see Table~\ref{tab:config} for details). To ensure a fair comparison with DiTs and SiTs, we consistently use a batch size of 256 during training. 
Additional experimental details, including hyperparameter settings and computing resources, are provided in Appendix~\ref{appen:main_setup}.

\begin{table}[t!]
\centering\small
\captionof{table}{
\textbf{Component-wise analysis} on ImageNet 256$\times$256. All models are SiT-L/2 trained for 400K iterations. All metrics except accuracy (Acc.) are measured with the SDE Euler-Maruyama sampler with NFE=250 and without classifier-free guidance. For Acc., we report linear probing results on the ImageNet validation set using the latent features aligned with the target representation. We fix $\lambda=0.5$ here. $\downarrow$ and $\uparrow$ indicate whether lower or higher values are better, respectively.
}
\vspace{-0.07in}
\begin{tabular}{c l c c c c c c c c}
\toprule
Iter. & \multicolumn{1}{l}{Target Repr.} &  Depth & Objective  &{FID$\downarrow$} & {sFID$\downarrow$} & {IS$\uparrow$} & {Pre.$\uparrow$} & {Rec.$\uparrow$} & {Acc.$\uparrow$}\\
\midrule
\rowcolor{gray!15}
400K & \multicolumn{3}{l}{Vanilla SiT-L/2 \citep{ma2024sit}} & 18.8  & 5.29 & \pz72.0 & 0.64 & 0.64 & N/A\\
\midrule
400K & \cellcolor{red!15}MAE-L    & 8 & NT-Xent  & 12.5 & 4.89 & \pz90.7 & 0.68 & 0.63 & 57.3 \\
400K & \cellcolor{red!15}DINO-B   & 8 & NT-Xent  & 11.9 & 5.00 & \pz92.9 & 0.68 & 0.63 & 59.3 \\
400K & \cellcolor{red!15}MoCov3-L & 8 & NT-Xent  & 11.9 & 5.06 & \pz92.2 & 0.68 & 0.64 & 63.0 \\
400K & \cellcolor{red!15}I-JEPA-H & 8 & NT-Xent  & 11.6 & 5.21 & \pz98.0 & 0.68 & 0.64 & 62.1\\
400K & \cellcolor{red!15}CLIP-L   & 8 & NT-Xent  & 11.0 & 5.25 & 100.4 & 0.67 & 0.66 & 67.2 \\
400K & \cellcolor{red!15}SigLIP-L & 8 & NT-Xent  & 10.2 & 5.15 & 107.0 & 0.69 & 0.64 & 68.8 \\
400K & \cellcolor{red!15}DINOv2-L & 8 & NT-Xent  & 10.0 & 5.09 & 106.6 & 0.68 & 0.65 & 68.1 \\
\midrule
400K & \cellcolor{yellow!15}DINOv2-B    & 8 & NT-Xent  & \pz9.7 & 5.13 & 107.5 & 0.69 & 0.64 & 65.7 \\
400K & \cellcolor{yellow!15}DINOv2-L    & 8 & NT-Xent  & 10.0 & 5.09 & 106.6 & 0.68 & 0.65 & 68.1 \\
400K & \cellcolor{yellow!15}DINOv2-g    & 8 & NT-Xent  & \pz9.8 & 5.22 & 108.9 & 0.69 & 0.64 & 65.7\\
\midrule
400K & DINOv2-L  & \cellcolor{green!15}6  & NT-Xent  & 10.3 & 5.23 & 106.5 & 0.69 & 0.65 & 66.2 \\
400K & DINOv2-L  & \cellcolor{green!15}8  & NT-Xent  & 10.0 & 5.09 & 106.6 & 0.68 & 0.65 & 68.1 \\
400K & DINOv2-L & \cellcolor{green!15}10 & NT-Xent  & 10.5 & 5.50 & 105.0 & 0.68 & 0.65 & 68.6 \\
400K & DINOv2-L  & \cellcolor{green!15}12 & NT-Xent  & 11.2 & 5.14 & 100.2 & 0.68 & 0.64 & 69.4 \\
400K & DINOv2-L  & \cellcolor{green!15}14 & NT-Xent  & 11.6 & 5.61 & \pz99.5 & 0.67 & 0.65 & 70.0 \\
400K & DINOv2-L  & \cellcolor{green!15}16 & NT-Xent  & 12.1 & 5.34 & \pz96.1 & 0.67 & 0.64 & 71.1 \\
\midrule
400K & DINOv2-L  & 8 & \cellcolor{blue!15}NT-Xent    & 10.0 & 5.09 & 106.6 & 0.68 & 0.65 & 68.1 \\
400K & DINOv2-L  & 8 & \cellcolor{blue!15}Cos. sim. & \pz9.9 & 5.34 & 111.9 & 0.68 & 0.65 & 68.2\\
\bottomrule
\end{tabular}
\label{tab:detailed_design}
\vspace{-0.15in}
\end{table}

\textbf{Evaluation.}
We report Fr\'echet inception distance (FID; \citealt{heusel2017gans}), sFID \citep{nash2021generating}, inception score (IS; \citealt{salimans2016improved}), precision (Pre.) and recall (Rec.)
\citep{kynkaanniemi2019improved} using 50,000 samples. We also include linear probing results (Acc.) and CKNNA~\citep{huh2024platonic} as discussed in Section~\ref{subsec:hypothesis}. We provide more details of each metric in Appendix~\ref{appen:eval}.

\textbf{Sampler.}
Following SiT \citep{ma2024sit}, we always use the SDE Euler-Maruyama sampler (for SDE with $w_t = \sigma_t$) and set the number of function evaluations (NFE) as 250 by default.

\textbf{Baselines.}
We use several recent diffusion-based generation methods as baselines, each employing different inputs and network architectures. Specifically, we consider the following four types of approaches: (a) \emph{Pixel diffusion}: ADM~\citep{dhariwal2021diffusion}, VDM$++$~\citep{kingma2024understanding}, Simple diffusion~\citep{hoogeboom2023simple}, CDM~\citep{ho2022cascaded}, (b) \emph{Latent diffusion with U-Net}: LDM~\citep{rombach2022high}, (c) \emph{Latent diffusion with transformer+U-Net hybrid models}: U-ViT-H/2 \citep{bao2022all}, DiffiT~\citep{hatamizadeh2023diffit}, and MDTv2-XL/2 \citep{gao2023mdtv2}, and (d) \emph{Latent diffusion with transformers}: MaskDiT \citep{zheng2024fast}, SD-DiT \citep{zhu2024sd}, DiT~\citep{Peebles2022DiT}, and SiT~\citep{ma2024sit}. Here, we refer to Transformer+U-Net hybrid models that contain skip connections, which are not originally used in pure transformer architecture. Detailed descriptions of each baseline method are provided in Appendix~\ref{appen:baselines}. 

\subsection{Component-wise analysis}
We answer the question of whether \sname leads to improved diffusion transformer training. As shown in Table~\ref{tab:detailed_design}, we discover that \sname consistently provides a substantially improved generation performance across various design choices, achieving a much better FID score than the vanilla model. Below, we provide a detailed analysis of the impact of each component.

\sethlcolor{red!15}
\hl{\textbf{Target representation.}}
We begin by analyzing the effect of using different pretrained self-supervised encoders as the target representation. Notably, there is a strong correlation between the quality of these encoders and the performance of the corresponding aligned diffusion transformers. When a diffusion transformer is aligned with a pretrained encoder that offers more semantically meaningful representations (\ie, better linear probing results), the model not only captures better semantics but also exhibits enhanced generation performance, as reflected by improved validation accuracy with linear probing and lower FID scores.

\begin{figure*}[t!]
\vspace{-0.1in}
\centering\small
\begin{subfigure}{.32\textwidth}
\centering
\includegraphics[width=.98\textwidth]{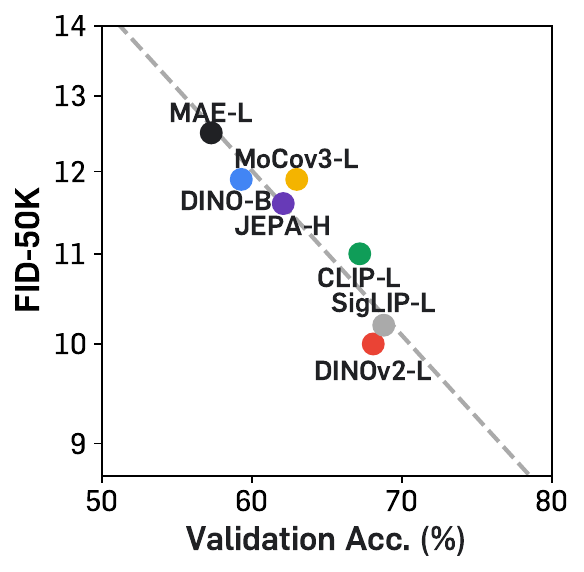}
\caption{Different visual encoders} 
\label{subfig:diff_encoders}
\end{subfigure}
\begin{subfigure}{.32\textwidth}
\centering
\includegraphics[width=\textwidth]{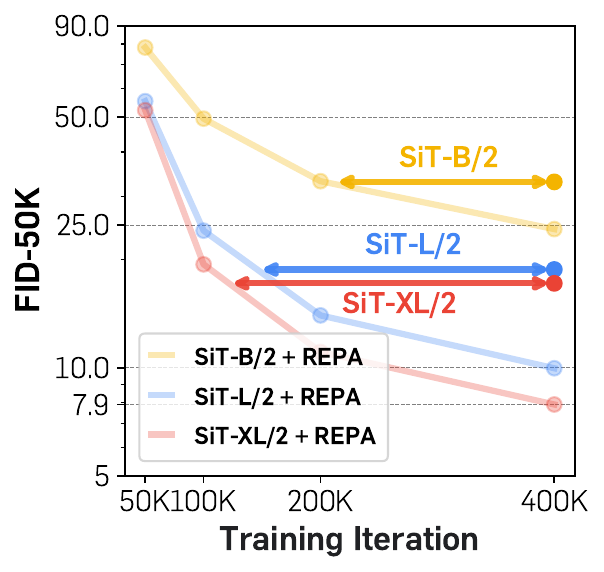}
\caption{Relative convergence} 
\label{subfig:scaling_law}
\end{subfigure}
\begin{subfigure}{.32\textwidth}
\centering
\includegraphics[width=.98\textwidth]{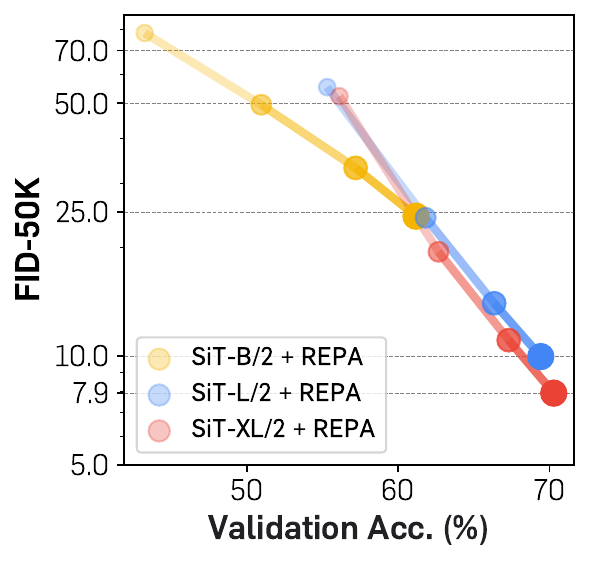}
\caption{Validation acc. vs. FID} 
\label{subfig:model_size}
\end{subfigure}
\caption{
\textbf{\sethlcolor{violet!15} \hl{Scalability} of \sname.} (a) Linear probing vs. FID plot of \sname with different target encoders (400K iterations). A stronger encoder improves both discrimination and generation performance. (b) The relative improvement of \sname over the vanilla model becomes increasingly significant as the model size grows. (c) With a fixed target encoder, larger models reach better performance more quickly. In the line plot, results are marked at 50K, 100K, 200K, and 400K iters.}
\label{fig:scalability}
\vspace{-0.1in}
\end{figure*}

\begin{figure*}[t!]
    \centering
    \includegraphics[width=.96\linewidth]{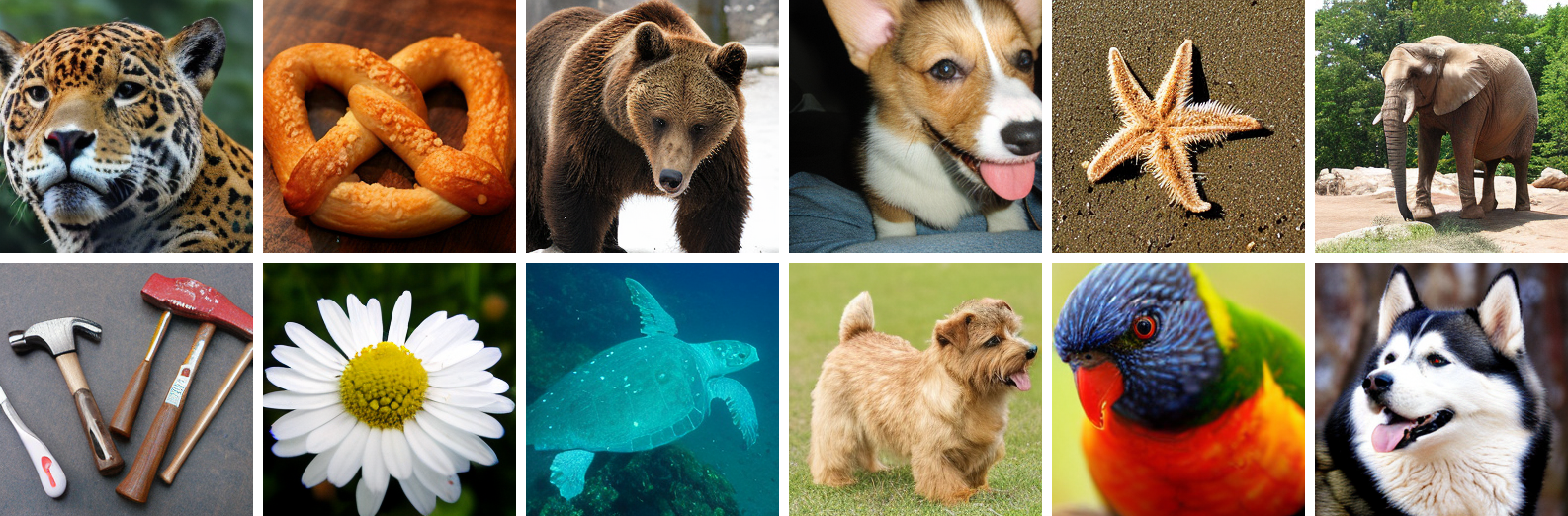}
    \caption{\textbf{Selected samples on ImageNet 256$\times$256} from the SiT-XL/2 + \sname model. We use classifier-free guidance with $w=4.0$. 
    }
    \label{fig:main_qual}
    \vspace{-0.15in}
\end{figure*}
\sethlcolor{yellow!15}
\hl{\textbf{Target encoder size.}}
Next, we investigate the impact of different target representation encoder sizes by evaluating various DINOv2 models (\ie, DINOv2-B, -L, -g). We observe that the performance differences are marginal, which we hypothesize is due to all DINOv2 models being distilled from the DINOv2-g model and thus sharing similar representations.

\sethlcolor{green!15}
\hl{\textbf{Alignment depth.}}
We also examine the effect of attaching the \sname loss to different layers. We find that regularizing only the first few layers (\eg, 8) in training is sufficient, as indicated by the linear probing results in Table~\ref{tab:detailed_design}. Interestingly, limiting regularization to the first few layers further enhances generation performance (\eg, adding \sname to layer 6 or 8 yields best results). We hypothesize that this enables the remaining layers to concentrate on capturing high-frequency details, building on a strong representation. In future experiments, we apply \sname to the first 8 layers.

\sethlcolor{blue!15}
\hl{\textbf{Alignment objective.}} We compare two simple training objectives for alignment: Normalized Temperature-scaled Cross Entropy (NT-Xent; \citealt{chen2020simple}) or negative cosine similarity (cos. sim.). Empirically, we find that NT-Xent offers advantages in the early stages (\eg, 50-100K iterations), but the gap diminishes over time. Thus, we opt for cos. sim. in future experiments.

\sethlcolor{violet!15}
\hl{\textbf{Scalability.}} Lastly, we investigate the scalability of \sname by varying the model sizes of both the target representation encoders and the diffusion transformers. In general, as summarized in Figure~\ref{subfig:diff_encoders}, aligning with stronger representations improves both the generation results and the linear probing performance. Moreover, the convergence speed-up from \sname becomes more significant as the diffusion transformer model increases in size. We demonstrate this by plotting FID-50K of different SiT models with and without \sname in Figure~\ref{subfig:scaling_law}: \sname achieves the same FID level more quickly with larger models. Lastly, Figure~\ref{subfig:model_size} highlights the relationship between linear probing results and FID scores as model size varies, while keeping the target representation encoder fixed as DINOv2-B. Larger models exhibit a steeper performance improvement (\ie, faster gains in both generation and linear evaluation) with longer training.

\begin{figure}[t!]
\vspace{-0.05in}
\begin{minipage}{0.415\textwidth}
\centering\small
\captionof{table}{\textbf{FID comparisons with vanilla DiTs and SiTs} on ImageNet 256$\times$256. We do not use classifier-free guidance (CFG). $\downarrow$ denotes lower values are better. Iter. indicates the training iteration.
}
\vspace{-0.05in}
\resizebox{\textwidth}{!}{%
\begin{tabular}{lccc}
\toprule
     Model & \#Params & Iter. & FID$\downarrow$  \\
     \midrule
     DiT-L/2 & 458M & 400K & 23.3 \\
     {\textbf{+ \sname (ours)}} & 458M & \textbf{400K}  & \textbf{15.6}\\
     \midrule
     DiT-XL/2 & 675M & 400K & 19.5\\
     {\textbf{+ \sname (ours)}}  & 675M & \textbf{400K}  & \textbf{12.3}\\
     \arrayrulecolor{black!40}\midrule
     DiT-XL/2 & 675M & {7M} & \pz9.6\\
     {\textbf{+ \sname (ours)}}  & 675M & \textbf{850K}  & \textbf{\pz9.6}\\
     \arrayrulecolor{black}\midrule
     SiT-B/2 & 130M & 400K & 33.0 \\
     {\textbf{+ \sname (ours)}} & 130M & \textbf{400K}  & \textbf{24.4}\\
     \arrayrulecolor{black!40}\midrule
     SiT-L/2 & 458M & 400K & 18.8 \\
     {\textbf{+ \sname (ours)}} & 458M & \textbf{400K}  & \textbf{\pz9.7}\\
     {\textbf{+ \sname (ours)}} & 458M & \textbf{700K} & \textbf{\pz8.4} \\
     \arrayrulecolor{black!40}\midrule
     SiT-XL/2 & 675M & 400K   & 17.2\\
     {\textbf{+ \sname (ours)}}  & 675M & \textbf{150K}  & \textbf{13.6}\\
     \arrayrulecolor{black!40}\midrule
     SiT-XL/2 & 675M & 7M   & \pz8.3\\
     {\textbf{+ \sname (ours)}}  & 675M & \textbf{400K}  & \textbf{\pz7.9}\\
     {\textbf{+ \sname (ours)}}  & 675M & \textbf{1M}  & \textbf{\pz6.4}\\
     {\textbf{+ \sname (ours)}}  & 675M & \textbf{4M}  & \textbf{\pz5.9}\\
 \arrayrulecolor{black}\bottomrule
\end{tabular}
\label{tab:wo_cfg}
}
\end{minipage}
~~
\begin{minipage}{0.55\textwidth}

\centering\large
\captionof{table}{
\textbf{System-level comparison} on ImageNet 256$\times$256 with CFG. $\downarrow$ and $\uparrow$ indicate whether lower or higher values are better, respectively. Results that include additional CFG scheduling are marked with an asterisk (*), where the guidance interval from~\citep{kynkaanniemi2024applying} is applied for REPA.}
\vspace{-0.06in}
\resizebox{\textwidth}{!}{%
\begin{tabular}{l c c c c c c}
\toprule
{\pz\pz Model} & Epochs  &  {\pz FID$\downarrow$} & {sFID$\downarrow$} & {IS$\uparrow$} & {Pre.$\uparrow$} & Rec.$\uparrow$ \\
\arrayrulecolor{black}\midrule

\multicolumn{7}{l}{\emph{Pixel diffusion}\vspace{0.02in}} \\
\pz\pz {ADM-U}  &\pz400 &  \pz3.94 & 6.14 &  186.7 & 0.82 & 0.52 \\
\pz\pz VDM$++$ & \pz560 & \pz2.40 & - &  225.3 & - & - \\
\pz\pz Simple diffusion & \pz800 & \pz2.77 & - & 211.8 & - & - \\
\pz\pz CDM & 2160 & \pz4.88 & - & 158.7 & - & - \\
\arrayrulecolor{black!40}\midrule

\multicolumn{7}{l}{\emph{Latent diffusion, U-Net}\vspace{0.02in}} \\
\pz\pz LDM-4 & \pz200  & \pz3.60 & - & 247.7 & {0.87} & 0.48 \\
\arrayrulecolor{black!40}\midrule

\multicolumn{7}{l}{\emph{Latent diffusion, Transformer + U-Net hybrid}\vspace{0.02in}} \\
\pz\pz U-ViT-H/2 & \pz240 & \pz2.29 & 5.68  & 263.9 & 0.82 & 0.57 \\ 
\pz\pz DiffiT* & - & \pz1.73 & - &  276.5 & 0.80 & 0.62 \\
\pz\pz MDTv2-XL/2* & 1080  &  \pz1.58 & 4.52 & 314.7  & 0.79 & {0.65}\\
\arrayrulecolor{black}\midrule

\multicolumn{7}{l}{\emph{Latent diffusion, Transformer}\vspace{0.02in}} \\
\pz\pz MaskDiT & 1600 &  \pz2.28 & 5.67 & 276.6 & 0.80 & 0.61 \\ 
\pz\pz SD-DiT & \pz480 & \pz3.23 & -    & -     & -    & -     \\
\arrayrulecolor{black!30}\cmidrule(lr){1-7}
\pz\pz {DiT-XL/2}   & 1400  &    \pz2.27 & 4.60 & {278.2} & {\textbf{0.83}} & 0.57  \\
\arrayrulecolor{black!30}\cmidrule(lr){1-7}
\pz\pz {SiT-XL/2}   & 1400 &     \pz2.06 & {4.50} & 270.3 & 0.82 & 0.59 \\
{\pz\pz{+ \sname (ours)}} & {\pz200} & \pz{1.96} & \textbf{4.49} & 264.0 & 0.82 & 0.60 \\
{\pz\pz{+ \sname (ours)}} & {\pz800} & {\pz1.80} & {{4.50}} & {{284.0}} & {0.81} & {0.61} \\
{\textbf{\pz\pz{+ \sname (ours)*}}} & {\textbf{\pz800}} & \textbf{\pz1.42} & {4.70} & {\textbf{305.7}} & {0.80} & {\textbf{0.65}} \\
\arrayrulecolor{black}\bottomrule
\end{tabular}
}
\label{tab:main}
\end{minipage}
\vspace{-0.15in}
\end{figure}

\subsection{System-level comparison}
Based on the analysis, we perform a system-level comparison between recent state-of-the-art diffusion model approaches and diffusion transformers with \sname.
First, we compare the FID values between vanilla DiT or SiT models and the same models trained with \sname. As shown in Table~\ref{tab:wo_cfg}, \sname shows consistent and significant improvement across all model variants. In particular, on SiT-XL/2, aligning representation leads to FID$=$7.9 at 400K iteration, which already exceeds the FID of the vanilla SiT-XL at 7M iteration. Note that the performance continues to improve with longer training; for instance, with SiT-XL/2, FID becomes 6.4 at 1M iteration and 5.9 at 4M iteration. We also qualitatively compare the progression of generation results in Figure~\ref{fig:qual_progression}, where we use the same initial noise across different models. The model trained with \sname exhibits better progression.

Finally, we provide a quantitative comparison between SiT-XL/2 with \sname and other recent diffusion model methods using classifier-free guidance~\citep{ho2022classifier}. Our method already outperforms the original SiT-XL/2 with 7$\times$ fewer epochs and it is further improved with longer training. At 800 epochs, SiT-XL/2 with \sname achieves FID of \bestfid with a classifier-free guidance scale of $w=1.35$, and achieves state-of-the-art FID of \bestfidinterval with a extra classifier-free guidance scheduling with guidance interval \citep{kynkaanniemi2024applying}.
We provide selected qualitative results of SiT-XL/2 with \sname in Figure~\ref{fig:main_qual} and more examples in Appendix~\ref{appen:more_qual}. Moreover, we provide experimental results on ImageNet 512$\times$512 and text-to-image generation in Appendix \ref{appen:512} and \ref{appen:t2i}; we show that \sname also provides significant improvements in such setups. 

\subsection{Ablation studies}
\textbf{Representation gap across different timesteps.}
We begin by comparing the semantic gap (measured through linear probing results) using outputs of the SiT models with different noise scale (\ie, different timesteps), and maximum CKNNA values using clean DINOv2-g representations. As shown in Figure~\ref{fig:abla_different_timestep}, \sname consistently reduces the representation gap across different noise levels, as indicated by better linear probing results and higher CKNNA values across all noise scales.

\textbf{Alignment to different visual encoders.}
In addition, we extend the analysis from Section~\ref{fig:sit_observation} to other visual encoders, not limited to the DINOv2 models. Specifically, we train SiT-L/2 models using \sname with MAE or MoCov3. As depicted in Figure~\ref{fig:abla_different_enc}, these models demonstrate higher CKNNA values across the corresponding target representations than the vanilla model. This indicates that \sname is effective in aligning various visual representations, not limited to DINOv2.

\begin{wrapfigure}[6]{r}{0.4\textwidth}
\centering\small
\captionof{table}{
Ablation study for $\lambda$.
}
\vspace{-0.1in}
\resizebox{0.38\textwidth}{!}{
\begin{tabular}{l c c c c}
     \toprule
     $\lambda$ & 0.25 & 0.5 & 0.75 & 1.0  \\
     \midrule
     FID$\downarrow$ & 8.6 & 7.9 & 7.8 & 7.8  \\
     IS$\uparrow$ & 118.6 & 122.6 & 124.4 & 124.8  \\
     \bottomrule
     & 
\end{tabular}
}
\label{tab:lambda}
\end{wrapfigure}
\textbf{Effect of $\lambda$}. We also examine the effect of the regularization coefficient $\lambda$ by training SiT-XL/2 models for 400K with different coefficients 0.25 to 1.0 and comparing the performance. As shown in Table~\ref{tab:lambda}, the performance is robust to the values and it is quite saturated after $\lambda=0.5$. 

\begin{figure*}[t!]
\vspace{-0.1in}
\centering\small
    \begin{minipage}[t]{0.48\textwidth}%
        \centering
        \begin{subfigure}[t]{0.49\linewidth}
            \includegraphics[width=.9\linewidth]{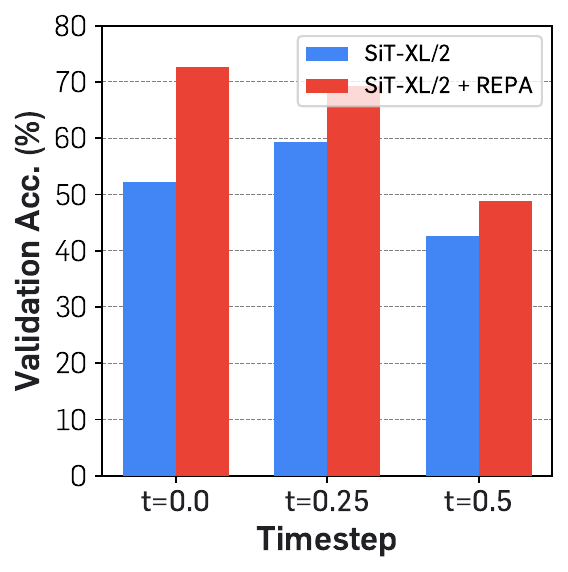}
            \caption{Linear probing}
        \end{subfigure}%
        \hfill
        \begin{subfigure}[t]{0.49\linewidth}
            \includegraphics[width=.9\linewidth]{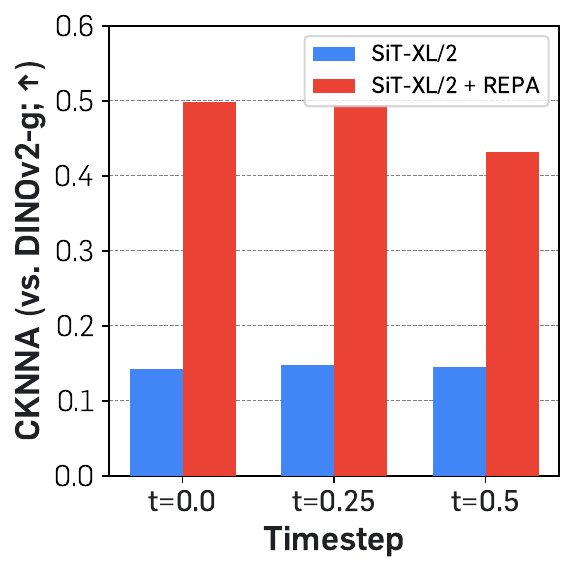}
            \caption{Alignment}
        \end{subfigure}
         \ifunderreview
\else
        \caption{\textbf{Representation gap across different timesteps.} We plot the linear probing results and maximum CKNNA values (using DINOv2-g) at different timesteps, comparing the vanilla SiT-XL/2 model and the same model trained using \sname. \sname consistently reduces the representation gap across different noise levels.}
        \fi        
    \label{fig:abla_different_timestep}
    \end{minipage}%
    ~~
    \begin{minipage}[t]{0.48\textwidth}
        \centering
        \begin{subfigure}[t]{0.49\linewidth}
            \includegraphics[width=.9\linewidth]{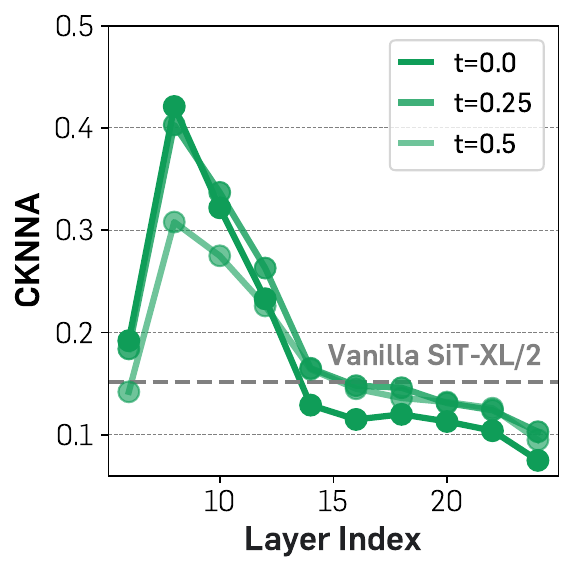}
            \caption{MoCov3-L}
        \end{subfigure}%
        \hfill
        \begin{subfigure}[t]{0.49\linewidth}
            \includegraphics[width=.9\linewidth]{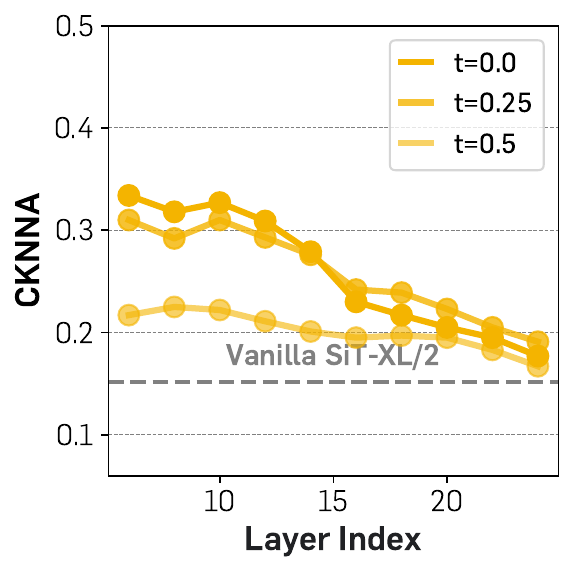}
            \caption{MAE-L}
        \end{subfigure}
        \caption{\textbf{Alignment to different target representations.} CKNNA values of SiT-L/2 models using \sname, with (a) MoCov3-L and (b) MAE-L as target representations. 
        After that, we measure CKNNA using these encoders. 
        \sname consistently improves the alignment regardless of the target representations.
        }
    \label{fig:abla_different_enc}
    \end{minipage}
    \vspace{-0.15in}
\end{figure*}


\vspace{-0.04in}
\section{Related Work}
\vspace{-0.04in}
\label{sec:related}

We discuss with the most relevant literature here and provide a more discussion in Appendix~\ref{appen:related}.

\textbf{Bridging diffusion models and representation learning.}
Many recent works have attempted to exploit or improve representations learned from diffusion models \citep{fuest2024diffusion}. First, there are {hybrid model approaches}: \citet{yang2022your} and \citet{deja2023learning} train a single model capable of both classification and diffusion-based generation. Also, \citet{tian2023addp} introduces a hybrid model capable of segmentation and generation. Next, several works have analyzed and exploited {representations in diffusion models}: \citet{xiang2023denoising} and \citet{mukhopadhyay2023diffusion} observe that the intermediate representations of diffusion models have discriminative properties. Moreover, Repfusion \citep{yang2023diffusion} and DreamTeacher \citep{li2023dreamteacher} propose knowledge distillation schemes using diffusion models to perform various downstream tasks.
Our work also shares some similarities, where we focus on designing a regularization for \emph{alignments} between recent self-supervised and diffusion representations and how they affect generation.
\nocite{zaidi2022pre}

\textbf{Diffusion models with external representations.}
Several recent studies have explored leveraging pretrained visual encoders to enhance efficiency and performance of diffusion models~\citep{pernias2024wrstchen, li2023return}. W\"urstchen \citep{pernias2024wrstchen} introduces a two-stage text-to-image diffusion model framework: a text-conditioned model that first generates a semantic map from a text prompt, followed by another diffusion model that synthesizes images based on the semantic map. RCG \citep{li2023return} focuses on unconditional generation, where a compact 1D latent vector is produced by a diffusion model and subsequently used as a label for image generation by a second diffusion model. We also exploit pretrained representations for improving the diffusion model but without the need of training an additional model that learns the representation distribution. 

\vspace{-0.04in}
\section{Conclusion}
\label{sec:conclusion}
\vspace{-0.04in}
In this paper, we have presented \sname, a simple regularization for improving diffusion transformers. In particular, we investigated whether diffusion transformer representations can be aligned with recent self-supervised representations, and if it can improve the generation performance of diffusion transformers. We showed \sname can significantly improves generation performance of diffusion transformers with faster convergence speed. We hope our work would facilitate many possible future research directions, including unifying discriminative and generative models and their representations or theoretical analysis; We provide more discussion in Appendix \ref{appen:limitation}.

\section*{Reproducibility Statement}
We provide hyperparameter details in Section~\ref{sec:exp} and Appendix~\ref{appen:main_setup}. We also release the implementation and model checkpoints to reproduce the results in the paper.

\section*{Acknowledgment}
This work was supported by Institute for Information \&
communications Technology Promotion(IITP) grant funded by the Korea government(MSIT) 
(No. RS-2019-II190075 Artificial Intelligence Graduate School Program (KAIST); No. RS-2024-00509279, Global AI Frontier Lab).

SY thanks the anonymous reviewers, Hankook Lee, Jihoon Tack, Sukmin Yun, and {Yisol Choi} for their insightful discussions; {Myungkyu Koo} and Daewon Choi for proofreading; and Kyungmin Lee for providing an excellent diffusion model implementation. SY also thanks Yewon Kim for the assistance in improving the visualizations in the paper.

JJ acknowledges support from the Institute of Information \& communications Technology Planning \& Evaluation (IITP) grant funded by the Korea government (MSIT) (No.~RS-2019-II190079, Artificial Intelligence Graduate School Program (Korea University)), and the IITP-ITRC (Information Technology Research Center) grant funded by the Korea government (MSIT) (No.~IITP-2025-RS-2024-00436857, 10\%).

SX thanks discussions with Boyang Zheng and Willis Ma. SX also acknowledges support from the IITP grant funded by the Korean Government (MSIT) (No. RS-2024-00457882, National AI Research Lab Project), the Open Path AI Foundation, Amazon Research Award, Google TRC program, and NSF Award IIS-2443404.

\bibliography{iclr2025_conference}
\bibliographystyle{iclr2025_conference}

\appendix
\clearpage
\section{Descriptions for Diffusion-based Models}
\label{appen:si}

We provide an overview of two types of generative models that we use in this paper, which learn the target distribution by training variants of a denoising autoencoder. We first explain denoising diffusion probabilistic models (DDPM) in Section~\ref{appen:subsec:ddpm} and stochastic interpolants in Section~\ref{appen:subsec:si}. For detailed explanations and rigorous proofs, please refer to the original papers \citep{albergo2023stochastic, ma2024sit} that provide excellent formulations and description.

\subsection{Denoising diffusion probabilistic models}
\label{appen:subsec:ddpm}
\emph{Diffusion models} \citep{sohl2015deep,ho2021denoising} model the target distribution $p(\rvx)$ via learning a gradual denoising process from Gaussian distribution $\mathcal{N}(\mathbf{0}, \mathbf{I})$ to $p(\rvx)$. Formally, diffusion models learn a \emph{reverse} process $p (\rvx_{t-1} | \rvx_{t})$ of the pre-defined \emph{forward} process $q(\rvx_t | \rvx_{0})$ that gradually adds the Gaussian noise starting from $p(\rvx)$ for $1\leq t \leq T$ with a fixed $T>0$.

For a given $\rvx_0 \sim p(\rvx)$, $q(\rvx_t | \rvx_{t-1})$ can be formalized as $q(\rvx_t| \rvx_{t-1}) \coloneqq \mathcal{N}(\rvx_t; \sqrt{1 - \beta_t}\rvx_0, \beta_t^2\mathbf{I})$, where $\beta_t \in (0,1)$ are pre-defined hyperparameters set to be small. In particular, DDPM~\citep{ho2021denoising} shows if one formalizes the reverse process $p (\rvx_{t-1} | \rvx_{t})$ (with ${\alpha}_t = 1 - \beta_t$. $\bar{\alpha}_t \coloneqq \prod_{i=1}^{t} \alpha_i$ for $1 \leq t \leq T$) as
\begin{align}
    p(\rvx_{t-1}|\rvx_{t}) \coloneqq \mathcal{N}\Big(\rvx_{t-1}; \frac{1}{\sqrt{\alpha_t}}\big(\rvx_t - \frac{\sigma_t^2}{\sqrt{1-\bar{\alpha}_t}}\bm{\epsilon}_{\bm{\theta}}(\rvx_t, t)\big), \mathbf{\Sigma}_\theta(\rvx_t, t)\Big),
\end{align}
then $\bm{\epsilon}_{\bm{\theta}}(\rvx_t, t)$ can be trained with a simple denoising autoencoder objective parameterized by $\bm{\theta}$:
\begin{align}
    \mathcal{L}_{\text{simple}} \coloneqq \mathbb{E}_{\rvx_\ast, \bm{\epsilon}, t} \Big[ ||\bm{\epsilon} - \bm{\epsilon}_{\bm{\theta}}(\rvx_t, t)||_2^2 \Big].
    \label{eq:ddpm}
\end{align}

For $\mathbf{\Sigma}_\theta(\rvx_t, t)$, \citep{ho2021denoising} shows it is enough to simply define it as $\sigma_t^2 \mathbf{I}$ with $\beta_t = \sigma_t^2$. After that, \citet{nichol2021improved} exhibits the performance can be improved if the model jointly learns $\mathbf{\Sigma}_\theta(\rvx_t, t)$ with $\bm{\epsilon}_{\bm{\theta}}(\rvx_t, t)$ in dimension-wise manner through the following objective:
\begin{equation}
     \mathcal{L}_{\text{vlb}}\coloneqq \exp (v \log \beta_t + (1-v) \log \tilde{\beta}_t),
\end{equation}
where $v$ denotes each component per dimension from the model output and $\tilde{\beta}_t = \frac{1 - \bar{\alpha}_{t-1}}{1 - \bar{\alpha}_{t}} \beta_t$.

With a sufficiently large $T$ and an appropriate scheduling of $\beta_t$, the distribution $p(\rvx_T)$ becomes almost an isotropic Gaussian distribution. Hence, one can generate a sample starting from a random noise and perform iterative reverse process $p(\rvx_{t-1}|\rvx_{t})$ to reach the data sample $\rvx_0$~\citep{ho2021denoising}.

\clearpage
\subsection{Stochastic interpolants}
\label{appen:subsec:si}
Different from DDPM, \emph{flow matching models}~\citep{esser2024scaling,lipman2022flow,liu2022flow} deal with the continuous time-dependent process with a data $\rvx_\ast \sim p(\rvx)$ and a Gaussian noise $\rvepsilon~\sim \mathcal{N}(\mathbf{0}, \mathbf{I})$ on $t \in [0, 1]$:
\begin{equation}
    \rvx_t = \alpha_t \rvx_0 + \sigma_t \rvepsilon,\quad \alpha_0 = \sigma_1 = 1,\,\, \alpha_1 = \sigma_0 = 0,
\end{equation}
where $\alpha_t$ and $\sigma_t$ are a decreasing and increasing function of $t$ (respectively). There exists a \emph{probability flow ordinary differential equation} (PF ODE) with a velocity field
\begin{equation}
\dot{\rvx}_t = \rvv (\rvx_t, t),
\end{equation}
where distribution of this ODE at $t$ is equal to the marginal $p_t(\rvx)$.

The velocity $\rvv (\rvx, t)$ is represented as the following sum of two conditional expectations
\begin{equation}
    \rvv (\rvx, t) = \mathbb{E}[\dot{\rvx}_t| \rvx_t = \rvx] = \dot{\alpha}_t \mathbb{E}[\rvx_\ast| \rvx_t = \rvx] + \dot{\sigma}_t\mathbb{E}[\rvepsilon| \rvx_t = \rvx], 
\end{equation}
which can be approximated with model $\rvv_{\theta}(\rvx_t, t)$ by minimizing the following training objective:
\begin{equation}
    \mathcal{L}_{\text{velocity}}(\theta)\coloneqq
    \mathbb{E}_{\rvx_\ast, \bm{\epsilon}, t}
    \Big[
    ||\rvv_{\theta}(\rvx_t, t) - \dot{\alpha}_t\rvx_\ast - \dot{\sigma}_t\rvepsilon||^2 
    \Big].
\end{equation}
Note that this also corresponds to the following reverse \emph{stochastic differential equation} (SDE):
\begin{equation}
    d\rvx_t = \rvv(\rvx_t, t)dt - \frac{1}{2}w_t \rvs(\rvx_t, t)dt + \sqrt{w_t} d\bar{\mathbf{w}}_t,
    \label{eq:sde_appendix}
\end{equation}
where the score $\rvs(\rvx_t, t)$ similarly becomes the conditional expectation 
\begin{equation}
    \rvs(\rvx_t, t) = -\frac{1}{\sigma_t}\mathbb{E}[\rvepsilon| \rvx_t = \rvx].
\end{equation}
Similar to $\rvv$, $\rvs$ can be approximated with a model $\rvs_{\theta}(\rvx, t)$ with the following objective:
\begin{equation}
    \mathcal{L}_{\text{score}}(\theta) \coloneqq 
    \mathbb{E}_{\rvx_\ast, \bm{\epsilon}, t}
    \Big[
    ||\sigma_t\rvs_{\theta}(\rvx_t, t) + \rvepsilon||^2 
    \Big].
\end{equation}
Here, since the score $\rvs(\rvx, t)$ can be directly computed using the velocity $\rvv(\rvx, t)$ for $t > 0$ as 
\begin{equation}
    \rvs(\rvx, t) = {\frac{1}{\sigma_t}} \cdot \frac{\alpha_t\rvv(\rvx, t) - \dot{\alpha}_t\rvx}{\dot{\alpha}_t\sigma_t - {\alpha}_t\dot{\sigma}_t},
\end{equation}
so it is enough to estimate only one of the two vectors.

\emph{Stochastic interpolants}~\citep{albergo2023stochastic} shows any $\alpha_t$ and $\sigma_t$ satisfy the three conditions
\begin{align*}
    1.~&\alpha_t^2 + \sigma_t^2 > 0,\,\, \forall t\in[0,1] \\
    2.~&\alpha_t~\text{and}~\sigma_t~\text{are differentiable},\,\, \forall t \in [0,1] \\
    3.~&\alpha_1 = \sigma_0 = 0,~\alpha_0 = \sigma_1 = 1,
\end{align*}
leads to a process that interpolates between $\rvx_0$ and $\rvepsilon$ without bias. Thus, one can use a simple interpolant by defining them as a simple function during training and inference, such as linear interpolants with $\alpha_t=1-t$ and $\sigma_t=t$ or variance-preserving (VP) interpolants with $\alpha_t=\cos(\frac{\pi}{2}t)$ and $\sigma_t=\cos(\frac{\pi}{2}t)$~\citep{ma2024sit}. One another advantage of stochastic interpolants is that the diffusion coefficient $w_t$ is independent in training any of a score or a velocity model. Thus, $w_t$ can be also explicitly chosen \emph{after training} when sampling with the reverse SDE. 

Note that existing score-based diffusion models, including DDPM~\citep{ho2021denoising}, can be similarly interpreted as an SDE formulation. In particular, their forward diffusion process can be interpreted as a pre-defined (discretized) forward SDEs that have an equilibrium distribution as $\mathcal{N} (\mathbf{0}, \mathbf{I})$ at $t \to \infty$, where the training is done on $[0,T]$ with sufficiently large $T$ (\eg, $T=1000$) that $p(\rvx_T)$ becomes almost isotropic Gaussian. Generation is done by solving the corresponding reverse SDE starting from a random Gaussian noise by assuming $\rvx_T \sim \mathcal{N}(\mathbf{0}, \mathbf{I})$, where $\alpha_t, \sigma_t$ and the diffusion coefficient $w_t$ is \emph{implicitly} chosen from the forward diffusion process, which might lead to over-complicated design space of score-based diffusion models \citep{karras2022edm}.

\clearpage
\section{Diffusion Transformer Architecture}
\label{appen:architecture_detail}
\begin{figure}[ht!]
    \centering
    \includegraphics[width=0.30\linewidth]{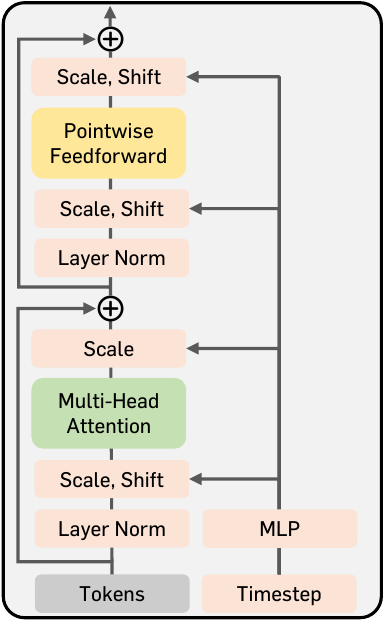}
    \caption{DiT block illustration.}
    \label{fig:dit_illstration}
\end{figure}

We strictly follow the architecture used in DiT~\citep{Peebles2022DiT} and SiT~\citep{ma2024sit}. The architecture is very similar to a vision transformer (ViTs; \citealt{dosovitskiy2021an}): an input is patchified, reshaped to a 1D sequence of patches with a length $N$, and then fed to the model. Similar to DiT and SiT, our architecture also uses a downsampled latent image $\rvz = E(\rvx)$ as an input, where $\rvx$ is a RGB image and $E$ is an encoder of the stable diffusion variational autoencoder (VAE) \citep{rombach2022high}. Different from the original ViT, our architecture also includes additional modulation layers at each attention block called AdaIN-zero layers. These layers scale and shift each hidden state with respect to the given timestep and additional conditions. We also consider a single multilayer perceptron (MLP) that projects a hidden state to the target representation space, which is only used in training. We provide an illustration of the DiT block in Figure~\ref{fig:dit_illstration}.

\section{Analysis Details}
\label{appen:analysis}

\subsection{Evaluation details}
\label{appen:subsec:metric}
\textbf{CKNNA} (Centered Kernel Nearest-Neighbor Alignment) is a \emph{relaxed version} of the popular Centered Kernel Alignment (CKA; \citealt{kornblith2019similarity}) that mitigates the strict definition of alignment. We generally follow the notations in the original paper for an explanation \citep{huh2024platonic}.

First, CKA have measured \emph{global} similarities of the models by considering all possible data pairs:
\begin{align}
    \mathrm{CKA}(\mathbf{K}, \mathbf{L}) = \frac{\mathrm{HSIC}(\mathbf{K}, \mathbf{L})}{\sqrt{\mathrm{HSIC}(\mathbf{K}, \mathbf{K})\mathrm{HSIC}(\mathbf{L}, \mathbf{L})}},
    \label{eq:cka}
\end{align}
where $\mathbf{K}$ and $\mathbf{L}$ are two kernel matrices computed from the dataset using two different networks. Specifically, it is defined as $\mathbf{K}_{ij} = \kappa (\phi_i, \phi_j)$ and $\mathbf{L}_{ij} = \kappa (\psi_i, \psi_j)$ where $\phi_i, \phi_j$ and $\psi_i, \psi_j$ are representations computed from each network at the corresponding data $\rvx_i, \rvx_j$ (respectively). By letting $\kappa$ as a inner product kernel, $\mathrm{HSIC}$ is defined as 
\begin{align}
    \mathrm{HSIC}(\mathbf{K}, \mathbf{L}) = \frac{1}{(n-1)^2}\Big(\sum_{i}\sum_j\big( \langle \phi_i, \phi_j \rangle -\mathbb{E}_{l}[ \langle \phi_i, \phi_l \rangle]\big)\big( \langle \psi_i, \psi_j \rangle -\mathbb{E}_{l}[ \langle \psi_i, \psi_l \rangle]\big) \Big).
    \label{eq:hsic}
\end{align}
CKNNA considers a relaxed version of Eq.~(\ref{eq:cka}) by replacing $\mathrm{HSIC}(\mathbf{K}, \mathbf{L})$ into $\mathrm{Align}(\mathbf{K}, \mathbf{L})$, where $\mathrm{Align}(\mathbf{K}, \mathbf{L})$ computes Eq.~(\ref{eq:hsic}) only using a $k$-nearest neighborhood embedding in the datasets:
\begin{align}
    \mathrm{Align}(\mathbf{K}, \mathbf{L}) = \frac{1}{(n-1)^2}\Big(\sum_{i}\sum_j \alpha(i, j)\big( \langle \phi_i, \phi_j \rangle -\mathbb{E}_{l}[ \langle \phi_i, \phi_l \rangle]\big)\big( \langle \psi_i, \psi_j \rangle -\mathbb{E}_{l}[ \langle \psi_i, \psi_l \rangle]\big) \Big),
    \label{eq:align}
\end{align}
where $\alpha(i, j)$ is defined as 
\begin{align}
    \alpha(i, j; k) = \mathbbm{1}[i \neq j\,\,\text{and}\,\, \phi_j \in \mathrm{knn} (\phi_i; k)\,\,\text{and}\,\,\psi_j \in \mathrm{knn} (\psi_i; k)],
\end{align}
so this term only considers $k$-nearest neighbors at each $i$. In this paper, we randomly sample 10,000 images in the validation set in ImageNet \citep{deng2009imagenet} and report CKNNA with $k=10$ based on observation in \citet{huh2024platonic} that smaller $k$ shows better a better alignment.

\textbf{Linear probing.}
We follow the setup used in DAE \citep{chen2024deconstructing}. Specifically, we use parameter-free batch normalization layer and train a linear layer for 90 epochs with a batch size of 16,384. We use the Adam optimizer~\citep{kingma2014adam} with cosine decay learning rate scheduler, where the initial learning rate is set to 0.001.

\subsection{DiT Analysis}
\label{appen:subsec:dit_analysis}
\begin{figure*}[ht!]
\centering\small
\begin{subfigure}{.3\textwidth}
\centering
\includegraphics[width=.96\textwidth]{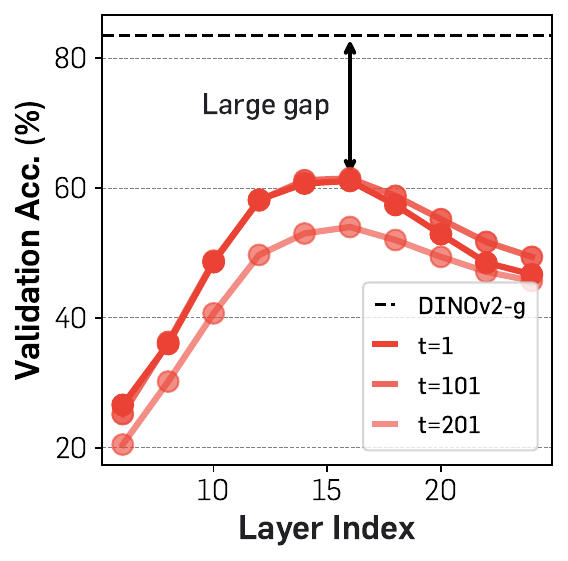}
\caption{Semantic gap} 
\label{subfig:dit_lin_eval}
\end{subfigure}
~~
\begin{subfigure}{.3\textwidth}
\centering
\includegraphics[width=1.\textwidth]{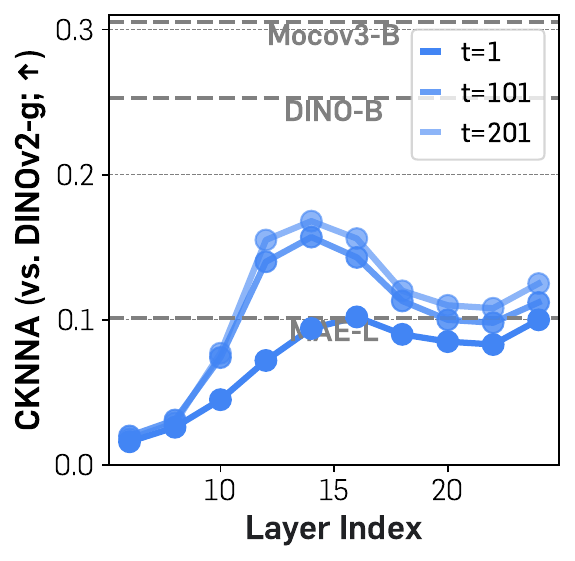}
\caption{Final alignment} 
\label{subfig:dit_cknna_dino}
\end{subfigure}
\caption{
\textbf{Empirical study with the pretrained DiT model.} Similar to Figure~\ref{fig:sit_observation}, we compare the semantic gap and measure the feature alignment between DINOv2-g and the DiT-XL/2 model trained with 7M iterations. (a) DiT learns meaningful (discriminative) representation but it still have a large gap between DINOv2. (b) Measured with CKNNA \citep{huh2024platonic}, DiT already shows a weak alignment with DINOv2, but its absolute value is still small. }
\label{fig:dit_observation}
\end{figure*}

We also perform a similar analysis have done in Figure~\ref{subfig:sit_lin_eval} (linear probing) and \ref{subfig:sit_cknna_dino} (CKA), and illustrate the result in Figure~\ref{fig:dit_observation}. Overall is shows a similar trend; the model includes discriminative representation but the gap is still large compared with DINOv2, as shown in the linear probing results, and also weakly aligned with DINOv2 representations. 

\subsection{Description of pretrained visual encoders}
\begin{itemize}[leftmargin=0.2in]
\item \textbf{MAE}~\citep{he2022masked} proposes a self-supervised representation learning objective for vision transformers, based on the reconstruction task of masked patches of input images.
\item \textbf{DINO}~\citep{caron2021emerging} is a self-supervised learning method based on self-distillation through the mean of momentum teacher network.
\item \textbf{MoCov3}~\citep{chen2021empirical} studies empirical study to train MoCo~\citep{he2020momentum, chen2020improved} on vision transformer and how they can be scaled up.
\item \textbf{CLIP}~\citep{radford2021learning} proposes a contrastive learning scheme on large image-text pairs.
\item \textbf{DINOv2}~\citep{oquab2024dinov} proposes a self-supervised learning method that combines pixel-level and patch-level discriminative objectives by leveraging advanced self-supervised techniques and a large pre-training dataset.
\item \textbf{I-JEPA}~\citep{assran2023self} predicts missing parts of an image by learning representations through joint-embedding, focusing on the context of the entire image without relying on pixel-level reconstruction.
\item \textbf{SigLIP}~\citep{zhai2023sigmoid} replaces the traditional softmax loss with a pairwise sigmoid loss, enhancing performance and efficiency of image-text representation learning.
\end{itemize}

Moreover, in Table~\ref{tab:dataset}, we also provide the datasets used for training of each of pretrained visual encoder. As shown in this table, better visual representations learned from massive amounts of image data provide more improvement, regardless of whether the dataset does not include ImageNet. Note that we do not fine-tune encoders (\eg, SigLIP and CLIP) with the ImageNet dataset, particularly if they were trained with another dataset, thereby separating the dataset leakage effect if we use these encoders. REPA also achieves significant improvements with these encoders, which validates that the improvement does not simply come from data leakages. 

\begin{table}[ht!]
\centering\small
\caption{Dataset analysis used for pretrained visual encoders.}
\begin{tabular}{llccc}
\toprule
    Method & Dataset  & w/ ImageNet-1K & Text sup. & FID$\downarrow$ \\
\midrule
     MAE    & ImageNet-1K & O   & X & 12.5 \\
     DINO   & ImageNet-1K & O   & X & 11.9 \\
     MoCov3 & ImageNet-1K & O   & X & 11.9 \\
     I-JEPA & ImageNet-1K & O   & X & 11.6 \\
     \midrule
     CLIP   & WIT-400M    & X   & O & 11.0 \\
     SigLIP & WebLi-4B    & X   & O & 10.2 \\
     DINOv2 & LVD-142M    & O   & X & 10.0 \\
\bottomrule
\end{tabular}
\label{tab:dataset}
\end{table}

\clearpage
\section{Hyperparameter and More Implementation Details}
\label{appen:main_setup}

\begin{table}[h!]
    \centering\small
    \caption{Hyperparameter setup.}
    \resizebox{\textwidth}{!}{%
    \begin{tabular}{l c c c c c}
        \toprule
         & {Figure 3} & Table 3 (SiT-B) & Table 3 (SiT-L) & Table 3 (SiT-XL) & Table 4 \\
        \midrule
        \textbf{Architecture} \\
        Input dim. & 32$\times$32$\times$4 & 32$\times$32$\times$4 & 32$\times$32$\times$4 & 32$\times$32$\times$4 & 32$\times$32$\times$4 \\
        Num. layers & 28 & 12 & 24 & 28 & 28 \\
        Hidden dim. & 1,152 & 768 & 1,024 & 1,152 & 1,152 \\
        Num. heads & 16 & 12 & 16 & 16 & 16 \\ 
        \midrule
        \textbf{\sname} \\
        $\lambda$ & 0.5 & 0.5 & 0.5 & 0.5 & 0.5 \\
        Alignment depth & 8 & 4 & 8 & 8 & 8 \\
        $\mathrm{sim}(\cdot, \cdot)$ & cos. sim.  & cos. sim. & NT-Xent & cos. sim.  & cos. sim.  \\
        Encoder $f(\rvx)$ & DINOv2-B  & DINOv2-B & DINOv2-L & DINOv2-B  & DINOv2-B\\
        \midrule
        \textbf{Optimization} \\
        Training iteration & 1M & 400K & 700K & 4M & 4M \\ 
        Batch size & 256 & 256 & 256 & 256 & 256 \\ 
        Optimizer & AdamW & AdamW & AdamW & AdamW & AdamW \\
        lr & 0.0001 & 0.0001 &  0.0001 & 0.0001 & 0.0001 \\
        $(\beta_1, \beta_2)$ & (0.9, 0.999) & (0.9, 0.999) & (0.9, 0.999) & (0.9, 0.999) & (0.9, 0.999) \\
        \midrule
        \textbf{Interpolants} \\
        $\alpha_t$ & $1-t$ & $1-t$ & $1-t$ & $1-t$ & $1-t$ \\
        $\sigma_t$ & $t$ & $t$ & $t$ & $t$ & $t$ \\
        $w_t$ & $\sigma_t$ & $\sigma_t$ & $\sigma_t$ & $\sigma_t$ & $\sigma_t$ \\
        Training objective & v-prediction & v-prediction & v-prediction & v-prediction & v-prediction \\
        Sampler & Euler-Maruyama & Euler-Maruyama & Euler-Maruyama & Euler-Maruyama & Euler-Maruyama \\
        Sampling steps & 250 & 250 & 250 & 250 & 250 \\
        Guidance & - & - & - & - &  1.35 \\
        \bottomrule
    \end{tabular}
    }
    \label{tab:hyperparam}
\end{table}

\textbf{Further implementation details.}
We implement our model based on the original SiT implementation~\citep{ma2024sit}. Throughout the experiments, we use the exact same structure as DiT~\citep{Peebles2022DiT} and SiT~\citep{ma2024sit}. We use AdamW~\citep{kingma2014adam, loshchilov2017decoupled} with constant learning rate of 1e-4, $(\beta_1, \beta_2)=(0.9, 0.999)$ without weight decay. To speed up training, we use mixed-precision (fp16) with a gradient clipping. We also pre-compute compressed latent vectors from raw pixels via stable diffusion VAE \citep{rombach2022high} and use these latent vectors. Because of this, we do not apply any data augmentation, but we find this does not lead to a big difference, as similarly observed in EDM2 \citep{karras2024analyzing}. We also use \texttt{stabilityai/sd-vae-ft-ema} decoder for decoding latent vectors to images. For MLP used for a projection, we use three-layer MLP with SiLU activations \citep{elfwing2018sigmoid}. We provide a detailed hyperparameter setup in Table~\ref{tab:hyperparam}. 

\textbf{Pretrained encoders.}
For MoCov3-B and -L models, we use the checkpoint in the implementation of RCG \citep{li2023return};\footnote{\url{https://github.com/LTH14/rcg}} for other checkpoints, we use their official checkpoints released in their official implementations. To adjust a different number of patches between the diffusion transformer and the pretrained encoder, we interpolate positional embeddings of pretrained encoders. 

\textbf{Sampler.}
For sampling, we use the Euler-Maruyama sampler with the SDE in Eq.~(\ref{eq:sde}) with a diffusion coefficient $w_t = \sigma_t$. We use the last step of the SDE sampler as 0.04, and it gives a significant improvement, similar to the original SiT paper \citep{ma2024sit}. 

\textbf{Computing resources.}
We use 8 NVIDIA H100 80GB GPUs for experiments; our training speed is about 5.4 step/s with a batch size of 256. Note that this can be further boosted with additional engineering (\eg, pre-computation of pretrained encoder features). 

\clearpage
\section{Evaluation Details}
\label{appen:eval}
We strictly follow the setup and use the same reference batches of ADM \citep{dhariwal2021diffusion} for evaluation, following their official implementation.\footnote{\url{https://github.com/openai/guided-diffusion/tree/main/evaluations}} We use NVIDIA H100 80GB GPUs or 4090Ti GPUs for evaluation and enable tf32 precision for faster generation, and we find the performance difference is negligible to the original fp32 precision.

In what follows, we explain the main concept of metrics that we used for the evaluation. 
\begin{itemize}[leftmargin=0.2in]
\item \textbf{FID}~\citep{heusel2017gans} measures the feature distance between the distributions of real and generated images. It uses the Inception-v3 network \citep{szegedy2016rethinking} and computes distance based on an assumption that both feature distributions are multivariate gaussian distributions.
\item \textbf{sFID}~\citep{nash2021generating} proposes to compute FID with intermediate spatial features of the Inception-v3 network to capture the generated images' spatial distribution.
\item \textbf{IS}~\citep{salimans2016improved} also uses the Inception-v3 network but use logit for evaluation of the metric. Specifically, it measures a KL-divergence between the original label distribution and the distribution of logits after the softmax normalization.
\item \textbf{Precision and recall}~\citep{kynkaanniemi2019improved} are based on their classic definitions: the fraction of realistic images and the fraction of training data manifold covered by generated data.
\end{itemize}
\section{Baselines}
\label{appen:baselines}
In what follows, we explain the main idea of baseline methods that we used for the evaluation.
\begin{itemize}[leftmargin=0.2in]
\item \textbf{ADM}~\citep{dhariwal2021diffusion} improves U-Net-based architectures for diffusion models and proposes classifier-guided sampling to balance the quality and diversity tradeoff.
\item \textbf{VDM++}~\citep{kingma2024understanding} proposes a simple adaptive noise schedule for diffusion models to improve training efficiency.
\item \textbf{Simple diffusion}~\citep{hoogeboom2023simple} proposes a diffusion model for high-resolution image generation by exploring various techniques to simplify a noise schedule and architectures.
\item \textbf{CDM}~\citep{ho2022cascaded} introduces cascaded diffusion models: similar to progressiveGAN \citep{karras2018progressive}, it trains multiple diffusion models starting from the lowest resolution and applying one or more super-resolution diffusion models for generating high-fidelity images.
\item \textbf{LDM}~\citep{rombach2022high} proposes latent diffusion models by modeling image distribution in a compressed latent space to improve the training efficiency without sacrificing the generation performance.
\item \textbf{U-ViT}~\citep{bao2022all} proposes a ViT-based latent diffusion model that incorporates U-Net-like long skip connections.
\item \textbf{DiffiT}~\citep{hatamizadeh2023diffit} proposes a time-dependent multi-head self-attention mechanism for enhancing the efficiency of transformer-based image diffusion models.
\item \textbf{MDTv2}~\citep{gao2023mdtv2} proposes an asymmetric encoder-decoder scheme for efficient training of a diffusion-based transformer. They also apply U-Net-like long-shortcuts in the encoder and dense input-shortcuts in the decoder.
\item \textbf{MaskDiT}~\citep{zheng2024fast} proposes an asymmetric encoder-decoder scheme for efficient training of diffusion transformers, where they train the model with an auxiliary mask reconstruction task similar to MAE \citep{he2022masked}.
\item \textbf{SD-DiT}~\citep{zhu2024sd} extends MaskdiT architecture but incorporates self-supervised discrimination objective using a momentum encoder.
\item \textbf{DiT}~\citep{Peebles2022DiT} proposes a pure transformer backbone for training diffusion models based on proposing AdaIN-zero modules.
\item \textbf{SiT}~\citep{ma2024sit} extensively analyzes how DiT training can be efficient by moving from discrete diffusion to continuous flow matching.
\end{itemize}

\clearpage
\section{Detailed Quantitative Results}
We provide evaluation results of different SiT models trained with \sname. All models are aligned with DINOv2-B representations with $\lambda=0.5$ and negative cosine similarity. We use the 4th layer hidden states for the base model and use the 8th layer hidden states for the large and xlarge model.
\begin{table}[ht!]
\centering\small
\caption{Detailed evaluation results with different model sizes. All results are reported without classifier-free guidance.}
\begin{tabular}{lcccccccc}
\toprule
     Model & \#Params & Iter. & FID$\downarrow$ & sFID$\downarrow$ & IS$\uparrow$ & Prec.$\uparrow$ & Rec.$\uparrow$ & Acc.$\uparrow$ \\
     \midrule
     \rowcolor{gray!15}
     SiT-B/2 \citep{ma2024sit} & 130M & 400K 
     & 33.0 & \pz6.46 & \pz43.7 & 0.53 & 0.63 & N/A \\
     {{+ \sname (ours)}} & 130M & {50K} 
     & 78.2 & 11.71 & \pz17.1 & 0.33 & 0.48 & 43.2 \\
     {{+ \sname (ours)}} & 130M & {100K} 
     & 49.5 & \pz7.00 & \pz27.5 & 0.46 & 0.59 & 50.9 \\
     {{+ \sname (ours)}} & 130M & {200K}
     & 33.2 & \pz6.68 & \pz43.7 & 0.54 & 0.63 & 50.9 \\
     {{+ \sname (ours)}} & 130M & {400K} 
     & 24.4 & \pz6.40 & \pz59.9 & 0.59 & 0.65 & 61.2\\
     \midrule
     \rowcolor{gray!15}
     SiT-L/2 \citep{ma2024sit} & 458M & 400K 
     & 18.8 & \pz5.29 & \pz72.0 & 0.64 & 0.64 & N/A \\
     {{+ \sname (ours)}} & 458M & {50K} 
     & 55.4 & \pz24.0 & \pz23.0 & 0.43 & 0.53 & 55.3 \\
     {{+ \sname (ours)}} & 458M & {100K} 
     & 24.1 & \pz6.25 & \pz55.7 & 0.62 & 0.60 & 61.8 \\
     {{+ \sname (ours)}} & 458M & {200K} 
     & 14.0 & \pz5.18  & \pz86.5 &  0.67 & 0.64 & 66.3\\
     {{+ \sname (ours)}} & 458M & {400K} 
     & 10.0 & \pz5.20 & 109.2 & 0.69 & 0.65 & 69.4\\
     \arrayrulecolor{black}\midrule
     \rowcolor{gray!15}
     SiT-XL/2 \citep{ma2024sit} & 675M & 7M   & \pz8.3 & \pz6.32 & 131.7 & 0.68 & 0.67 & N/A \\
     {{+ \sname (ours)}} & 675M & {50K} 
     & 52.3 & 31.24 & \pz24.3 & 0.45 & 0.53 & 56.1\\
     {{+ \sname (ours)}} & 675M & {100K} 
     & 19.4 & \pz6.06 & \pz67.4 & 0.64 & 0.61 & 62.9\\
     {{+ \sname (ours)}} & 675M & {200K} 
     & 11.1 & \pz5.05 & 100.4 & 0.69 & 0.64 & 67.3 \\
     {{+ \sname (ours)}} & 675M & {400K} 
     & \pz7.9 & \pz5.06 & 122.6 & 0.70 & 0.65 & 70.3 \\
     {{+ \sname (ours)}} & 675M & {4M} 
     & \pz5.9 & \pz5.73 & 157.8 & 0.70 & 0.69 & 74.6 \\
\bottomrule
\end{tabular}
\label{tab:detailed_quantitative}
\end{table}

We also provide SiT-XL/2+\sname at 4M iteration with classifier-free guidance with different class-free guidance scales.
\begin{table}[ht!]
\centering\small
\caption{Detailed evaluation results of SiT-XL+\sname at 4M iteration with different classifier-free guidance scale $w$.}
\begin{tabular}{lccccccccc}
\toprule
     Model & \#Params & Iter. & $w$ & FID$\downarrow$ & sFID$\downarrow$ & IS$\uparrow$ & Prec.$\uparrow$ & Rec.$\uparrow$  \\
     \midrule
     \rowcolor{gray!15}
     SiT-XL/2 \citep{ma2024sit} & 675M  & 7M  & 1.500 & 2.06 & 4.50 & 270.3 & 0.82 & 0.59  \\
     {{+ \sname (ours)}} & 675M & {4M} & 1.300 & 1.80 & 4.55 & 268.6 & 0.80 & 0.63 \\
     {{+ \sname (ours)}} & 675M & {4M} & 1.325 & 1.79 & 4.51 & 276.8 & 0.81 & 0.62 \\
     {{+ \sname (ours)}} & 675M & {4M} & 1.350 & 1.80 & 4.50 & 284.0 & 0.81 & 0.61 \\
     {{+ \sname (ours)}} & 675M & {4M} & 1.375 & 1.84 & 4.48 & 291.7 & 0.82 & 0.61 \\
     {{+ \sname (ours)}} & 675M & {4M} & 1.400 & 1.90 & 4.48 & 297.5 & 0.82 & 0.60 \\
     \bottomrule
\end{tabular}
\label{tab:detailed_quantitative_cfg}
\end{table}

Moreover, we provide the results with the guidance interval \citep{kynkaanniemi2024applying}.

\begin{table}[ht!]
\centering\small
\resizebox{\textwidth}{!}{%
\begin{tabular}{lccccccccc}
\toprule
     Model & \#Params & Iter. & Interval & $w$ & FID$\downarrow$ & sFID$\downarrow$ & IS$\uparrow$ & Prec.$\uparrow$ & Rec.$\uparrow$  \\
     \midrule
     \rowcolor{gray!15}
     SiT-XL/2 \citep{ma2024sit} & 675M  & 7M &[0, 1] & 1.50 & 2.06 & 4.50 & 270.3 & 0.82 & 0.59  \\
     {{+ \sname (ours)}} & 675M & {4M} & [0, 0.8] & 2.00 & 2.23 & 4.40 & 360.9 & 0.84 & 0.6 \\
     {{+ \sname (ours)}} & 675M & {4M} & [0, 0.75] &2.00 & 1.78 & 4.50 & 346.2 & 0.82 & 0.62 \\
     {{+ \sname (ours)}} & 675M & {4M} & [0, 0.7] &2.00 & 1.48 & 4.67 & 324.0 & 0.82 & 0.62 \\
     {{+ \sname (ours)}} & 675M & {4M} & [0, 0.65] &2.00 & 1.44 & 4.88 & 308.8 & 0.79 & 0.65 \\
     {{+ \sname (ours)}} & 675M & {4M} & [0, 0.6] & 2.00 & 1.56 & 5.11 & 290.7 & 0.78 & 0.66 \\
     {{+ \sname (ours)}} & 675M & {4M} & [0, 0.7] &1.90 & 1.45 & 4.68 & 317.6 & 0.80 & 0.64 \\
     {{+ \sname (ours)}} & 675M & {4M} & [0, 0.7] &1.80 & 1.42 & 4.70 & 305.7 & 0.80 & 0.64 \\
     \bottomrule
\end{tabular}
}
\caption{Detailed evaluation results of SiT-XL+\sname at 4M iteration with different classifier-free guidance scale $w$. We apply the guidance interval \citep{kynkaanniemi2024applying}.}
\label{tab:detailed_quantitative_cfg_interval}
\end{table}

\begin{figure}[h!]
    \centering
    \includegraphics[width=.46\linewidth]{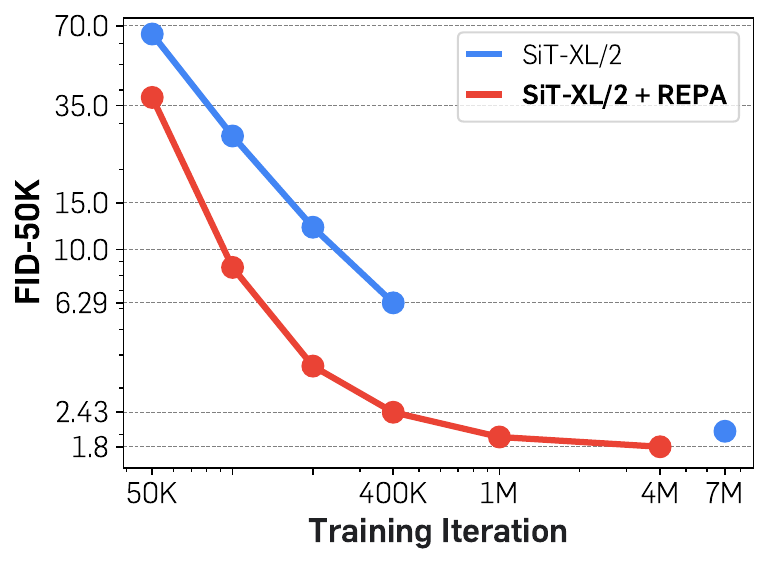}
    \caption{\textbf{Training iteration vs. FID plot.} All values are measured using a classifier-free guidance. \sname demonstrates a notable speedup and enhanced performance.}
    \label{fig:teaser_cfg}
\end{figure}

\clearpage
Finally, in Figure~\ref{fig:teaser_cfg}, we provide a plot similar to Figure~\ref{fig:teaser} using SiT-XL/2 but FIDs are obtained with a classifier-guidance scale $w=1.35$. Similar to Figure~\ref{fig:teaser}, \sname provides great speedup and performance improvement compared with the vanilla model. 

\clearpage
\section{More Qualitative Results}
\label{appen:more_qual}
\begin{figure}[ht!]
    \centering
    \includegraphics[width=\linewidth]{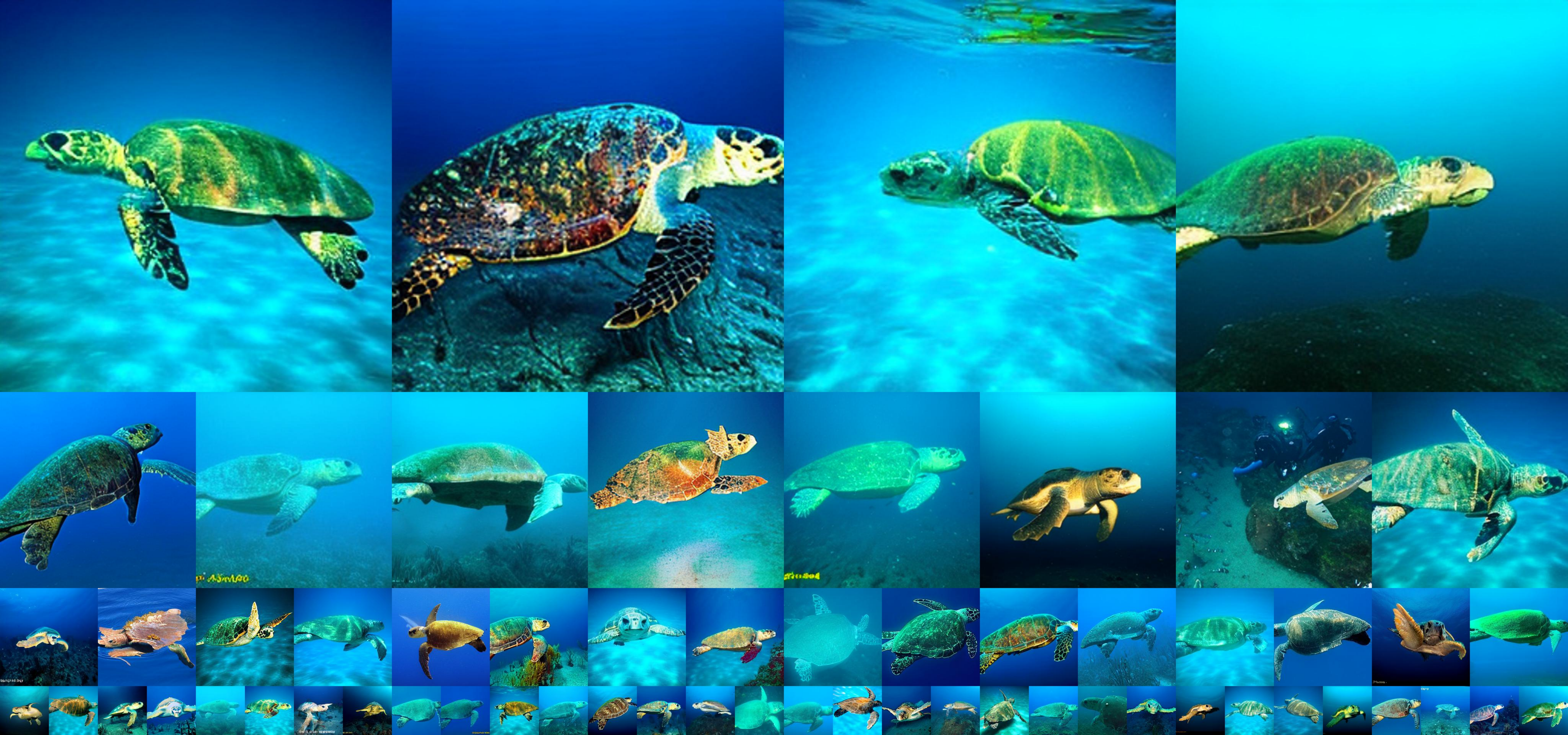}
    \caption{\textbf{Uncurated generation results of SiT-XL/2 + \sname.} We use classifier-free guidance with $w=4.0$. Class label = ``loggerhead sea turtle'' (33).}
\end{figure}
\begin{figure}[ht!]
    \centering
    \includegraphics[width=\linewidth]{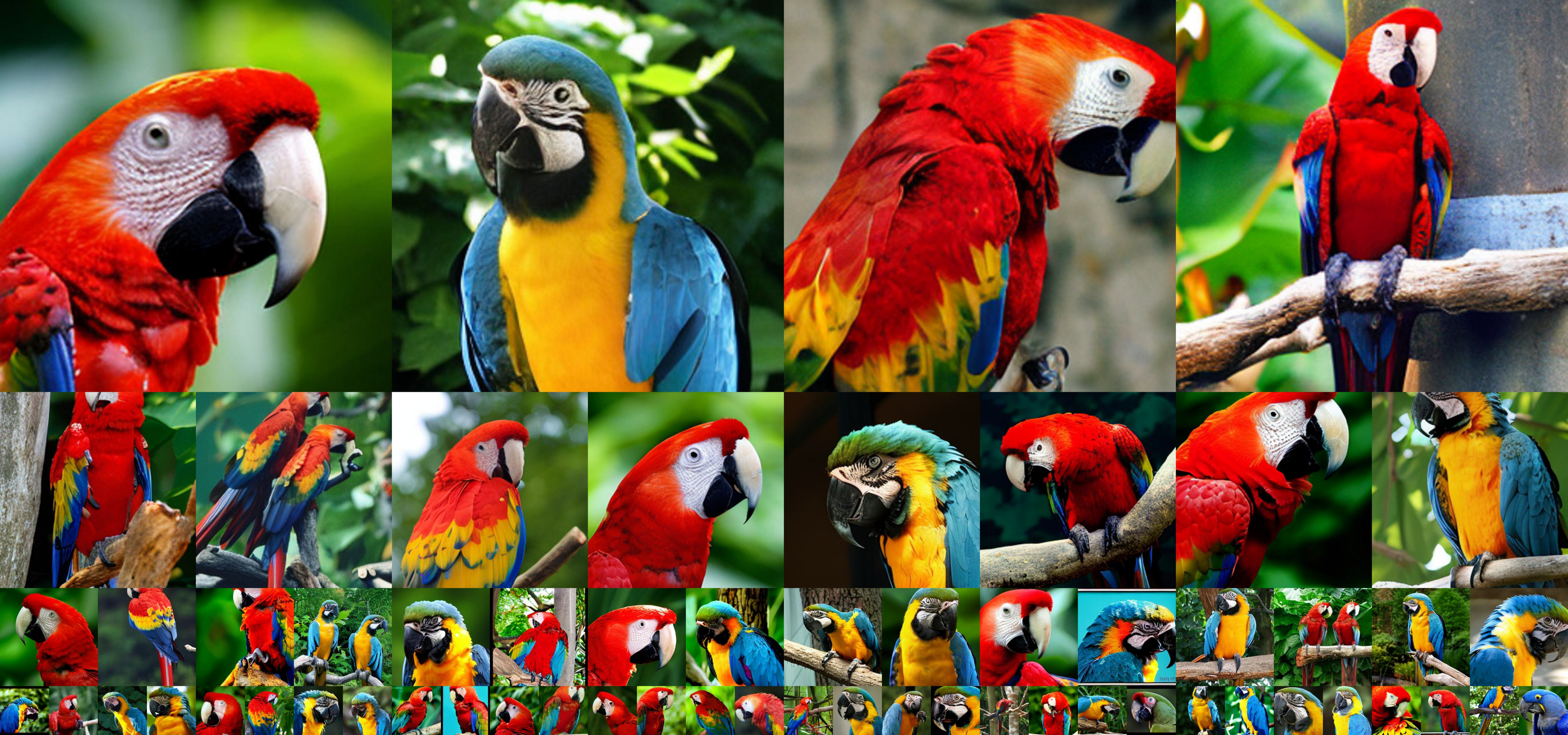}
    \caption{\textbf{Uncurated generation results of SiT-XL/2 + \sname.} We use classifier-free guidance with $w=4.0$. Class label = ``macaw'' (88).}
\end{figure}
\clearpage
\begin{figure}[ht!]
    \centering
    \includegraphics[width=\linewidth]{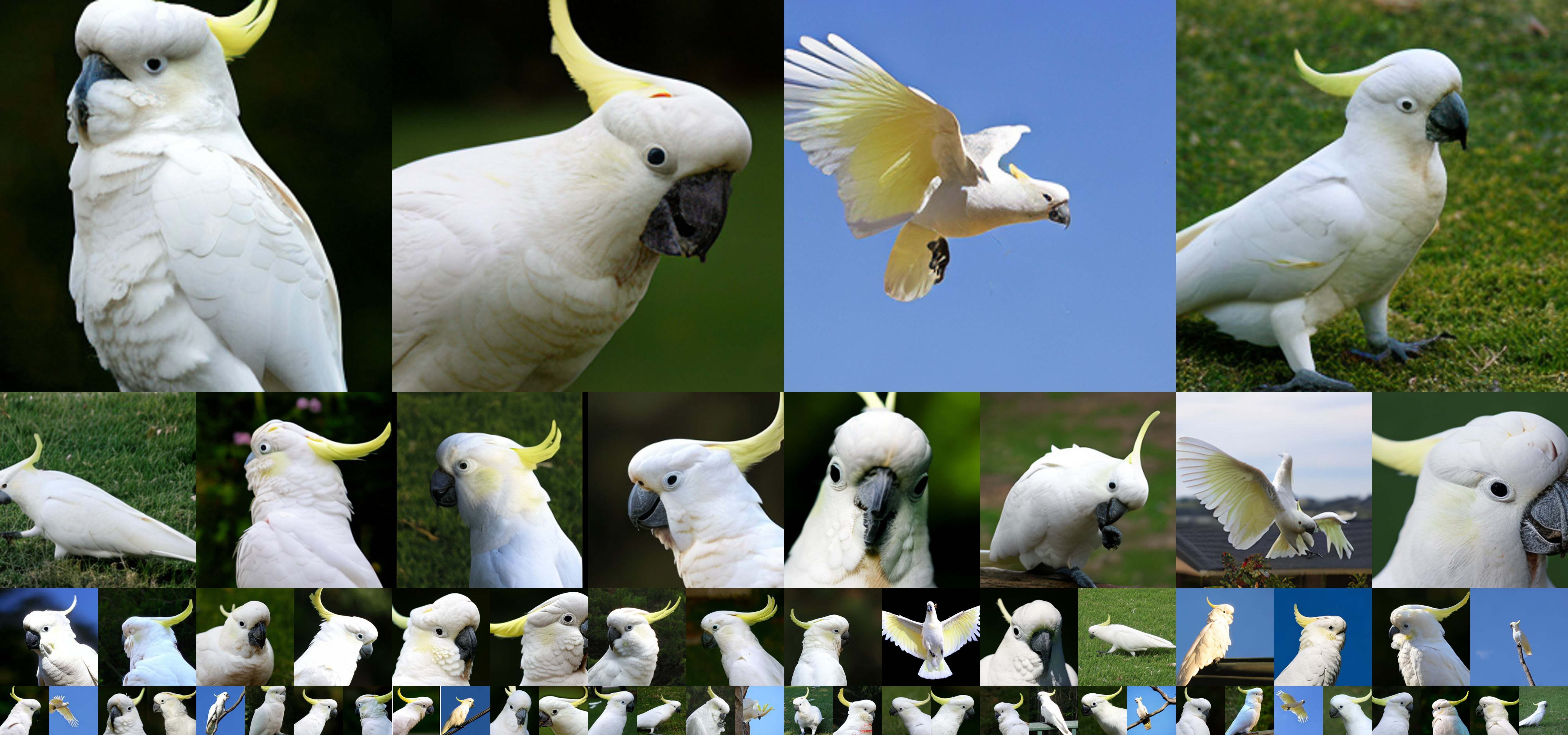}
    \caption{\textbf{Uncurated generation results of SiT-XL/2 + \sname.} We use classifier-free guidance with $w=4.0$. Class label = ``sulphur-crested cockatoo'' (89).}
\end{figure}
\begin{figure}[ht!]
    \centering
    \includegraphics[width=\linewidth]{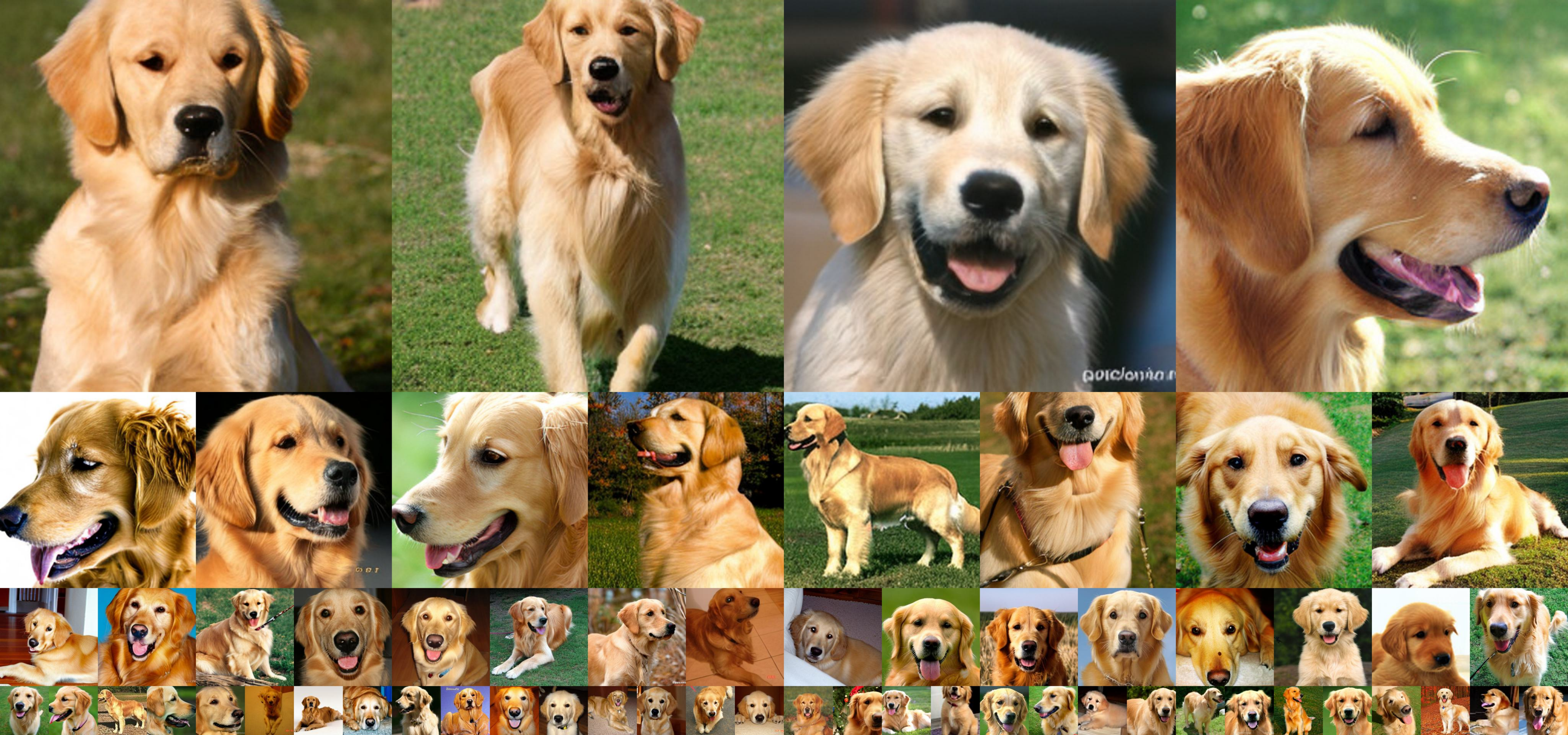}
    \caption{\textbf{Uncurated generation results of SiT-XL/2 + \sname.} We use classifier-free guidance with $w=4.0$. Class label = ``golden retriever'' (207).}
\end{figure}
\clearpage
\begin{figure}[ht!]
    \centering
    \includegraphics[width=\linewidth]{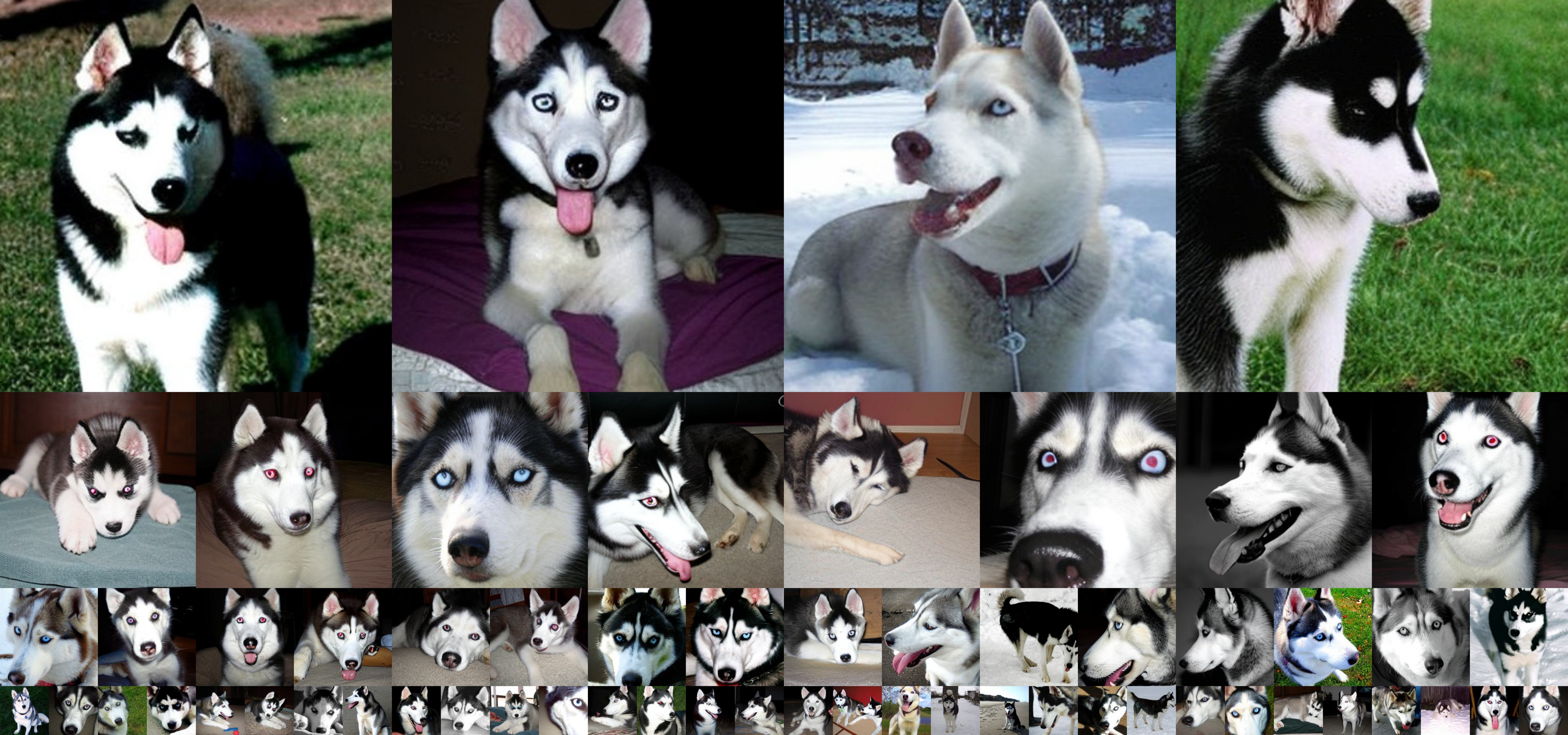}
    \caption{\textbf{Uncurated generation results of SiT-XL/2 + \sname.} We use classifier-free guidance with $w=4.0$. Class label = ``husky'' (250).}
\end{figure}
\begin{figure}[ht!]
    \centering
    \includegraphics[width=\linewidth]{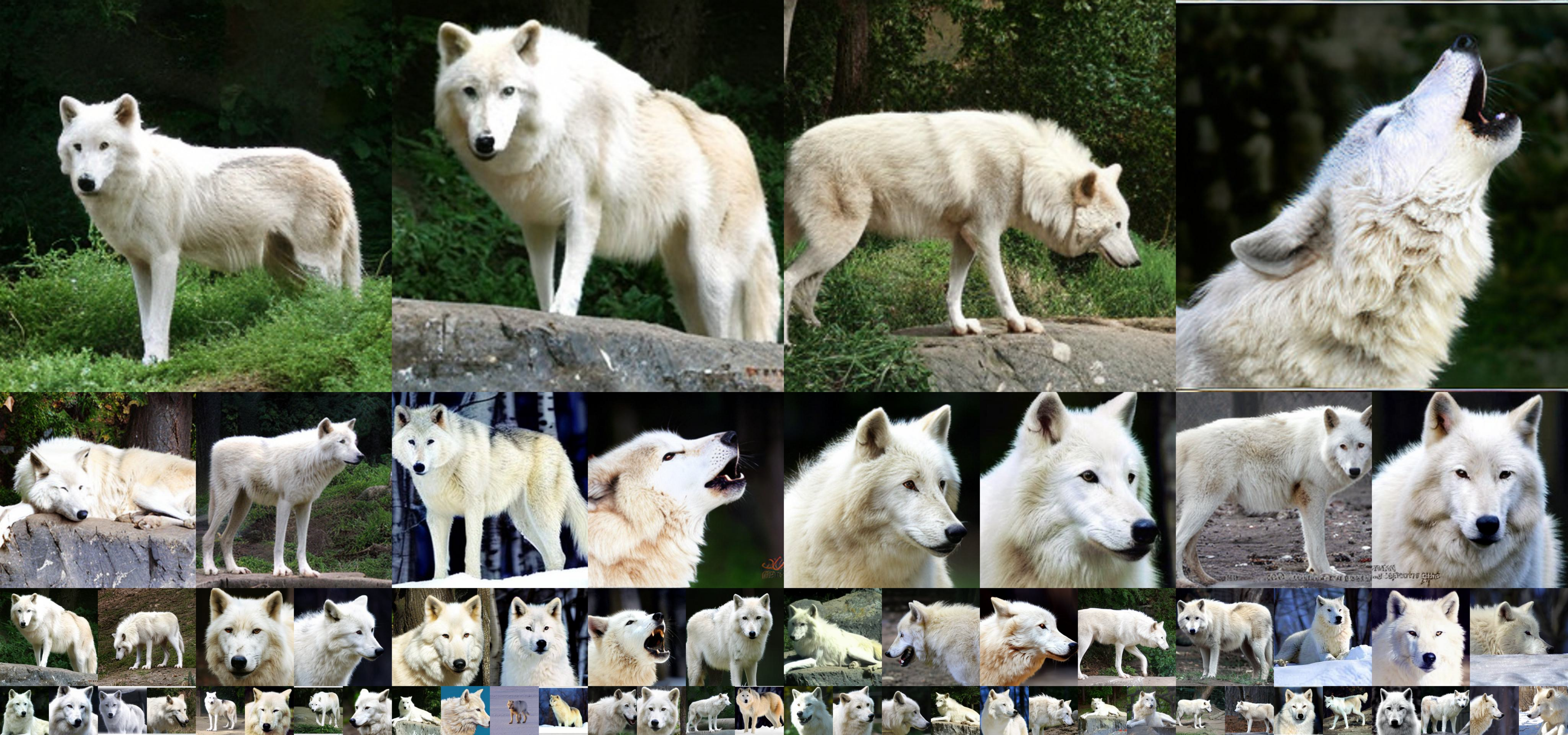}
    \caption{\textbf{Uncurated generation results of SiT-XL/2 + \sname.} We use classifier-free guidance with $w=4.0$. Class label = ``arctic wolf'' (270).}
\end{figure}
\clearpage
\begin{figure}[ht!]
    \centering
    \includegraphics[width=\linewidth]{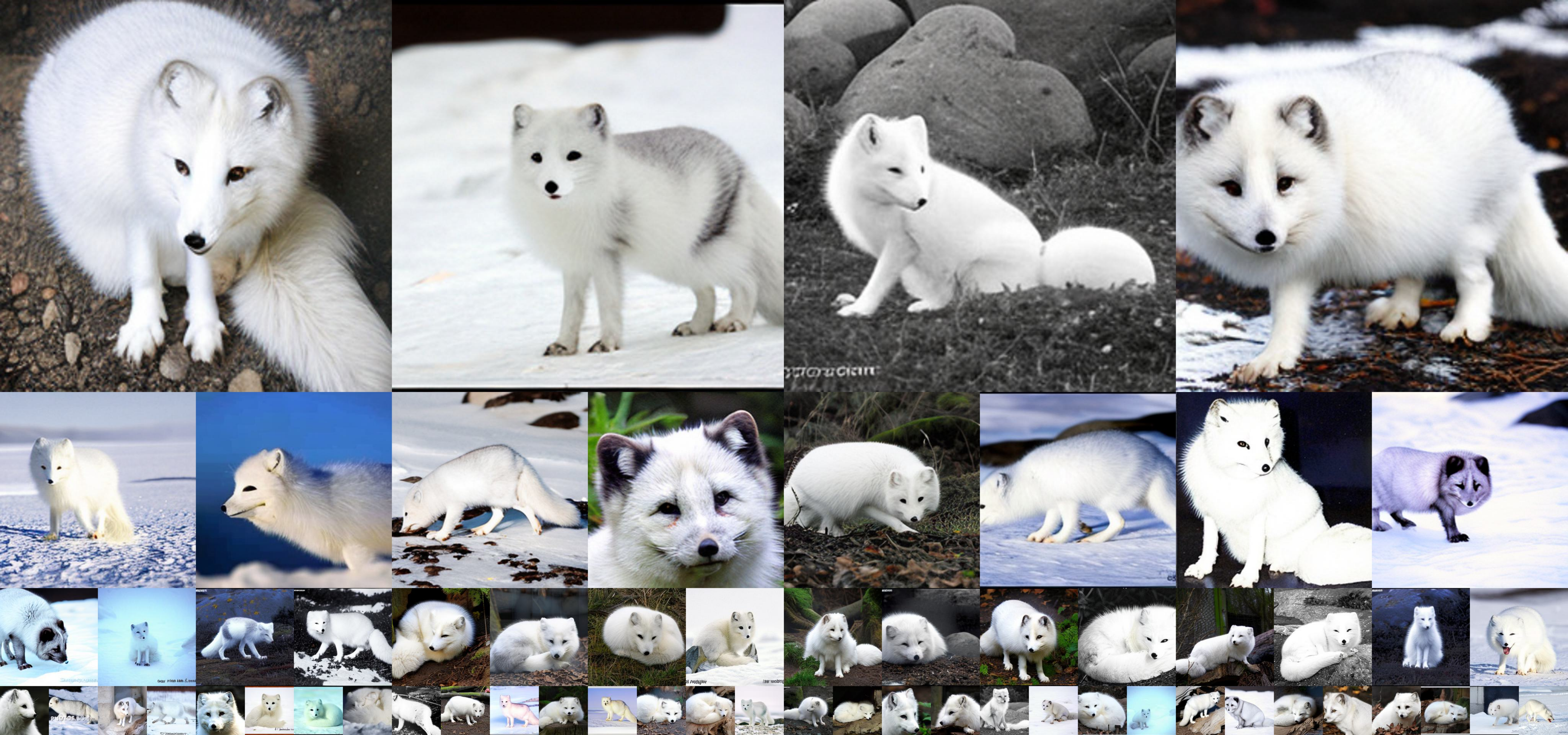}
    \caption{\textbf{Uncurated generation results of SiT-XL/2 + \sname.} We use classifier-free guidance with $w=4.0$. Class label = ``arctic fox'' (279).}
\end{figure}
\begin{figure}[ht!]
    \centering
    \includegraphics[width=\linewidth]{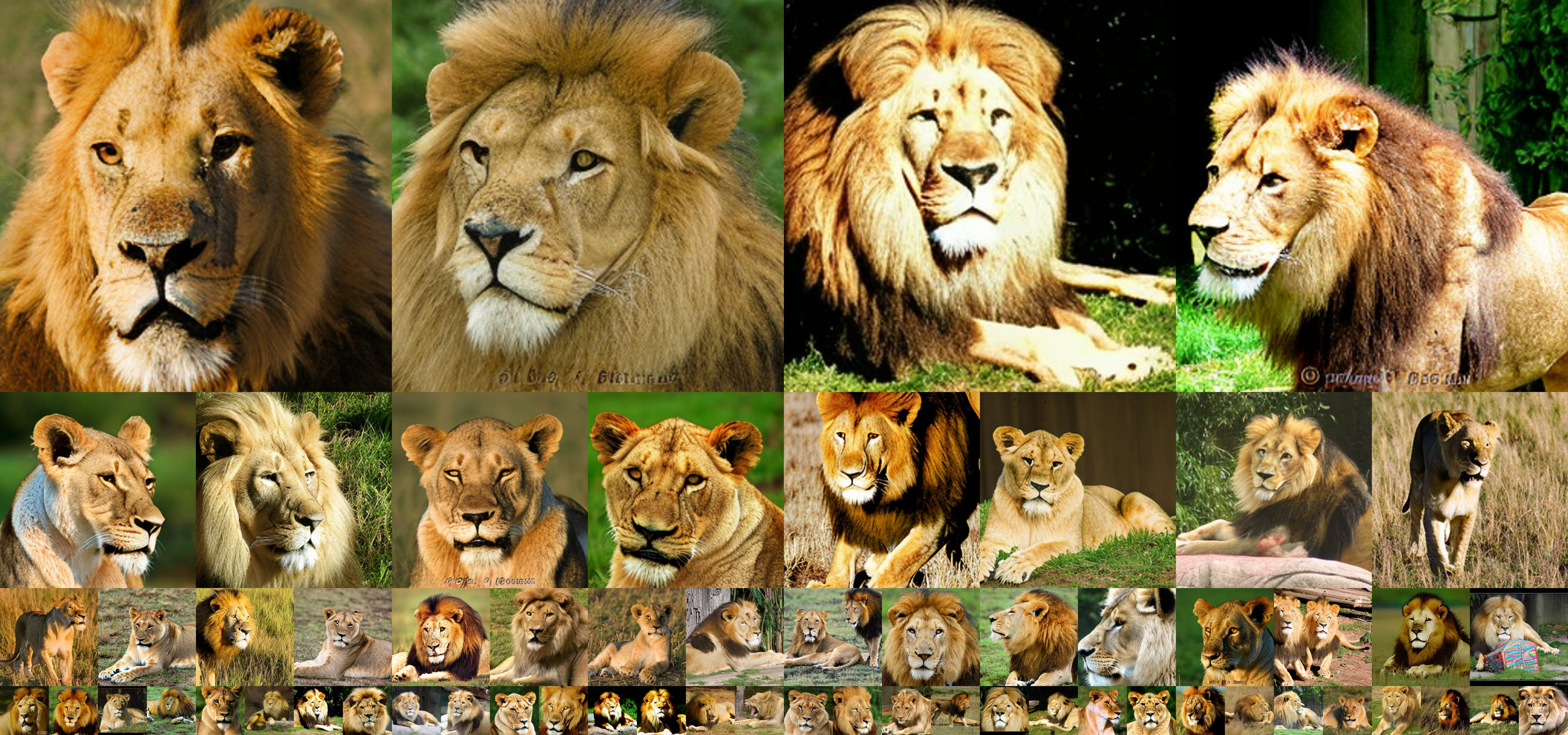}
    \caption{\textbf{Uncurated generation results of SiT-XL/2 + \sname.} We use classifier-free guidance with $w=4.0$. Class label = ``lion'' (291).}
\end{figure}
\clearpage
\begin{figure}[ht!]
    \centering
    \includegraphics[width=\linewidth]{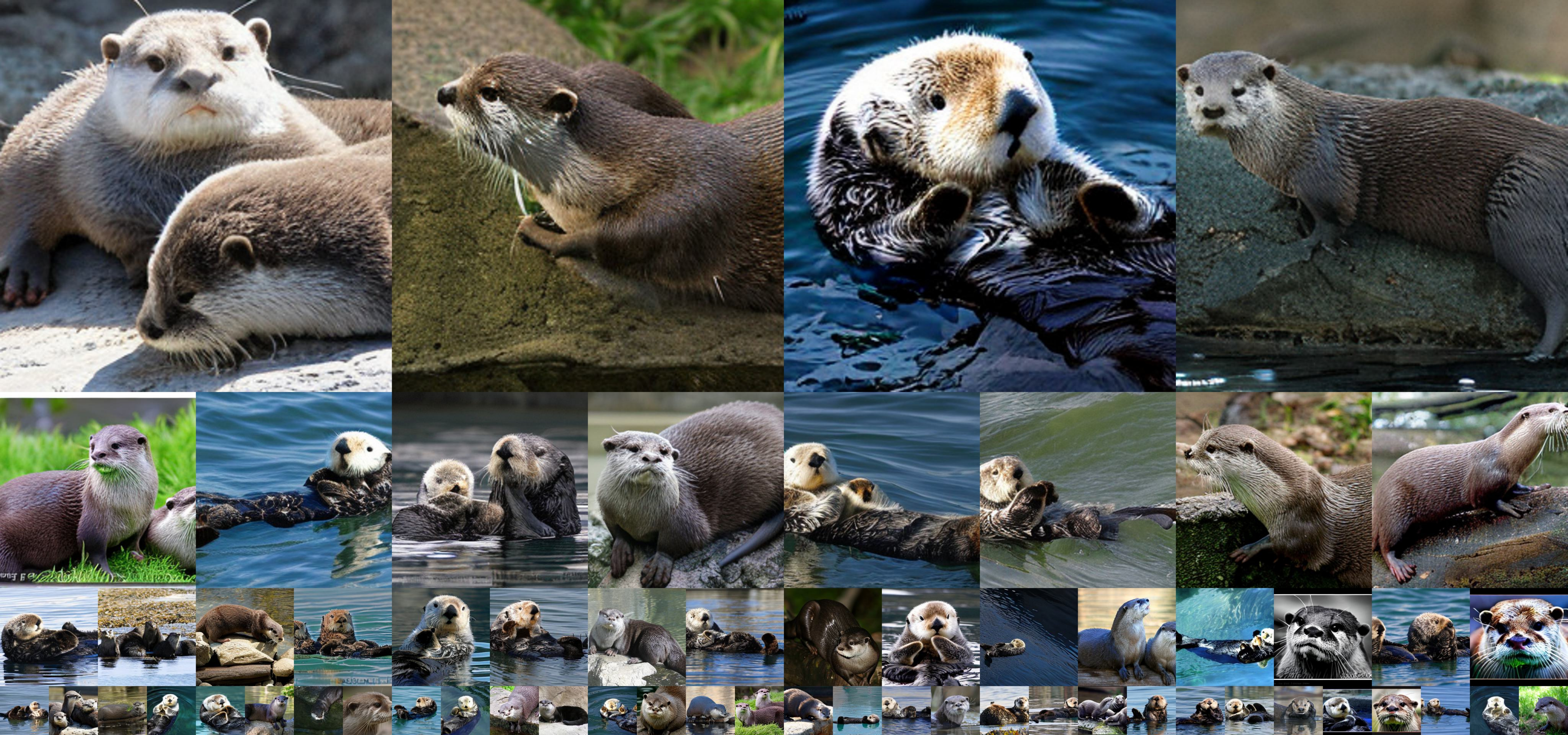}
    \caption{\textbf{Uncurated generation results of SiT-XL/2 + \sname.} We use classifier-free guidance with $w=4.0$. Class label = ``otter'' (360).}
\end{figure}
\begin{figure}[ht!]
    \centering
    \includegraphics[width=\linewidth]{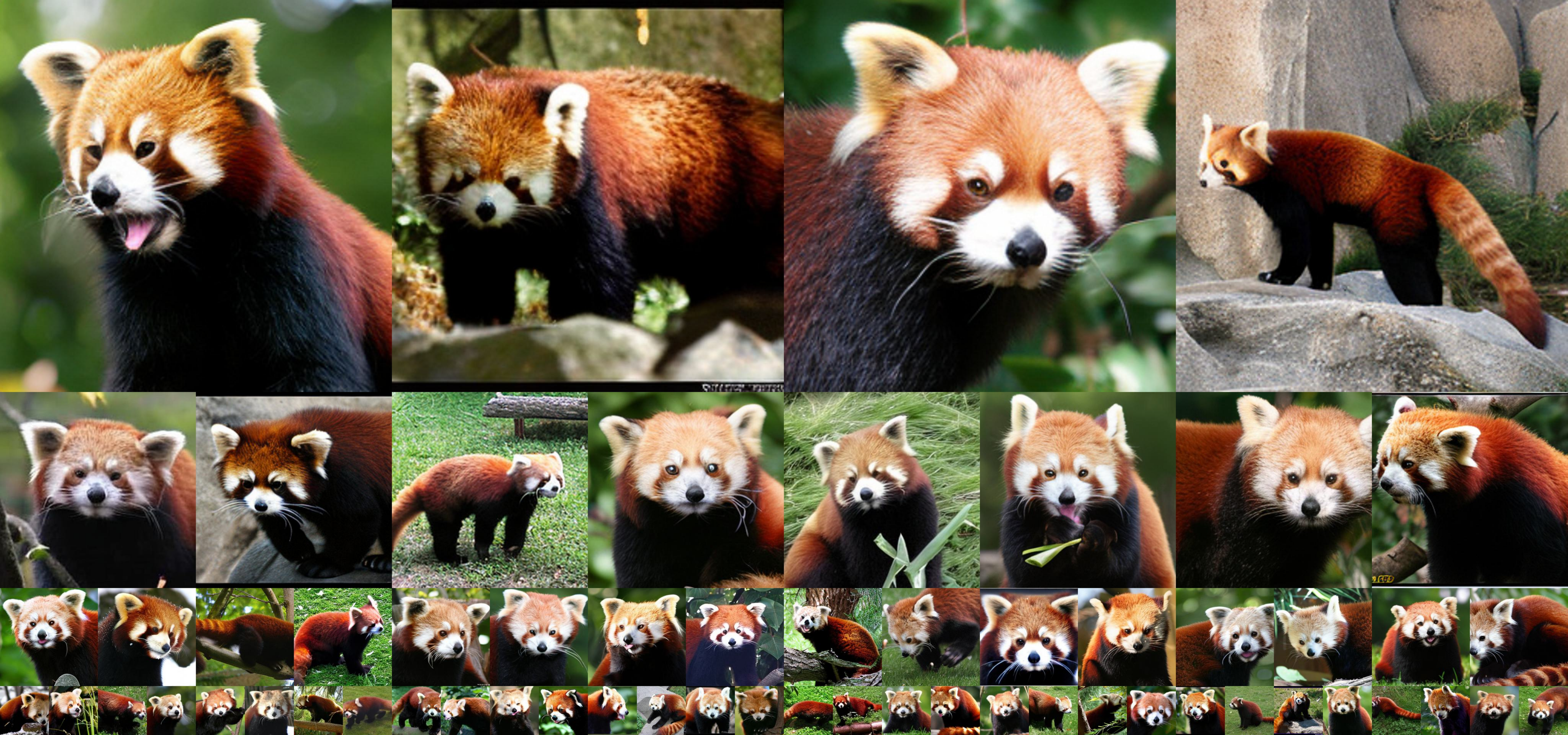}
    \caption{\textbf{Uncurated generation results of SiT-XL/2 + \sname.} We use classifier-free guidance with $w=4.0$. Class label = ``red panda'' (387).}
\end{figure}
\clearpage
\begin{figure}[ht!]
    \centering
    \includegraphics[width=\linewidth]{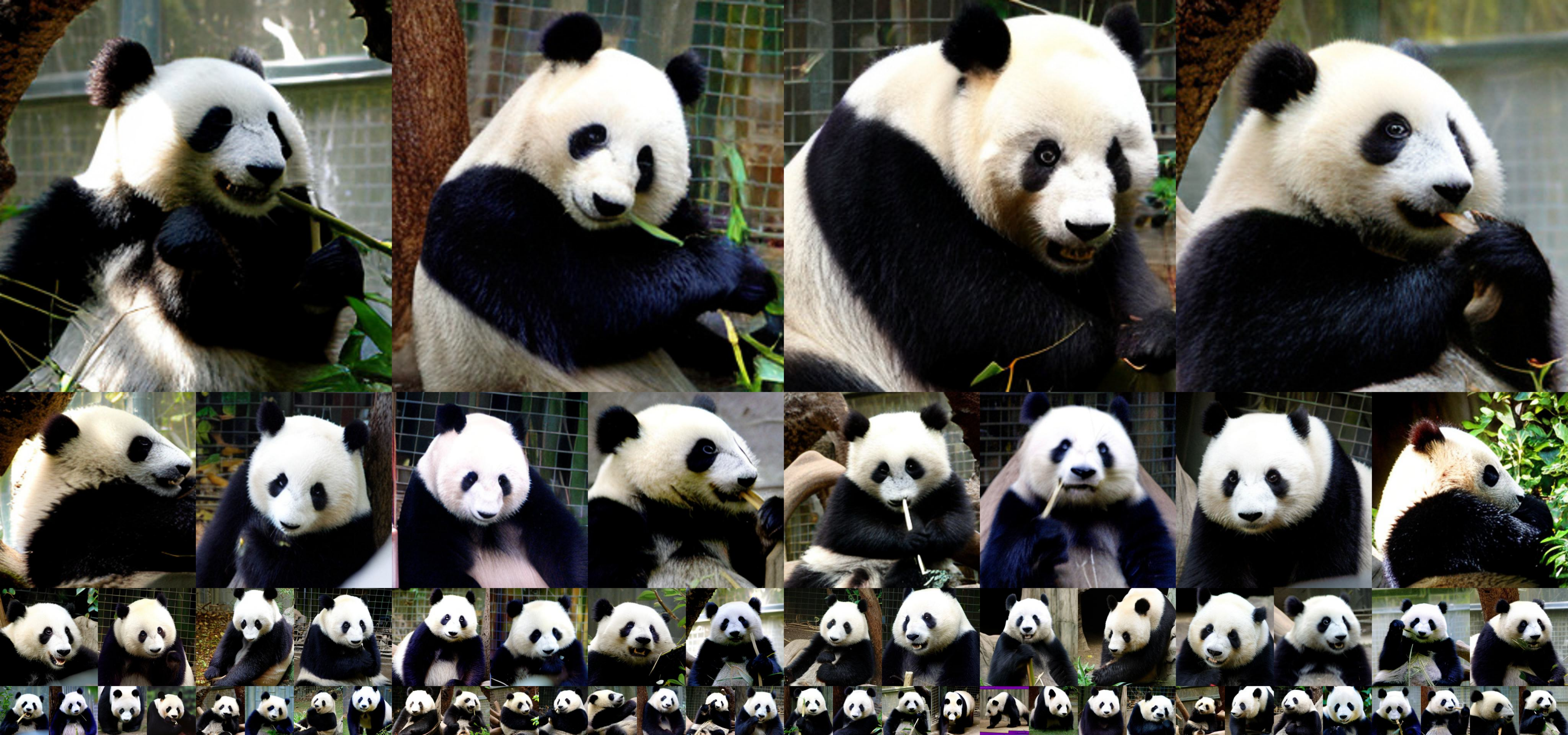}
    \caption{\textbf{Uncurated generation results of SiT-XL/2 + \sname.} We use classifier-free guidance with $w=4.0$. Class label = ``panda'' (388).}
\end{figure}
\begin{figure}[ht!]
    \centering
    \includegraphics[width=\linewidth]{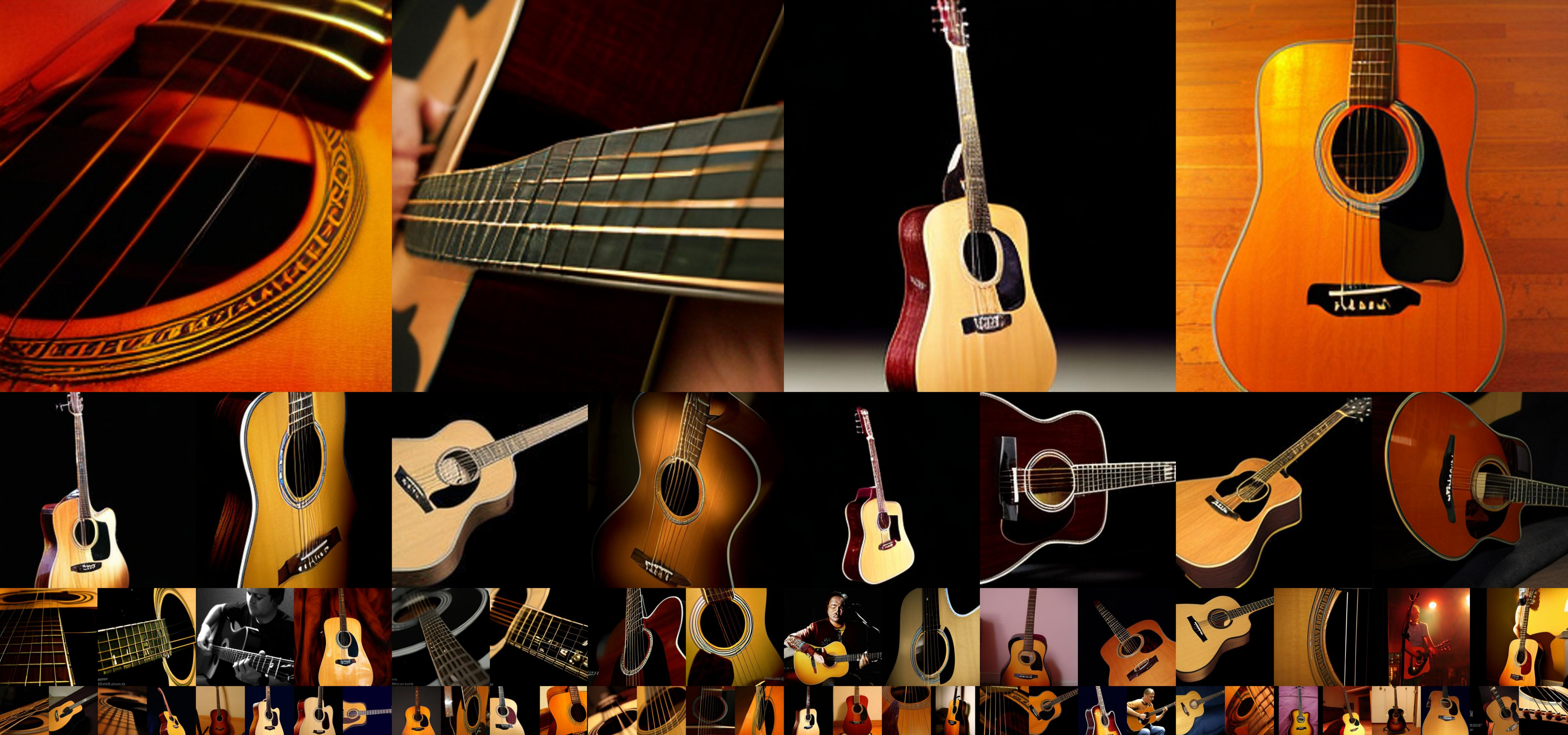}
    \caption{\textbf{Uncurated generation results of SiT-XL/2 + \sname.} We use classifier-free guidance with $w=4.0$. Class label = ``acoustic guitar'' (402).}
\end{figure}
\clearpage
\begin{figure}[ht!]
    \centering
    \includegraphics[width=\linewidth]{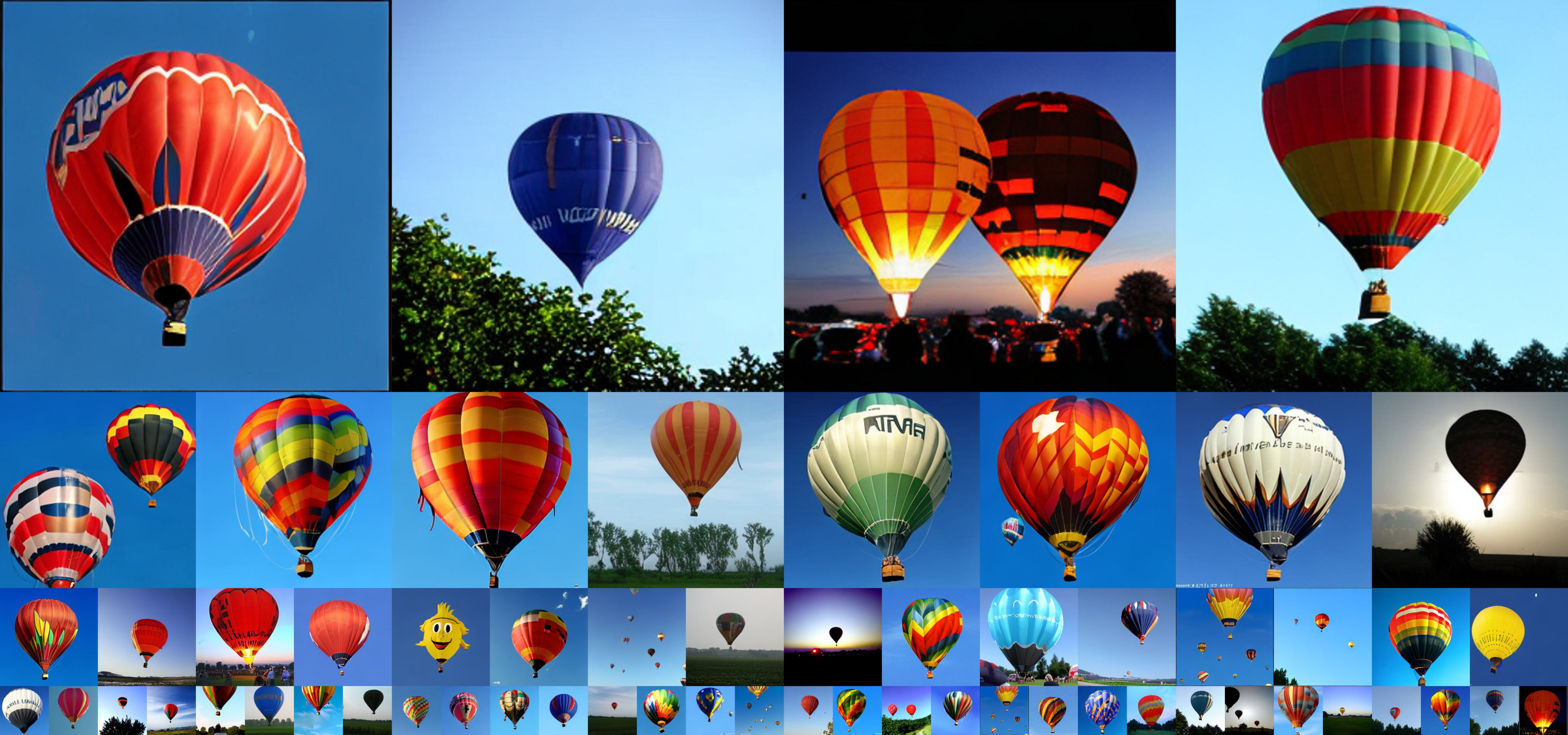}
    \caption{\textbf{Uncurated generation results of SiT-XL/2 + \sname.} We use classifier-free guidance with $w=4.0$. Class label = ``balloon'' (417).}
\end{figure}
\begin{figure}[ht!]
    \centering
    \includegraphics[width=\linewidth]{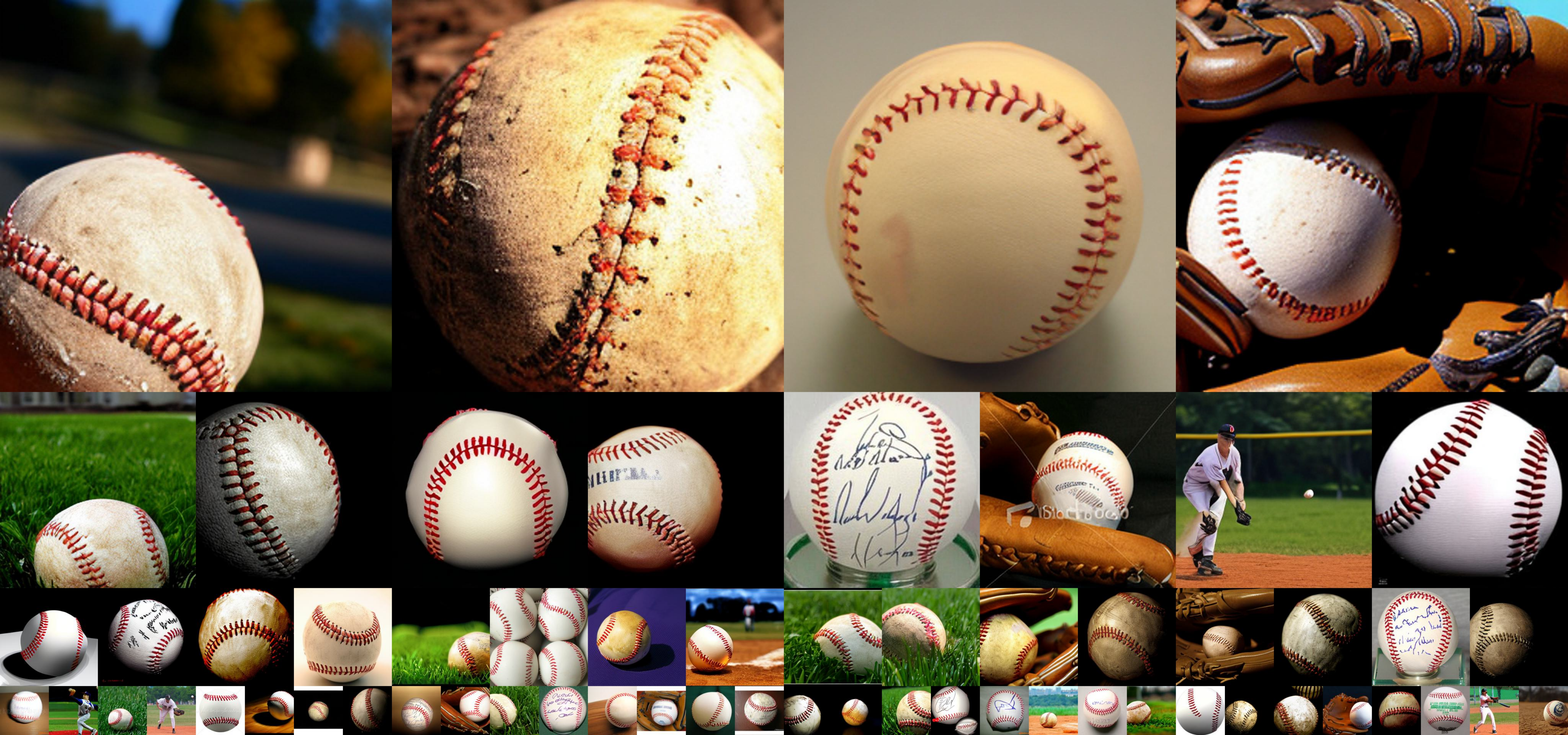}
    \caption{\textbf{Uncurated generation results of SiT-XL/2 + \sname.} We use classifier-free guidance with $w=4.0$. Class label = ``baseball'' (429).}
\end{figure}
\clearpage
\begin{figure}[ht!]
    \centering
    \includegraphics[width=\linewidth]{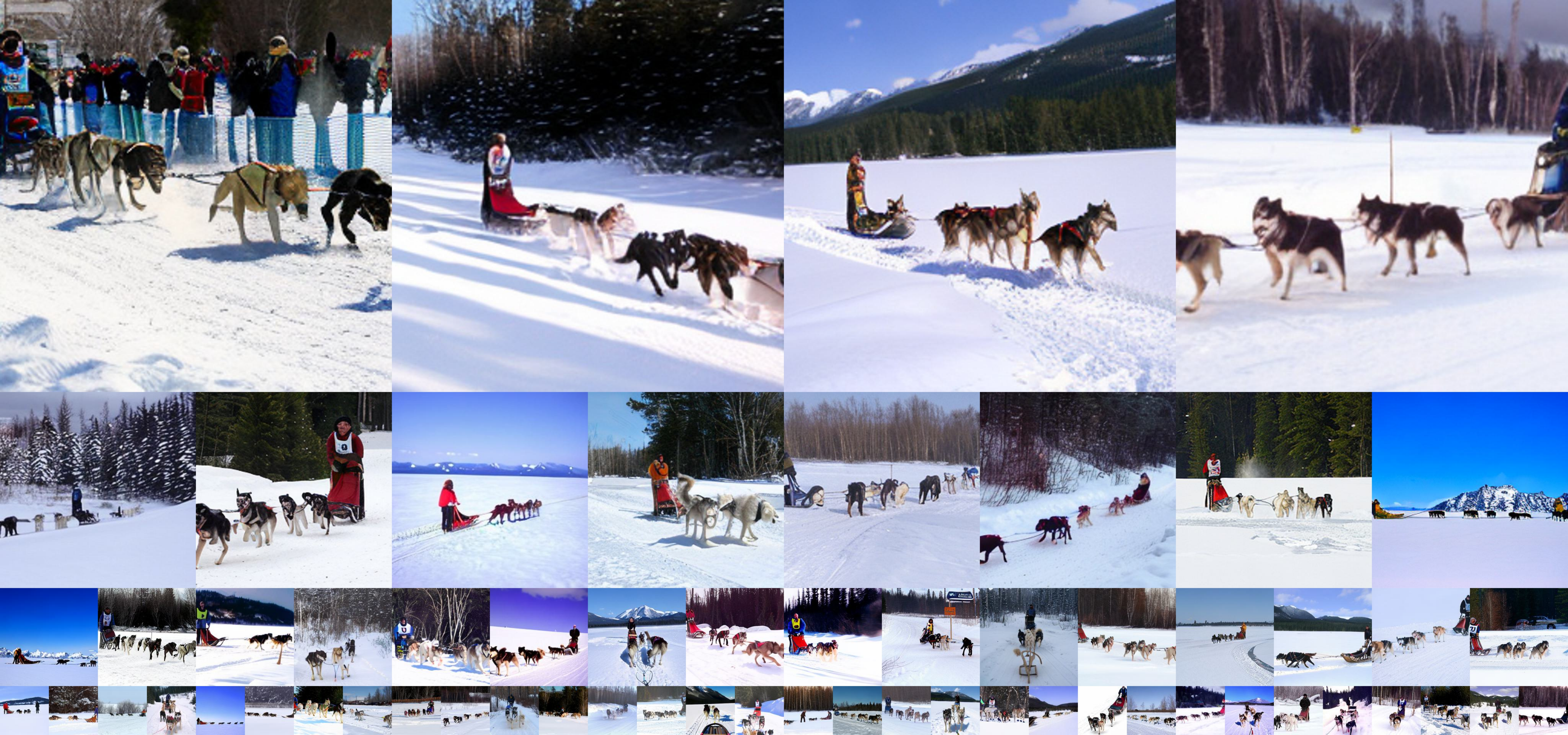}
    \caption{\textbf{Uncurated generation results of SiT-XL/2 + \sname.} We use classifier-free guidance with $w=4.0$. Class label = ``dog sled'' (537).}
\end{figure}
\begin{figure}[ht!]
    \centering
    \includegraphics[width=\linewidth]{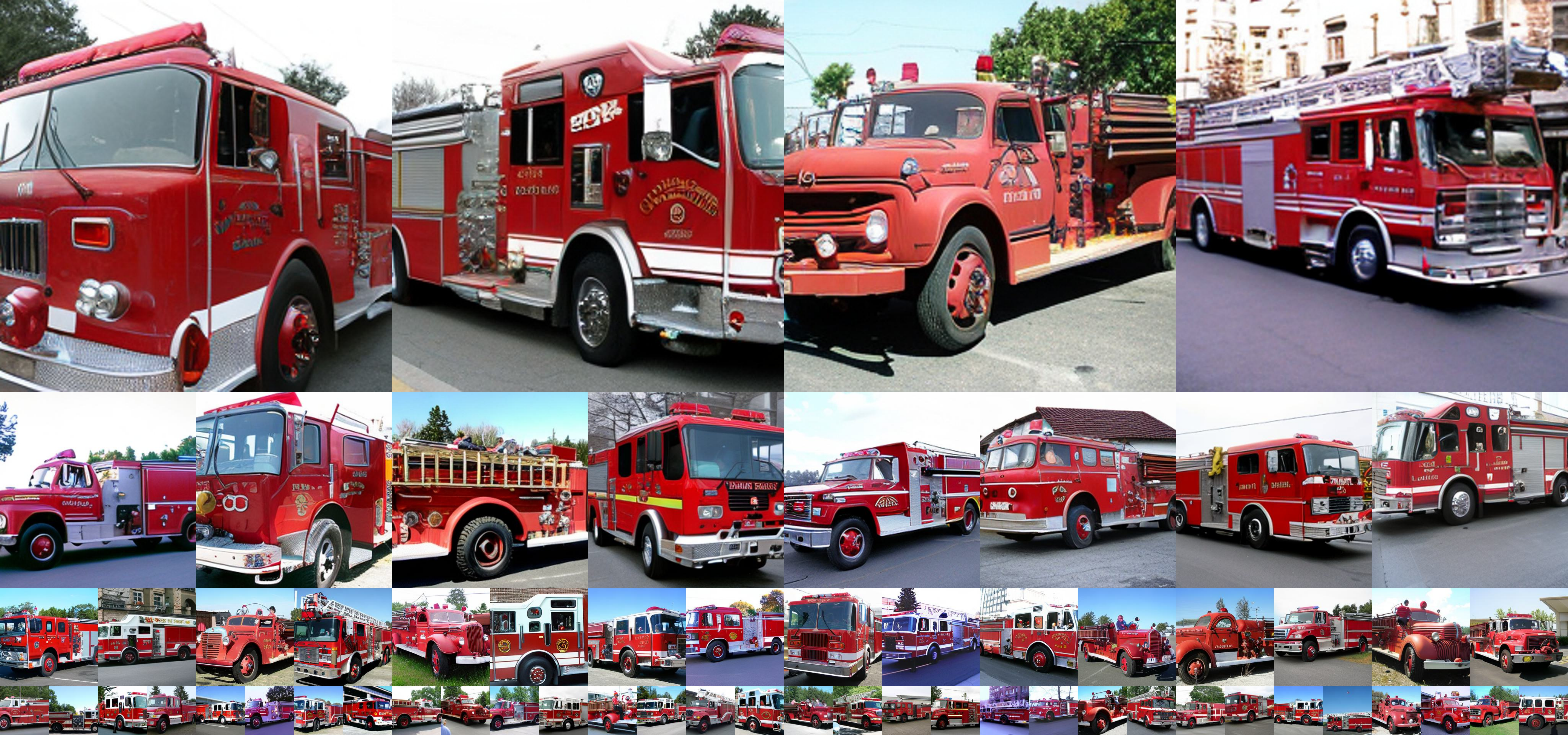}
    \caption{\textbf{Uncurated generation results of SiT-XL/2 + \sname.} We use classifier-free guidance with $w=4.0$. Class label = ``fire truck'' (555).}
\end{figure}
\clearpage
\begin{figure}[ht!]
    \centering
    \includegraphics[width=\linewidth]{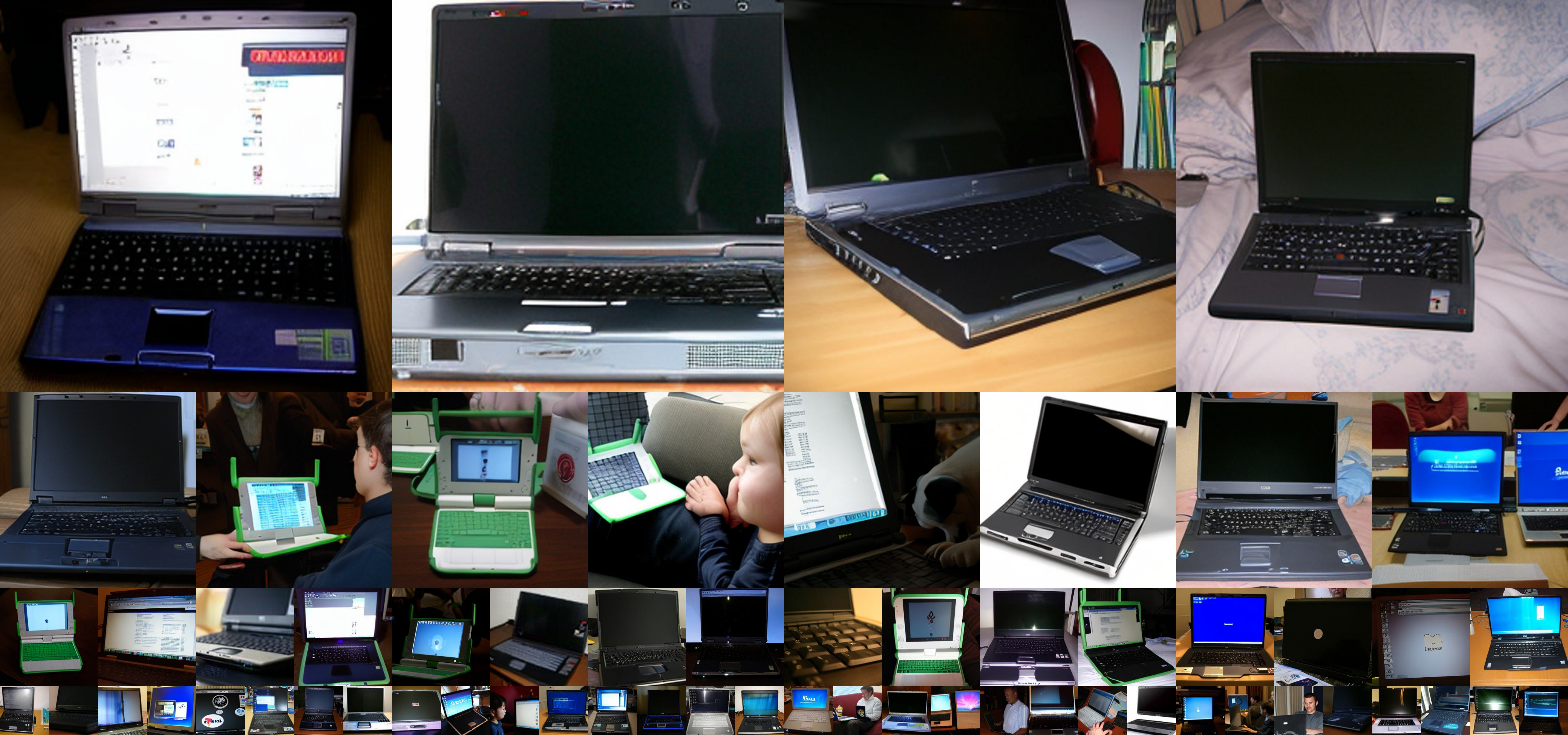}
    \caption{\textbf{Uncurated generation results of SiT-XL/2 + \sname.} We use classifier-free guidance with $w=4.0$. Class label = ``laptop'' (620).}
\end{figure}
\begin{figure}[ht!]
    \centering
    \includegraphics[width=\linewidth]{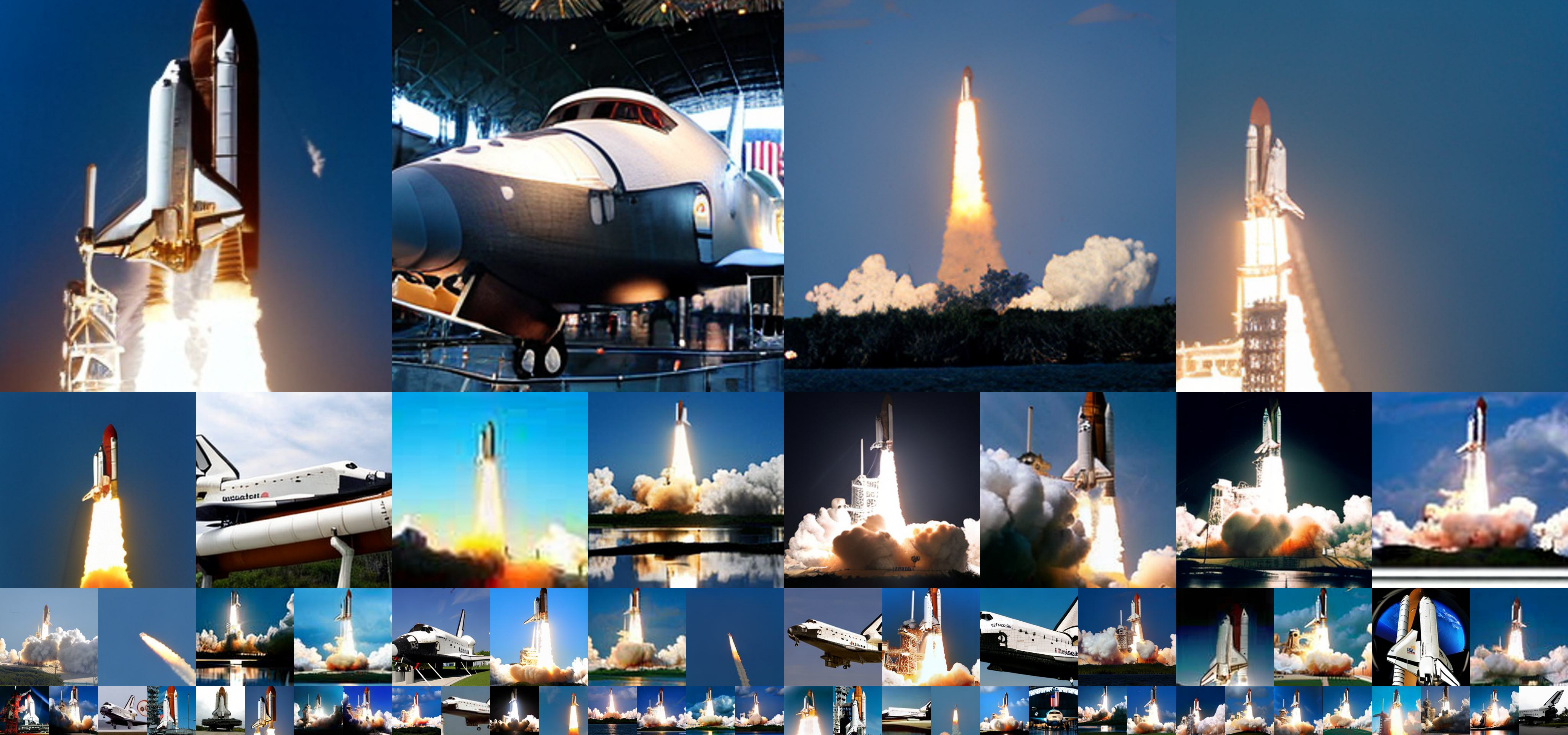}
    \caption{\textbf{Uncurated generation results of SiT-XL/2 + \sname.} We use classifier-free guidance with $w=4.0$. Class label = ``space shuttle'' (812).}
\end{figure}
\clearpage
\begin{figure}[ht!]
    \centering
    \includegraphics[width=\linewidth]{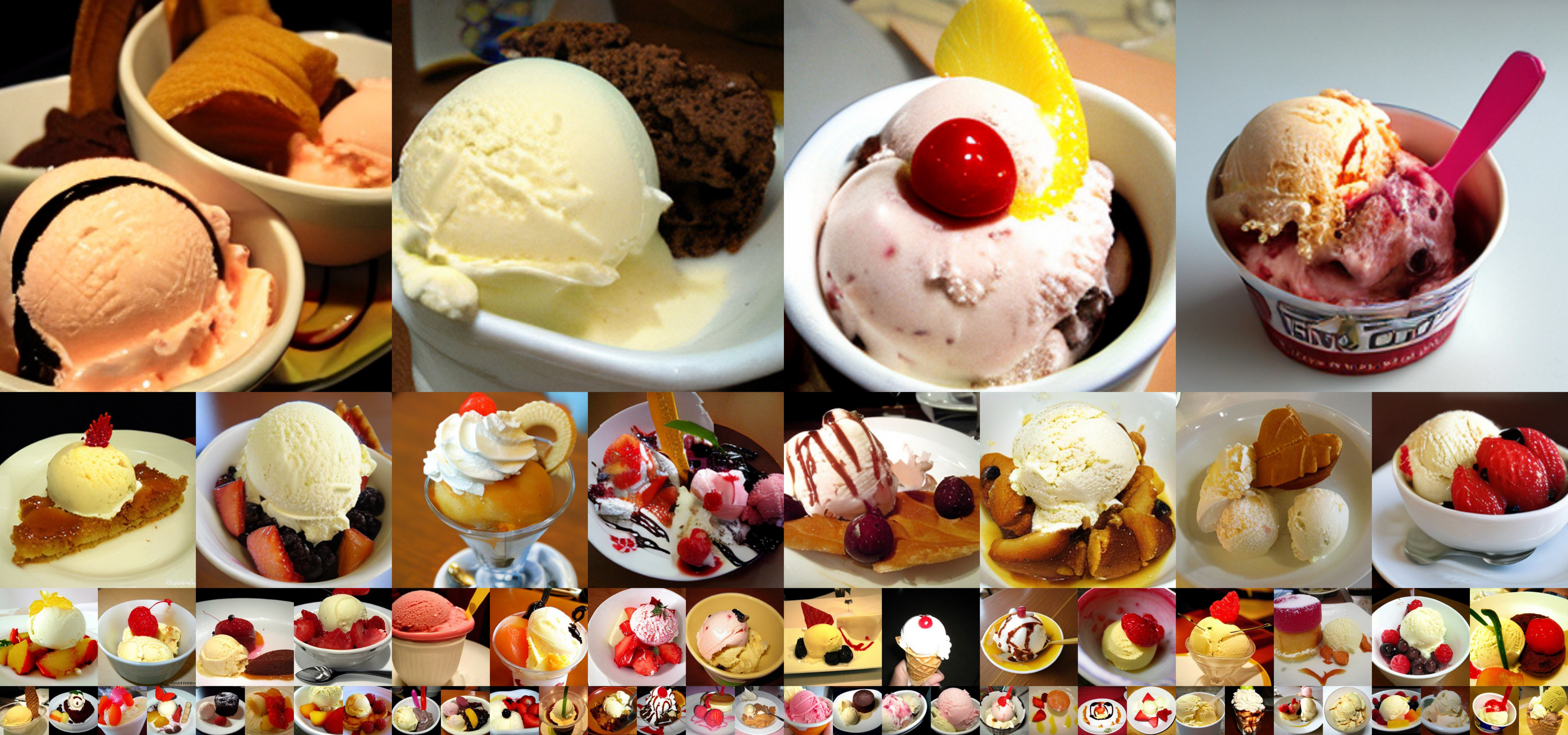}
    \caption{\textbf{Uncurated generation results of SiT-XL/2 + \sname.} We use classifier-free guidance with $w=4.0$. Class label = ``ice cream'' (928).}
\end{figure}
\begin{figure}[ht!]
    \centering
    \includegraphics[width=\linewidth]{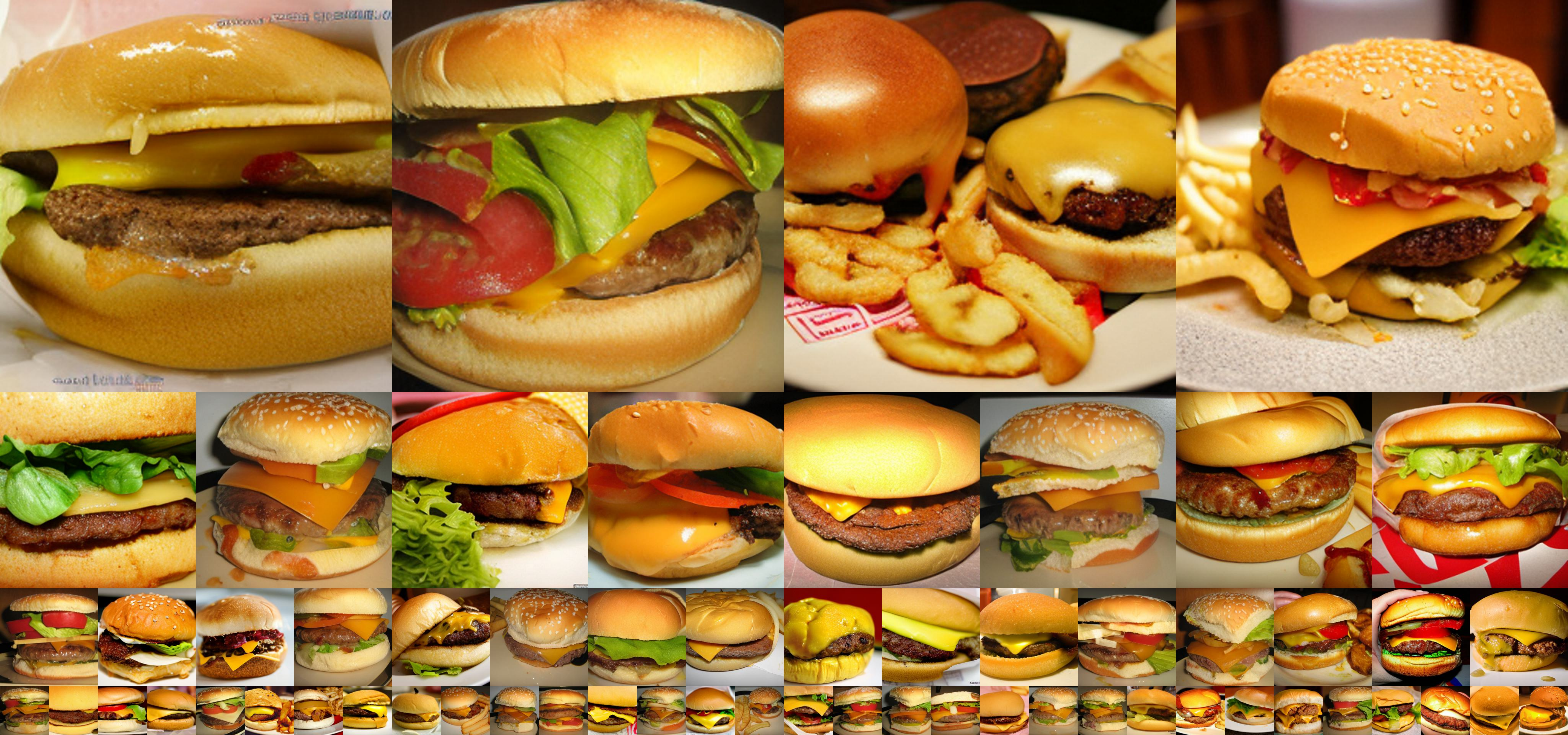}
    \caption{\textbf{Uncurated generation results of SiT-XL/2 + \sname.} We use classifier-free guidance with $w=4.0$. Class label = ``cheeseburger'' (933).}
\end{figure}
\clearpage
\begin{figure}[ht!]
    \centering
    \includegraphics[width=\linewidth]{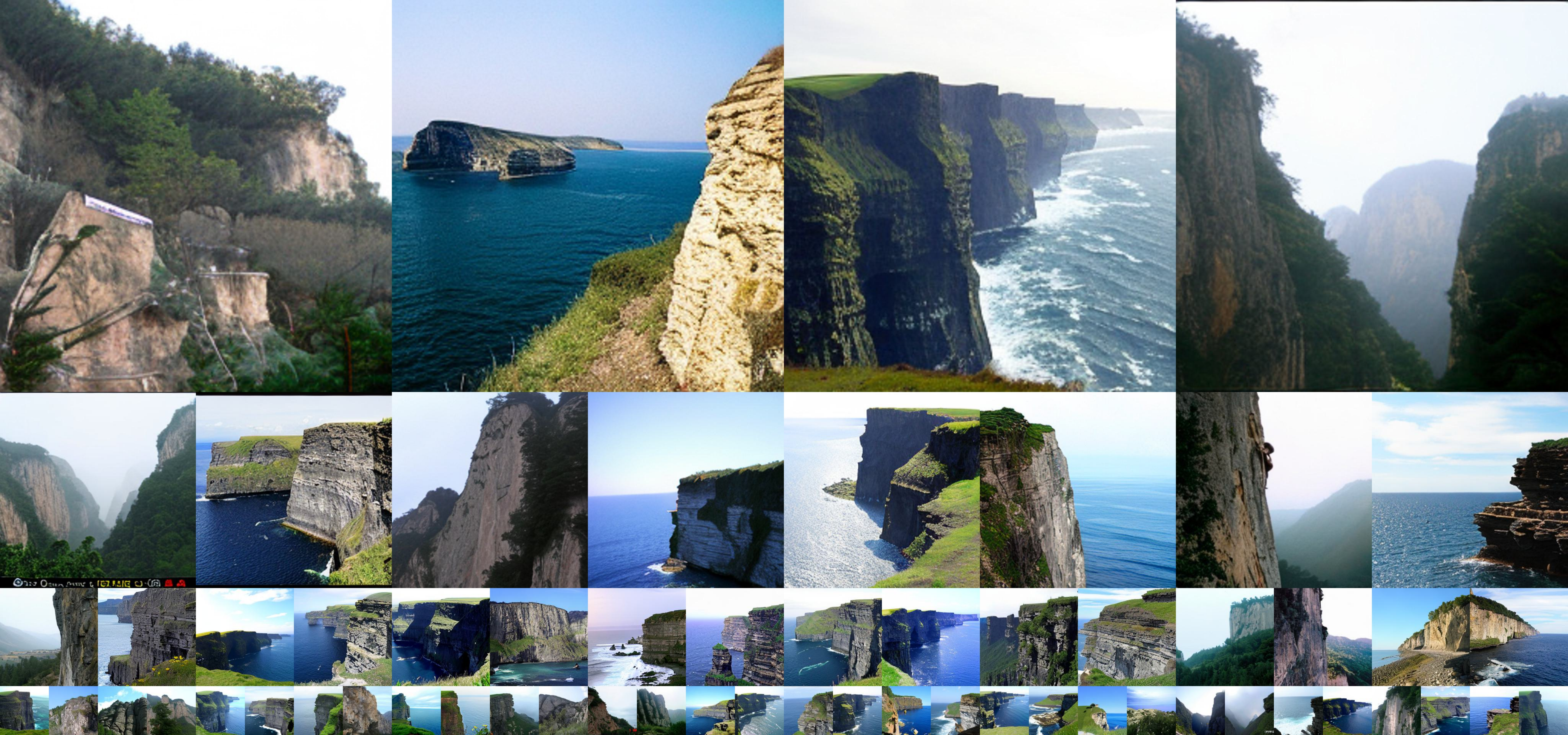}
    \caption{\textbf{Uncurated generation results of SiT-XL/2 + \sname.} We use classifier-free guidance with $w=4.0$. Class label = ``cliff drop-off'' (972).}
\end{figure}
\begin{figure}[ht!]
    \centering
    \includegraphics[width=\linewidth]{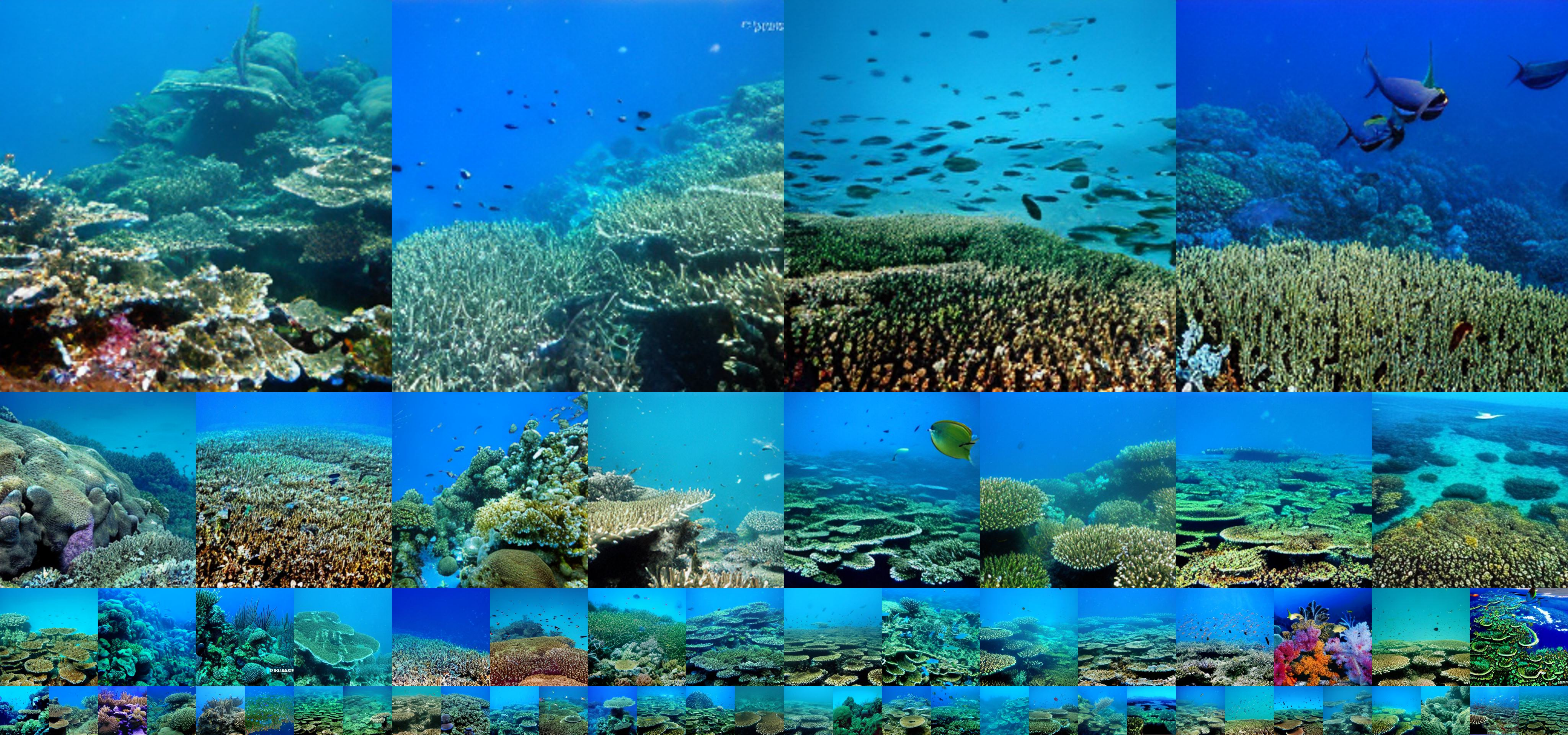}
    \caption{\textbf{Uncurated generation results of SiT-XL/2 + \sname.} We use classifier-free guidance with $w=4.0$. Class label = ``coral reef'' (973).}
\end{figure}
\clearpage
\begin{figure}[ht!]
    \centering
    \includegraphics[width=\linewidth]{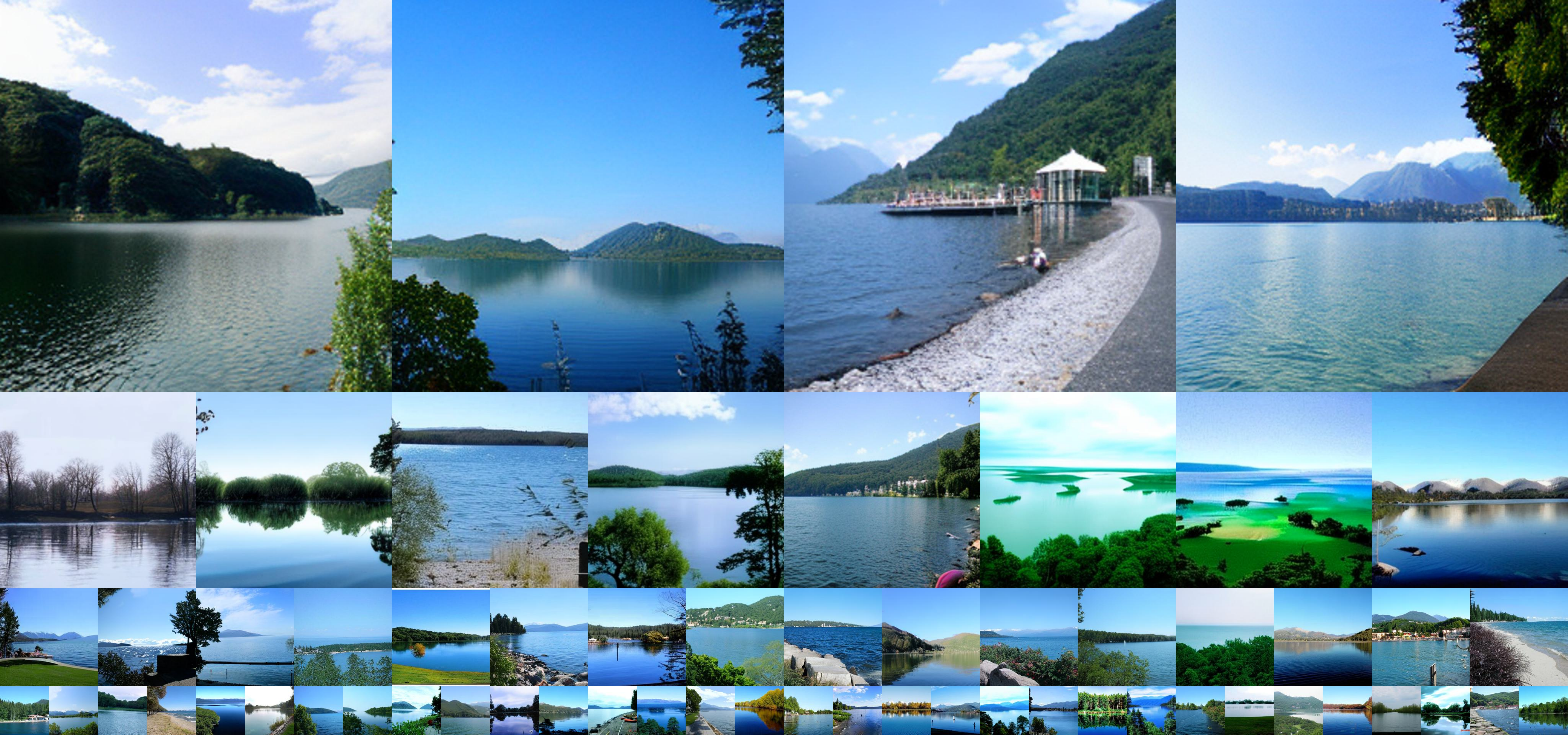}
    \caption{\textbf{Uncurated generation results of SiT-XL/2 + \sname.} We use classifier-free guidance with $w=4.0$. Class label = ``lake shore'' (975).}
\end{figure}
\begin{figure}[ht!]
    \centering
    \includegraphics[width=\linewidth]{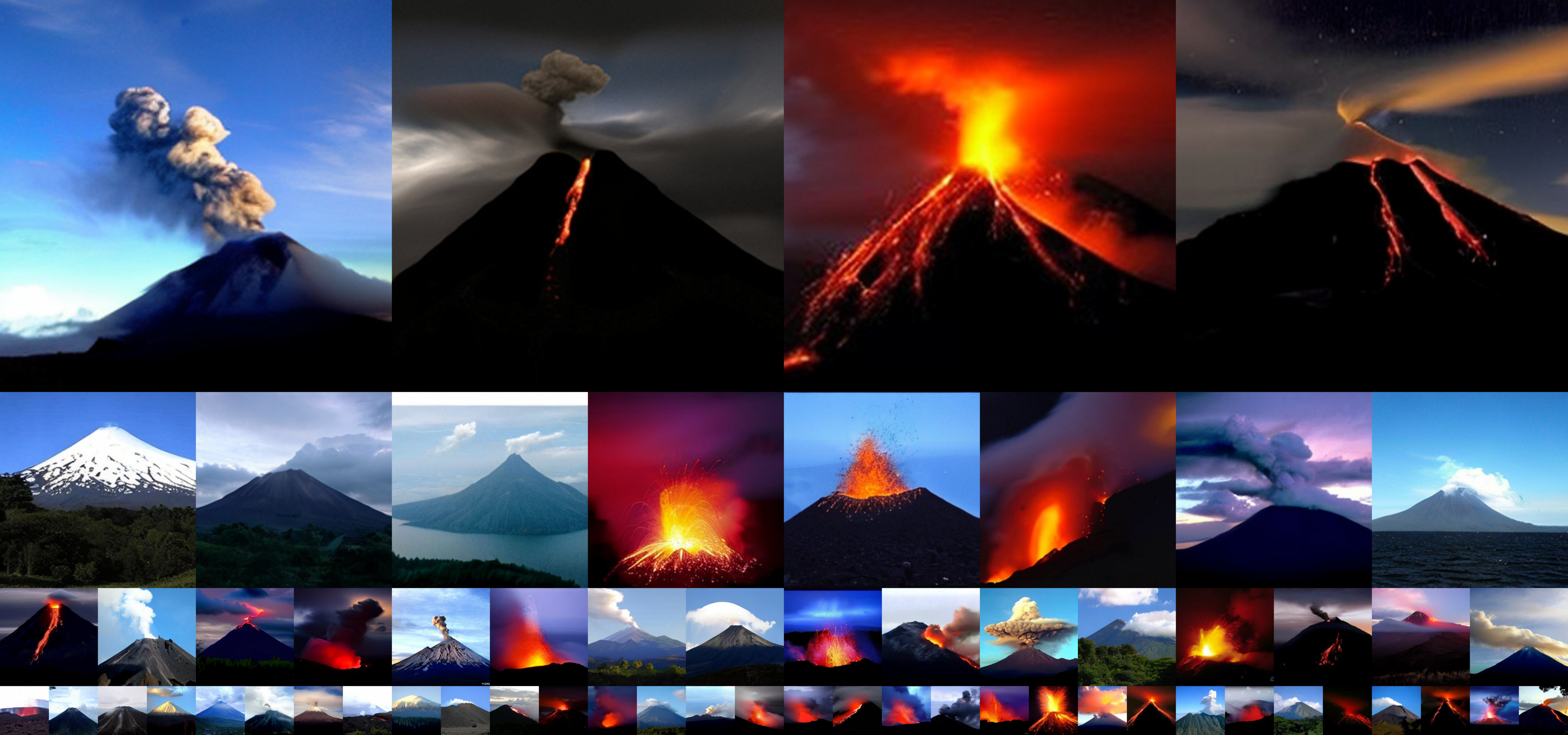}
    \caption{\textbf{Uncurated generation results of SiT-XL/2 + \sname.} We use classifier-free guidance with $w=4.0$. Class label = ``volcano'' (980).}
\end{figure}
\clearpage

\section{More Discussion on Related Work}
\label{appen:related}
\textbf{Pretrained visual encoders for generative models.}
First, there have been several approaches in generative adversarial network (GAN; \citealt{goodfellow2014generative}) that try to accelerate training with better convergence using pretrained visual encoders \citep{sauer2021projected, kumari2022ensembling,sauer2022stylegan,sauer2023stylegan,kang2023scaling}. They usually use pretrained visual encoders as a discriminator by leveraging their intermediate features. This approach has also been applied to the distillation of diffusion models with adversarial objectives~\citep {sauer2023adversarial,sauer2024fast,kang2024distilling}. Another line of work tries to exploit the pretrained visual encoders for improving diffusion model training from scratch~\citep{pernias2024wrstchen, li2023return}, usually by training two diffusion models where one model generates the pretrained representations and the other model generates the target data conditioned on the generated representation. Our method also tries to improve diffusion model training through pretrained visual encoders, but our motivation is in the alignment between the diffusion model representation and recent self-supervised visual representations.

\textbf{Denoising transformers.}
Many recent works have tried to use transformer backbones for diffusion or flow-based model training. First, several works like U-ViT~\citep{bao2022all}, MDT~\citep{gao2023mdtv2}, and DiffiT~\citep{hatamizadeh2023diffit} show transformer-based backbones with \emph{skip connections} can be an effective backbone for training diffusion models. Intriguingly, DiT~\citep{Peebles2022DiT} show skip connections are not even necessary components, and a pure transformer architecture can be a scalable architecture for training diffusion-based models. Based on DiT, SiT~\citep{ma2024sit} shows the model can be further improved with continuous stochastic interpolants~\citep{albergo2023stochastic}. Moreover, VDT~\citep{lu2023vdt} and Latte \citep{ma2024latte} show DiTs can be extended for video generation through a space-time attention~\citep{arnab2021vivit}. Based on these improvements, Pixart-$\alpha$~\citep{chen2023pixart}, Pixart-$\Sigma$~\citep{chen2024pixartsigma}, Stable diffusion 3~\citep{esser2024scaling} show pure transformers can be scaled up for challenging text-to-image generation, and CMD~\citep{yu2024efficient}, WALT~\citep{gupta2023photorealistic}, and Sora~\citep{videoworldsimulators2024} demonstrates their success in text-to-video generation. Our work analyzes and improves the training of DiT (and SiT) architecture based on a simple feature matching regularization to the early layers.

\textbf{Generative models with auxiliary self-supervised tasks.} MaskDiT \citep{zheng2024fast} combines mask reconstruction in MAE \citep{he2022masked} to diffusion model training for faster diffusion model training. Similarly, SD-DiT \citep{zhu2024sd} shows diffusion model training can be improved with an auxiliary discriminative self-supervised loss. MAGE \citep{li2023mage} bridge MAE training and masked image modeling \citep{chang2022maskgit} by adjusting the masking ratio in training, which leads to a single model both capable of discrimination and generation tasks. Our method also has a similarity to these works, where our training scheme has an additional distillation loss to projection of diffusion transformer hidden states.

\textbf{Denoising as self-supervised learning task.}
There have been some attempts to improve {self-supervised learning with denoising}: \citet{abstreiter2021diffusion} extends the diffusion objective for a better representation learning scheme, and \citet{chen2024deconstructing} deconstructs diffusion models to improve denoising-based representation learning. \citet{hudson2024soda} introduces an encoder that learns a representation by guiding a diffusion with its output as a compact latent vector. \citet{zaidi2022pre} focuses on the molecular domain and proposes a pretraining scheme based on denoising. Our work also analyzes representational gap between popular self-supervised networks and those learned from pretrained diffusion models.
\clearpage

\section{ImageNet 512$\times$512 Experiment}
\label{appen:512}
To validate the scalability of \sname with respect to the input image resolution, we conduct an additional experiment on ImageNet 512$\times$ 512. We strictly follow the setup used in our ImageNet 256$\times$256 experiment except an input dimension. Specifically, the input dimension for SiT becomes a $64\times64\times4$ compressed latent image from $512\times512\times3$ image pixels using stable diffusion VAE \citep{rombach2022high}. Moreover, we use an image resized to $448 \times 448$ resolution for an input to the DINOv2 \citep{oquab2024dinov} encoder with an interpolation of the positional embedding. 

\begin{figure*}[ht!]
\vspace{-0.05in}
    \centering
    \includegraphics[width=.95\linewidth]{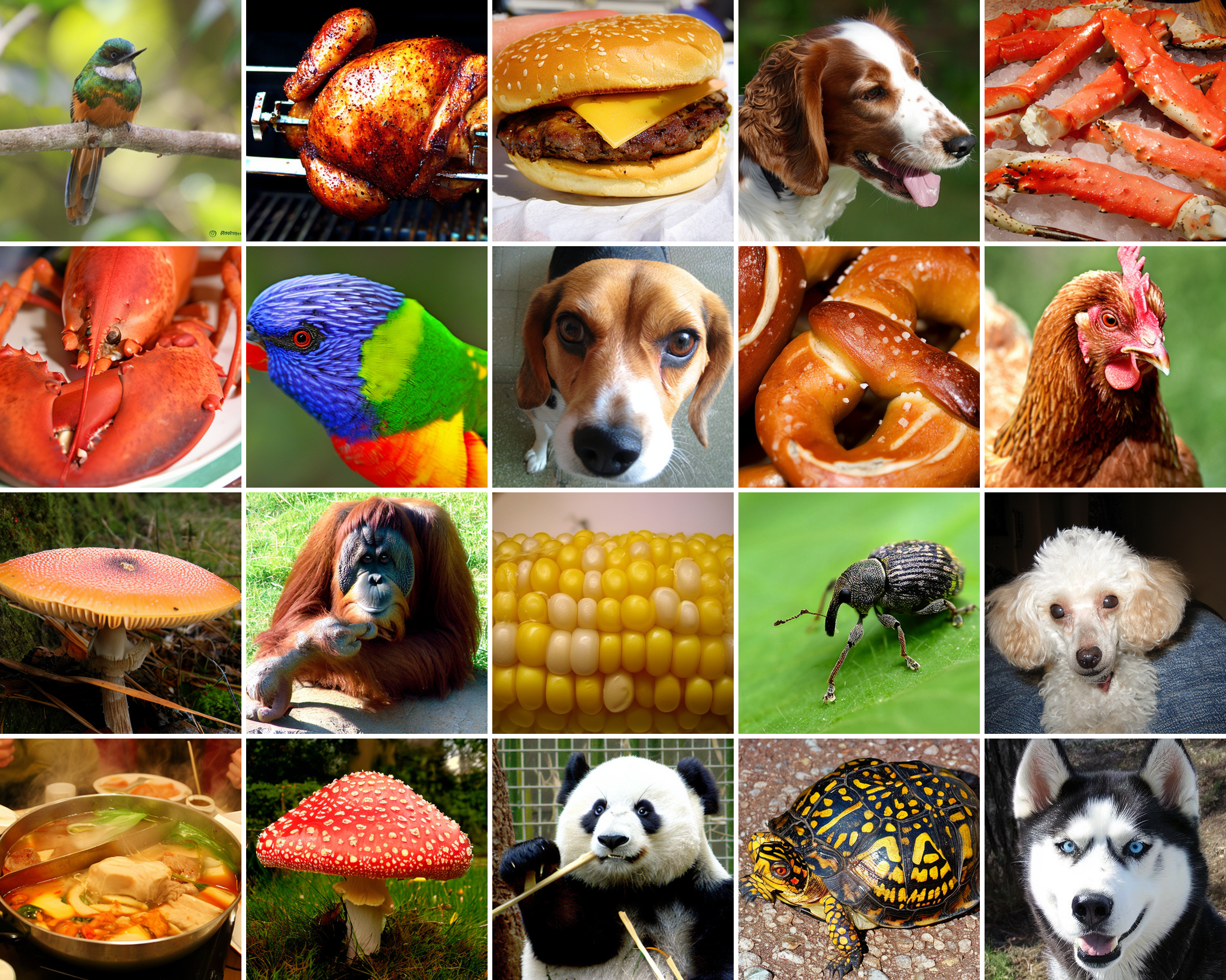}
    \caption{\textbf{Samples on ImageNet 512$\times$512} from SiT-XL/2+\sname using CFG with $w=4.0$.
    }
    \label{fig:512}
    \vspace{-0.2in}
\end{figure*}

We provide quantitative results in Table~\ref{tab:512} and qualitative result in Figure~\ref{fig:512}. Notably, as shown in this table, the model (with REPA) already outperforms the vanilla SiT-XL/2 in terms of four metrics (FID, sFID, IS, and Prec) using $>$3$\times$ fewer training iterations.

\begin{table}[h!]
\centering\small
\caption{\textbf{System-level comparison} on ImageNet 512$\times$512. We use CFG with $w=1.35$. 
}
\begin{tabular}{l c c c c c c}
\toprule
{\pz\pz Model} & Epochs  &  {\pz FID$\downarrow$} & {sFID$\downarrow$} & {IS$\uparrow$} & {Pre.$\uparrow$} & Rec.$\uparrow$ \\
\arrayrulecolor{black}\midrule
\multicolumn{7}{l}{\emph{Pixel diffusion}\vspace{0.02in}} \\
\pz\pz {VDM$++$}                   & -   & 2.65 & -    & 278.1 & -    & - \\
\pz\pz {ADM-G, ADM-U}              & 400 & 2.85 & 5.86 & 221.7 & 0.84 & 0.53 \\
\pz\pz Simple diffusion (U-Net)    & 800 & 4.28 & -    & 171.0 & - & - \\
\pz\pz Simple diffusion (U-ViT, L) & 800 & 4.53 & -    & 205.3 & - & - \\
\arrayrulecolor{black!40}\midrule
\multicolumn{7}{l}{\emph{Latent diffusion, Transformer}\vspace{0.02in}} \\
\pz\pz MaskDiT                     & 800 & 2.50 & 5.10 & 256.3 & 0.83 & 0.56 \\ 
\arrayrulecolor{black!30}\cmidrule(lr){1-7}
\pz\pz {DiT-XL/2}                  & 600 & 3.04 & 5.02 & 240.8 & 0.84 & 0.54  \\
\arrayrulecolor{black!30}\cmidrule(lr){1-7}
\pz\pz {SiT-XL/2}                  & 600 & 2.62 & 4.18 & 252.2 & 0.84 & {0.57} \\
{\pz\pz{+ \sname (ours)}}         &\pz80 & {2.44} & 4.21 & 247.3 &  0.84 & 0.56   \\
{\pz\pz{+ \sname (ours)}}         & 100  &  {2.32} & \textbf{4.16} & {255.7} & \textbf{0.84} & {0.56}  \\
{\pz\pz{+ \sname (ours)}}         & 200  & \textbf{2.08} & 4.19 & \textbf{274.6} & 0.83 & \textbf{0.58} \\
\arrayrulecolor{black}\bottomrule
\end{tabular}
\label{tab:512}
\end{table}

\clearpage
\section{Text-to-Image Generation Experiment}
\label{appen:t2i}
We also validate \sname in text-to-image generation. We mostly follow the experimental setup used in U-ViT \citep{bao2022all} unless otherwise specified: we train the model from scratch on a train split of the MS-COCO dataset \citep{lin2014microsoft} and use a validation split for evaluation. We use MMDiT \citep{esser2024scaling}, a simple variant of DiT that design attention layers to be jointly computed with image patches and text embeddings. We train MMDiT models for 150K iterations with a batch size of 256. We set a hidden dimension as 768 and a model depth as 24, and we use the CLIP \citep{radford2021learning} text encoder to compute text prompts from captions.

\begin{figure*}[ht!]
    \centering
    \includegraphics[width=\linewidth]{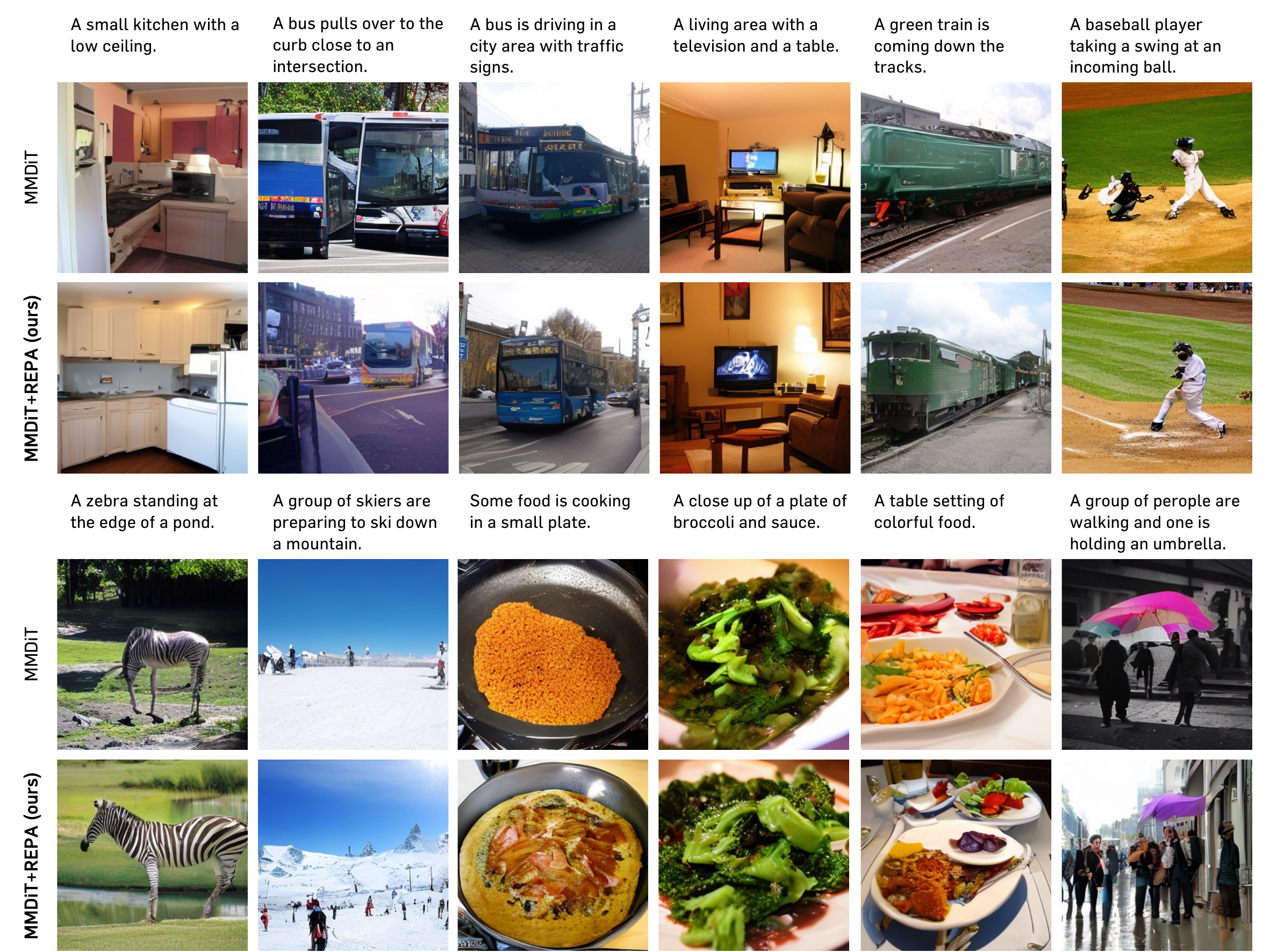}
    \caption{\textbf{Qualitative comparison on text-to-image generation (MS-COCO)}. We use classifier-free guidance with $w=4.0$.
    }
    \label{fig:t2i}
\end{figure*}

We report the results in Table~\ref{tab:t2i} and Figure~\ref{fig:t2i}. 
First, as shown in the qualitative comparison in Figure~\ref{fig:t2i}, \sname shows consistently better results than the vanilla model. Moreover, as shown in Table~\ref{tab:t2i}, \sname also shows considerable improvements in T2I generation, highlighting the importance of alignment of visual representations even under the presence of text representations. 

In this respect, we strongly believe that training large-scale T2I models with large-scale data using \sname will be a promising direction in the future.

\clearpage
\begin{table}[h!]
    \centering\small
\resizebox{.6\textwidth}{!}{
    \begin{tabular}{lcc}
    \toprule
    Method & Type & FID \\
    \midrule
     AttnGAN \citep{xu2018attngan}     & GAN                & 35.49 \\
     DM-GAN \citep{zhu2019dm}          & GAN                & 32.64 \\
     VQ-Diffusion \citep{gu2022vector} & Discrete Diffusion & 19.75 \\
     DF-GAN \citep{tao2022df}          & GAN                & 19.32 \\
     XMC-GAN \citep{zhang2021cross}    & GAN                & \pz9.33 \\
     Frido  \citep{fan2023frido}       & Diffusion          & \pz8.97 \\
     LAFITE \citep{zhou2022lafite}     & GAN                & \pz8.12 \\
    \arrayrulecolor{black!30}\midrule
    U-Net \citep{bao2022all}           & Diffusion          & \pz7.32 \\
    U-ViT-S/2 \citep{bao2022all}       & Diffusion          & \pz5.95 \\
    U-ViT-S/2 (Deep) \citep{bao2022all}& Diffusion          & \pz5.48 \\
    \arrayrulecolor{black!30}\midrule
    MMDiT (ODE; NFE=50)                & Diffusion          & \pz6.05 \\
    \textbf{MMDiT+REPA (ODE; NFE=50)}  & Diffusion          & \textbf{\pz4.73} \\
    \arrayrulecolor{black!30}\midrule
    MMDiT (SDE; NFE=250)               & Diffusion          & \pz5.30 \\
    \textbf{MMDiT+REPA (SDE; NFE=250)} & Diffusion          & \textbf{\pz4.14} \\
    \arrayrulecolor{black}\bottomrule
    \end{tabular}
}
    \caption{Quantitative comparison on text-to-image generation (MS-COCO). We use classifier-free guidance with $w=2.0$ following the setup in \citep{bao2022all}.}
    \label{tab:t2i}
\end{table}

\section{Feature Map Visualization}
\begin{figure*}[ht!]
    \centering
    \begin{subfigure}[b]{\textwidth}
        \centering
        \includegraphics[width=\textwidth]{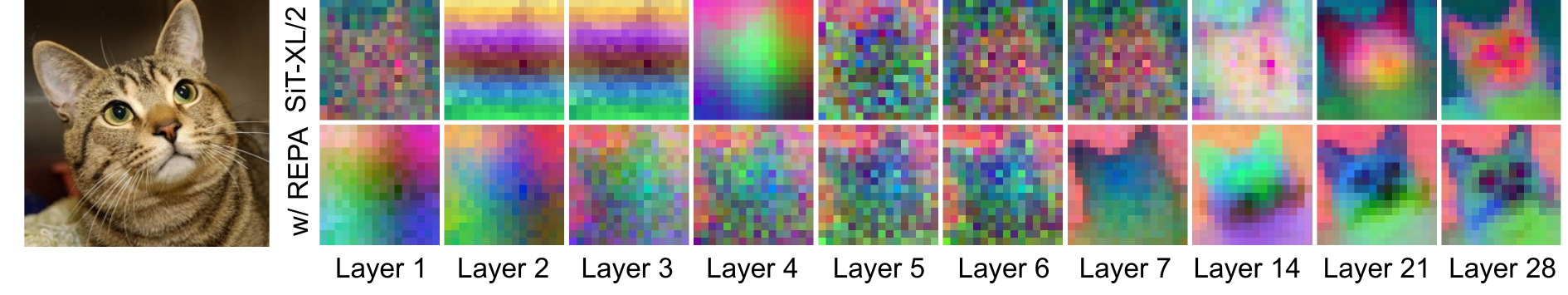}
        \caption{$t=0.5$}
    \end{subfigure}
    
    \begin{subfigure}[b]{\textwidth}
        \centering
        \includegraphics[width=\textwidth]{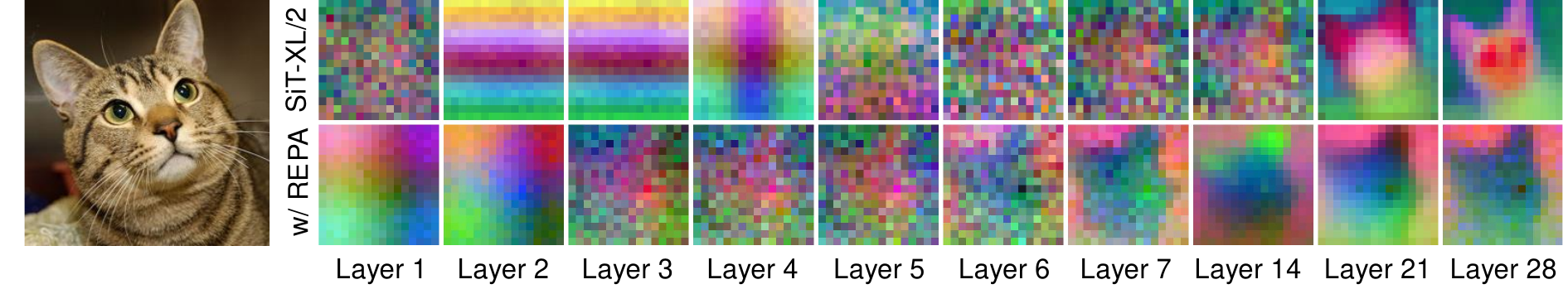}
        \caption{$t=0.7$}
    \end{subfigure}%

    \begin{subfigure}[b]{\textwidth}
        \centering
        \includegraphics[width=\textwidth]{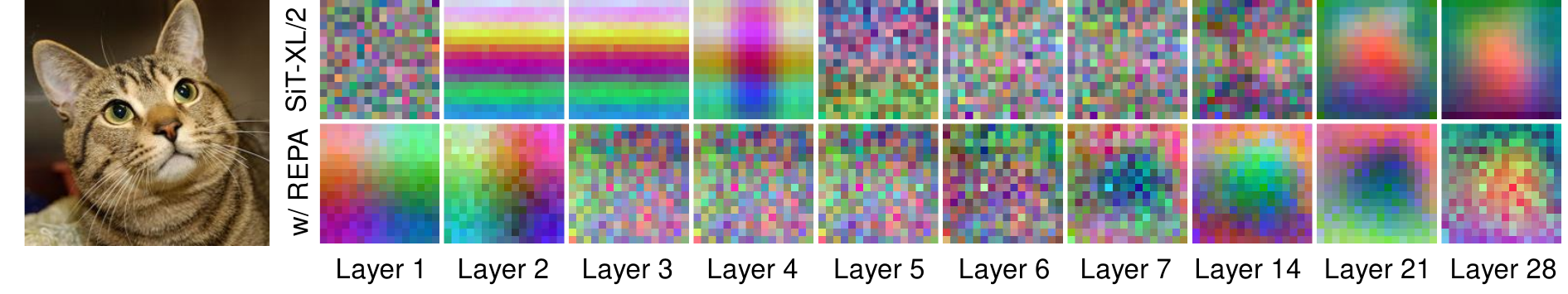}
        \caption{$t=0.9$}
    \end{subfigure}%
    \caption{\textbf{PCA visualization} of layer-wise features of SiT-XL/2 and SiT-XL/2+REPA.}
    \label{fig:pca}
\end{figure*}

We provide PCA visualizations of feature map, similar to those in DINOv2 \citep{oquab2024dinov}. As shown in Figure~\ref{fig:pca}, \sname shows coarse-to-fine feature maps, while the vanilla model tends to show noisy feature map particular for large $t$.

\clearpage
\section{Limitations and Future Work}
In this section, we enumerate several possible future research directions in what follows.

\textbf{Alignment depth.}
Recall that we empirically showed that applying REPA to layer 8 is more beneficial than to later transformer layer embeddings (see Table~\ref{tab:detailed_design}); conducting extensive analysis of this result will be an interesting direction to further improve \sname.

\textbf{Different input data types.}
We mainly focused on latent diffusion in the image domain. Exploring REPA with pixel-level diffusion or on other data domains like videos would be an interesting future work. Moreover, based on our text-to-image generation results on MS-COCO, training large-scale text-to-image diffusion models with REPA will also be an interesting direction.

\textbf{Theoretical analysis.}
Exploring theoretical insights into why REPA works well will also be an exciting future direction. For instance, it will be interesting to explore the relationship between representations learned with an instance discrimination objective and a denoising objective.

\textbf{Time-varying \sname.}
We think one of the interesting possible directions can be designing a weight function based on a noise schedule used in the diffusion process. We have not explored this in this work as our main focus is more on performing extensive analysis on other perspectives, such as target representations used for alignment, alignment depth, scalability of the method, \emph{etc.} We leave this as an exciting direction for future work.

\label{appen:limitation}

\end{document}